\documentclass[10pt,twocolumn,letterpaper]{article}

\usepackage{titling}
\usepackage{ifthen}
\newboolean{main}
\setboolean{main}{true}
\newboolean{suppmat}
\setboolean{suppmat}{true}

\usepackage{cvpr}              %
\usepackage[dvipsnames]{xcolor}

\usepackage{graphicx}
\usepackage{amsmath}
\usepackage{amssymb}
\usepackage{booktabs}
\usepackage{bm}

\usepackage{cancel}

\usepackage{svg}

\usepackage{tikz}
\usetikzlibrary{shapes.geometric}
\usetikzlibrary{graphs}
\usetikzlibrary{shapes.misc}
\usetikzlibrary{arrows.meta,decorations.pathreplacing, positioning, calc}
\usetikzlibrary{through}
\usetikzlibrary{intersections}

\usepackage{breqn}
\usepackage{booktabs}
\usepackage{makecell}
\usepackage{multirow}
\usepackage{nicematrix}

\usepackage[noend,ruled,linesnumbered]{algorithm2e}
\SetKwInput{KwInput}{Input}                %
\SetKwInput{KwOutput}{Output}              %
\SetKwInput{KwInitialize}{Initialize}      %

\usepackage{amsthm}

\usepackage{multirow}
\usepackage{makecell}

\usepackage{array}
\newcolumntype{L}[1]{>{\raggedright\let\newline\\\arraybackslash\hspace{0pt}}m{#1}}
\newcolumntype{C}[1]{>{\centering\let\newline\\\arraybackslash\hspace{0pt}}m{#1}}
\newcolumntype{R}[1]{>{\raggedleft\let\newline\\\arraybackslash\hspace{0pt}}m{#1}}

\usepackage[pagebackref,breaklinks,colorlinks]{hyperref}
\hypersetup{
    colorlinks=true,
    allcolors=Black, %
    pdfpagemode=FullScreen,
    }

\SetKwComment{Comment}{\color{RoyalBlue}$\triangleright$\ }{}

\usepackage[capitalize,nameinlink]{cleveref}
\crefdefaultlabelformat{#2\textbf{#1}#3}
\crefname{equation}{}{}
\creflabelformat{equation}{#2\textup{\bf #1}#3}
\crefname{equation}{Eq.}{Eqs.}
\Crefname{equation}{Equation}{Equations}
\crefname{figure}{Fig.}{Figs.}
\Crefname{figure}{Figure}{Figures}
\crefname{table}{Tab.}{Tabs.}
\Crefname{table}{Table}{Tables}
\crefname{section}{Sec.}{Secs.}
\Crefname{section}{Section}{Sections}
\crefname{problem}{Problem}{Problems}
\Crefname{problem}{Problem}{Problems}
\crefname{definition}{Definition}{Definitions}
\Crefname{definition}{Definition}{Definitions}
\crefname{lemma}{Lemma}{Lemmas}
\Crefname{lemma}{Lemma}{Lemmas}
\crefname{theorem}{Thm.}{Thms.}
\Crefname{theorem}{Theorem}{Theorems}
\crefname{remark}{Rmk.}{Rmks.}
\Crefname{remark}{Remark}{Remarks}
\crefname{enumi}{item}{items}
\Crefname{enumi}{Item}{Items}
\crefname{algocf}{Alg.}{Algs.}
\Crefname{algocf}{Algorithm}{Algorithms}
\crefname{assumption}{Asm.}{Asms.}
\Crefname{assumption}{Assumption}{Assumptions}
\crefname{ALC@unique}{line bla}{lines}
\Crefname{ALC@unique}{Line bla}{Lines}

\newcommand\colorSurface{Black}
\newcommand\colorSqOne{Gray}
\newcommand\colorSqTwo{blue}
\newcommand\colorSqThree{ForestGreen}

\newcommand\colorSqOneText{Gray!60!Black}
\newcommand\colorSqTwoText{blue!60!Black}
\newcommand\colorSqThreeText{ForestGreen!60!Black}

\newcommand\colorAxis{black}
\newcommand\colorCircle{RoyalBlue}

\tikzset{
    point/.style={
        very thick,
        cross out,
        inner sep=0pt,
        minimum width=6pt,
        minimum height=6pt,
    },
}

\def\paperID %
\def\confName{\xspace}
\def\confYear{\xspace}

\def\cvprPaperID{122} %
\def\confName{3DV}
\def\confYear{2024}

\definecolor{amber}{rgb}{1.0, 0.75, 0.0}

\usepackage{enumitem}

\newlist{rquestions}{enumerate}{1}
\setlist[rquestions,1]{
    label={\bf RQ\arabic*:},
    ref=\arabic*, %
    labelwidth=!,
    align=left,
    itemindent=0pt,
    leftmargin=30pt,
    }
\creflabelformat{rquestionsi}{#2#1#3} %
\crefname{rquestionsi}{research question number}{research questions number} %
\Crefname{rquestionsi}{Research question number}{Research questions number}

\newlist{tree}{enumerate}{1}
\setlist[tree,1]{
    label={{\bf Tree \arabic*:}},
    ref=\arabic*, %
    labelwidth=!,
    align=left,
    itemindent=0pt,
    leftmargin=34pt,
    }
\creflabelformat{treei}{#2\textup{\bf #1}#3} %
\crefname{treei}{Tree}{Trees} %
\Crefname{treei}{Tree}{Trees}

\newcommand{\paperpar}[1]{%
\vspace{.10cm}\noindent%
{\bf #1:}%
}

\makeatletter
  \def\my@tag@font{\normalsize}
  \def\maketag@@@#1{\hbox{\m@th\normalfont\my@tag@font#1}}
  \let\amsmath@eqref\eqref
  \renewcommand\eqref[1]{{\let\my@tag@font\relax\amsmath@eqref{#1}}}
\makeatother

\usepackage{graphicx}
\usepackage{tikz}
\usetikzlibrary{positioning}
\usetikzlibrary{shapes.geometric}
\usetikzlibrary{spy}
\usepackage{graphicx}
\usepackage{calc}
\usepackage{wrapfig}

\def\magscale{0.5}
\def\scale{.49}
\def\figwidth{\scale\columnwidth}

\newsavebox{\boxA}\savebox{\boxA}{\includegraphics[width=\figwidth]{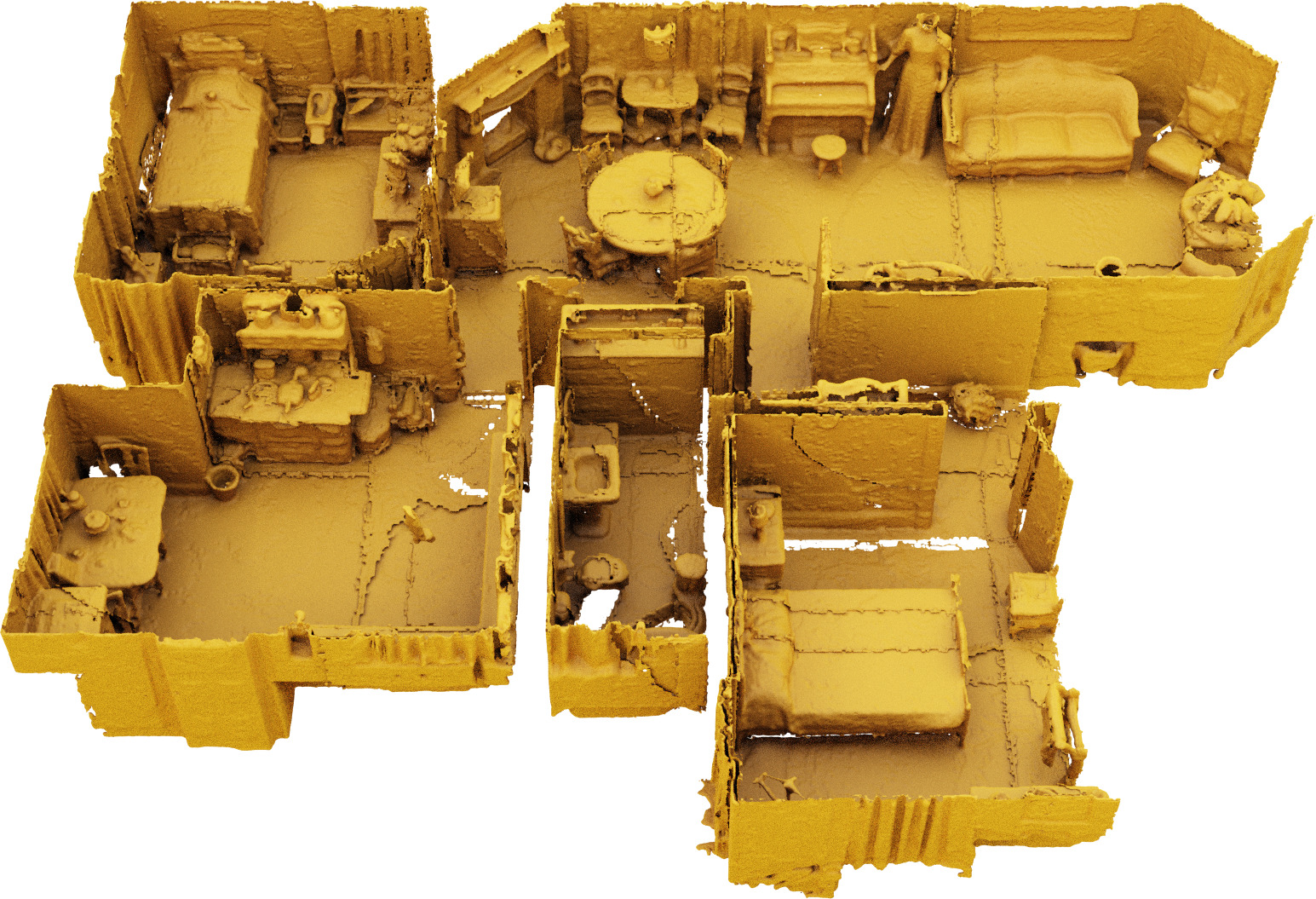}}
\newlength\hA
\newlength\wA

\newcommand\ZoomPictureA[5]{%
\begin{tikzpicture}[
        spy using outlines={%
        rectangle, %
        magnification=#5,
        size=\magscale\hA,
        connect spies
        }
    ]
    \node[
    	inner sep=0pt,
    	anchor=south west,
    	outer sep=0pt
    ] {\pgfimage[width=\wA]{#1}};
    \spy[blue,very thick] on (#2\wA,#3\hA) in node[anchor=south west] at (0,-25pt);
\end{tikzpicture}}

\newsavebox{\boxB}\savebox{\boxB}{\includegraphics[width=\figwidth]{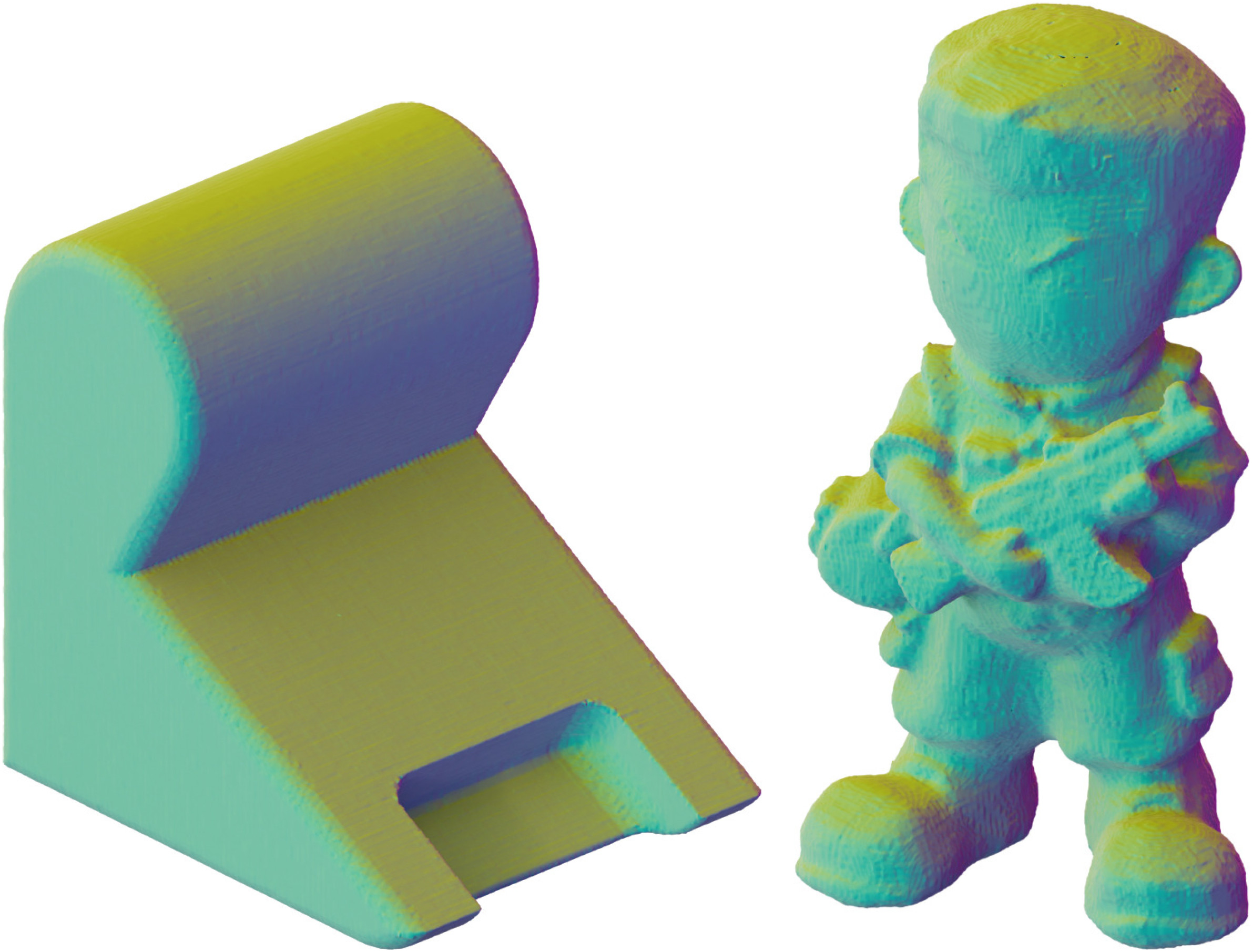}}
\newlength\hB
\newlength\wB

\newcommand\ZoomPictureB[5]{%
\begin{tikzpicture}[
        spy using outlines={%
        rectangle, %
        magnification=#5,
        size=\magscale\hB,
        connect spies
        }
    ]
    \node[
    	inner sep=0pt,
    	anchor=south west,
    	outer sep=0pt
    ] {\pgfimage[width=\wB]{#1}};
    \spy[blue,very thick] on (#2\wB,#3\hB) in node[anchor=south east] at (\wB,-30pt);
\end{tikzpicture}}

\begin{document}

\ifthenelse{\boolean{main}}
{
    
    \title{
    Oriented-grid Encoder for 3D Implicit Representations%
    }
    
    \author{Arihant Gaur\textsuperscript{1}\thanks{Equal contribution. Work done while an intern at MERL.} \hspace{50pt} G. Dias Pais\textsuperscript{1,2}\footnotemark[1] \hspace{50pt}  Pedro Miraldo\textsuperscript{1}\\[5pt]
    \begin{minipage}{0.5\textwidth}
        \centering
        \textsuperscript{1}Mitsubishi Electric Research Labs (MERL)\\
    \end{minipage}\hfill
    \begin{minipage}{0.47\textwidth}
        \centering
        \textsuperscript{2}Instituto Superior T\'ecnico, Lisboa\\
    \end{minipage}
    }

\maketitle%

    \begin{abstract}
Encoding 3D points is one of the primary steps in learning-based implicit scene representation. Using features that gather information from neighbors with multi-resolution grids has proven to be the best geometric encoder for this task. However, prior techniques do not exploit some characteristics of most objects or scenes, such as surface normals and local smoothness. This paper is the first to exploit those 3D characteristics in 3D geometric encoders explicitly. In contrast to prior work on using multiple levels of details, regular cube grids, and trilinear interpolation, we propose 3D-oriented grids with a novel cylindrical volumetric interpolation for modeling local planar invariance. In addition, we explicitly include a local feature aggregation for feature regularization and smoothing of the cylindrical interpolation features. We evaluate our approach on ABC, Thingi10k, ShapeNet, and Matterport3D, for object and scene representation. Compared to the use of regular grids, our geometric encoder is shown to converge in fewer steps and obtain sharper 3D surfaces. When compared to the prior techniques, our method gets state-of-the-art results.
    \end{abstract}

\section{Introduction}\label{sec:introduction}

\begin{figure}
    \centering%
    \input{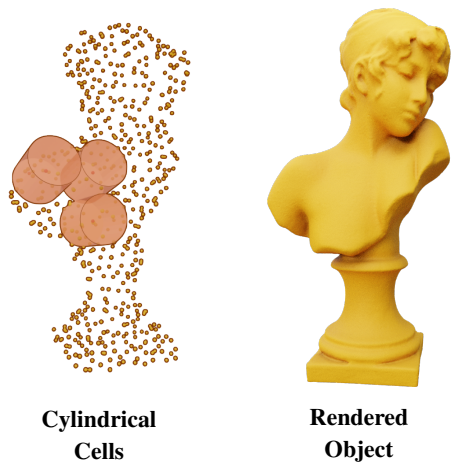}%
    \caption{%
\textbf{Teaser.} The proposed multi-resolution oriented grid extends the octree representation with the object's normal directions; cells are rotated to an orientation tree and the respective LOD (left). During training, the feature of a sampled point within the rotated cell is interpolated according to a new cylindrical interpolation scheme; neighboring cell features are aggregated with a 3DCNN. These features can be used in the current state-of-the-art decoders for object representation, such as SDFs and Occupancy. The object is rendered from the implicit representation (right).\\[0pt]%
 }
    \label{fig:teaser}
\end{figure}

There are many different ways of representing 3D surfaces. Implicit surface representations tell us that a point with coordinates $x$, $y$, and $z$ belongs to an object surface if $\mathcal{F}(x,y,z) = 0$, which defines the object. This type of 3D representation is advantageous since it is concise and guarantees continuity. Most learning-based 3D implicit representations start with encoding 3D points, then decoding their features into the chosen representation, defining $\mathcal{F}(.)$. 
Two kinds of encoders are used, most of the times in parallel: {\it i)} mapping the 3D coordinates of each point alone to a higher dimensional vector space, here denoted as positional encoder; and {\it ii)} 3D points gathering information about their neighbors, named grid-based. Multi-layer Perceptrons (MLPs) are usually considered a suitable choice for decoders. This paper focuses on the grid-based encoders, as illustrated in~\cref{fig:teaser}.

Although many neural implicit 3D surface representation methods have been proposed, such as \cite{hane2017hierarchical,groueix2018papier,mescheder2019occupancy,park2019deepsdf,tancik2020fourier,takikawa2021neural,lindell2022bacon}, %
previous techniques using geometric encoders still do not consider some of the object's underlying geometric characteristics, like normals, and only utilize its spatial localization. Since 3D surface representation is bound to a specific object/scene, we should use all the object's structural characteristics to model our geometric encoder.
Therefore, we propose a novel-oriented multiscale grid-based encoder. We consider a tree representation with multiple Levels-of-Detail (LOD) that capture different detail resolution~\cite{takikawa2021neural,yu2021plenoctrees}.
In contrast, the grids are constructed at each level using the object's surface occupancy and normals' orientation.
We then aim to answer the following research question:
{\it Does geometrically aligning the cell to the surface's normal help in representation accuracy?}

Our geometric encoder is the first to use surface orientations explicitly. From a multi-resolution surface representation, an abstract cell grid is aligned with the surface's normal (deeper LOD will have better-aligned normals). In addition to the surface alignment, we add the assumption that regular (constructed) surfaces are locally planar (smooth surfaces). Taking this assumption, the grid alignment with the surface is locally defined up to a rotation around its vector's normal surface, which we explicitly use to encode smoothness. The grid orientation and smoothness constraints are modeled by encoding features using cylinder grids, as shown in~\cref{fig:teaser}. Given the nature of the cylindrical coordinates (there are no corner-sharing features between neighboring grid cells), the new grid structure does not guarantee a smooth relationship between neighborhood cells. Thus, we propose using a shared 3D convolutional kernel for local feature aggregation. We summarize our contributions next:
\begin{enumerate}
    \item {\bf Oriented grids [\cref{subsec:irregular_grids}]:} A new sparse 3D tree representation aligning cells to the surface normals;\label{item:contributions_1}
    \item {\bf Cylindrical interpolation [\cref{subsec:cylindrical_inter}]:} A novel cylindrical interpolation scheme for cells in \cref{item:contributions_1};\label{item:contribution_2}
    \item {\bf Local feature aggregation [\cref{subsec:localfeatagg}]:} The interpolated feature vectors in \cref{item:contribution_2} are conditioned with a shared 3D convolutional kernel;\label{item:contributions_3}
    \item Results show that our method obtains state-of-the-art surface representation while capturing the object's structural regularity in more detail, lowering low-level roughness, and obtaining sharper 3D renderings.
\end{enumerate}

    \section{Related Work}\label{sec:related_work}

This section provides an overview of the related work on surface representation within the field of geometry processing, specifically from the perspective of our proposed method. We focus on geometric encoders, highlighting the differences concerning our method. We discuss various interpolation methods and list available pipelines for neural implicit representations.  

\paperpar{Grid-based encoders} 3D object neural encoders can be represented as either feature-based or a combination of feature and grid-based. The feature-based methods~\cite{qi2017pointnet,qi2017pointnet++,wang2017cnn,peng2021shape, yifan2019differentiable,riegler2020free,attal2020matryodshka,broxton2020immersive,dai2020neural,chabra2020deep,yifan2021iso,chibane20ifnet,park2019deepsdf,mescheder2019occupancy,peng2020convolutional,peng2021shape,chibane2020ndf,guillard2022udf,williams2022neural,boulch2022poco,huang2023neural}, encode a feature representation of a dataset of objects from the input point clouds and decode them to the desired output representation. Typically, such encoders have a large memory requirement and are more applicable in generalizability across all objects, often at the cost of quality. As a result, these methods frequently generate overly smoothed objects that lack intricate details. This paper focuses on \textbf{3D object implicit representation}~\cite{sitzmann2020implicit,lindell2022bacon,tancik2020fourier,chibane2020ndf}, that is, object bounded and not a general object encoder.

Recent advancements in neural representations~\cite{mildenhall2020nerf,wang2021neus} have led to the widespread use of encoders that combine grids and feature-based approaches. These grid-based encoders use a 3D grid to simplify the feature space and create an explicit 3D embedding~\cite{mescheder2019occupancy,peng2020convolutional}.
Recent works used grid-based encoders for neural 3D representations~\cite{fridovich2022plenoxels,tancik2020fourier,lindell2022bacon,chen2022tensorf}. These representations allowed the construction of a per-object implicit representation while preserving far more details.
More recent approaches have employed multiscale representations, such as octrees~\cite{riegler2017octnet,wang2017cnn,wang2018adaptive,wang2020deep,wang2021unsupervised,yu2021plenoctrees,takikawa2021neural}. They can a capture higher level of information in data structures~\cite{chien1986volume,laine2010efficient}. 
Plenoctrees and NGLOD~\cite{yu2021plenoctrees,takikawa2021neural} leverage the octree representation to achieve a smaller model and capture multiple resolutions of the 3D object.
Instant-NGP~\cite{muller2022instant} employs the octree representation with hash encoding to various tasks, including 3D object representation, and has demonstrated improved training efficiency.
Due to the success of multiscale representation, we use a tree-based representation, which decimates the original mesh into a set of grids for each LOD, according to an octree. %

\begin{figure}
    \centering
    \begin{subfigure}[T]{.35\linewidth}
    \centering
    \scalebox{0.34}{
        \begin{tikzpicture}
    \def\s{8cm}
    \def\ss{4cm}
    \def\angX{0.6}
    \def\angY{1}
    \def\arrowSize{1cm}

    \draw [\colorSurface, line width= 0.7mm] plot [smooth] coordinates { (\ss*0,\ss*1.85) (\ss*0.4,\ss*1.65) (\ss*1.0,\ss*1.45) (\ss*1.43,\ss*1.1) (\ss*1.33,\ss*0.45) (\ss*1.9,\ss*0)};
    \node (L) at (10mm, \ss + 25mm) [text=\colorSurface, rotate=-20, inner sep=0pt, font=\Large, minimum height=1em]{\bf Surface};

    \draw[fill=\colorSqOne,fill opacity=0.03, line width= 1.0mm, \colorSqOne] (0,0) rectangle (\s,\s) node (SqOne) {};

    \draw[fill=\colorSqTwo,fill opacity=0.35, line width= 1mm, \colorSqTwo] (\s, \s) rectangle (\s / 2, \s / 2) node (SqTwo) {};
    \draw[fill=\colorSqTwo,fill opacity=0.15, line width= .7mm, dashed, dash pattern=on 4pt off 1.9pt, \colorSqTwo] (\s / 2, 0) rectangle (\s, \s / 2) node (SqTwo) {};
    \draw[fill=\colorSqTwo,fill opacity=0.15, line width= .7mm, dashed, dash pattern=on 4pt off 1.9pt, \colorSqTwo] (0, \s / 2 ) rectangle (\s / 2, \s) node (SqTwo) {};

    \draw[fill=\colorSqThree,fill opacity=0.45, line width= 1.0mm, \colorSqThree] (\s / 2, \s / 2) rectangle (\s / 2 + \s / 4, \s / 2 + \s / 4) node (SqThree) {};
    \draw[fill=\colorSqThree,fill opacity=0.15, line width= .7mm, dashed, \colorSqThree] (\s / 4, \s / 2) rectangle (\s / 4 + \s / 4, \s / 2 + \s / 4) node (SqThree) {};
    \draw[fill=\colorSqThree,fill opacity=0.15, line width= .7mm, dashed, \colorSqThree] (\s / 4, 3 * \s / 4 ) rectangle (\s / 4 + \s / 4, \s ) node (SqThree) {};
    \draw[fill=\colorSqThree,fill opacity=0.15, line width= .7mm, dashed, \colorSqThree] (0, 3 * \s / 4 ) rectangle (\s / 4, \s ) node (SqThree) {};
    \draw[fill=\colorSqThree,fill opacity=0.15, line width= .7mm, dashed, \colorSqThree] (3 * \s / 4, 0 ) rectangle (\s , \s / 4 ) node (SqThree) {};
    \draw[fill=\colorSqThree,fill opacity=0.15, line width= .7mm, dashed, \colorSqThree] (2 * \s / 4, 0 ) rectangle (3 * \s / 4 , \s / 4 ) node (SqThree) {};    
    \draw[fill=\colorSqThree,fill opacity=0.15, line width= .7mm, dashed, \colorSqThree] (\s / 2, \s / 4 ) rectangle (3* \s / 4 , \s / 2 ) node (SqThree) {};
    
    \fill [\colorSurface] (\ss*1.15,\ss*1.37) circle [x radius=1mm, y radius=1mm];

    \draw[line width= 0.7mm, -{Stealth[length=3mm]}, \colorSurface] (\ss*1.15,\ss*1.37) -- (\ss*1.15 + \angX*\arrowSize, \ss*1.37 + \angY*\arrowSize);

    \node (A) at (\ss*1.15 + \angX*\arrowSize - 0.3cm, \ss*1.37 + \angY*\arrowSize) [text=\colorSurface, inner sep=0pt, font=\Large, minimum height=1em]{$\mathbf{n}$};

    \node[point, draw=\colorSqOneText] at (\ss,\ss) {};
    \node[point, draw=\colorSqTwoText] at (\ss + \ss / 2,\ss + \ss / 2) {};
    \node[point, draw=\colorSqThreeText] at (\ss + \ss / 4,\ss + \ss / 4) {};

\end{tikzpicture}
        }
    \caption{Structured octree.}
    \label{oriented_tree_rep1}
    \scalebox{0.34}{
        \begin{tikzpicture}
        \def\s{8cm}
        \def\ss{4cm}
        \def\angX{0.6}
        \def\angY{1}
        \def\arrowSize{1cm}

        \draw [\colorSurface, line width= 0.7mm] plot [smooth] coordinates { (\ss*0,\ss*1.85) (\ss*0.4,\ss*1.65) (\ss*1.0,\ss*1.45) (\ss*1.43,\ss*1.1) (\ss*1.33,\ss*0.45) (\ss*1.9,\ss*0)};
        \node (L) at (10mm, \ss + 25mm) [text=\colorSurface, rotate=-20, inner sep=0pt, font=\Large, minimum height=1em]{\bf Surface};

        \draw[fill=\colorSqThree,fill opacity=0.05, line width= 1.0mm, \colorSqOne] (0,0) rectangle (\s,\s) node (SqOne) {};
        \draw[fill=\colorSqThree,fill opacity=0.15, line width= 1.0mm, \colorSqTwo, rotate around={45:((\s / 2 + \s / 4, \s / 2 + \s / 4) )}] (\s, \s) rectangle (\s / 2, \s / 2) node (SqTwo) {};
        \draw[fill=\colorSqThree,fill opacity=0.25, line width= 1.0mm, \colorSqThree, rotate around={67:((\ss + \ss / 4,\ss + \ss / 4))}] (\s / 2, \s / 2) rectangle (\s / 2 + \s / 4, \s / 2 + \s / 4) node (SqThree) {};

        \fill [\colorSurface] (\ss*1.15,\ss*1.37) circle [x radius=1mm, y radius=1mm];
    
        \draw[line width= 0.7mm, -{Stealth[length=3mm]}, \colorSurface] (\ss*1.15,\ss*1.37) -- (\ss*1.15 + \angX*\arrowSize, \ss*1.37 + \angY*\arrowSize);
    
        \node (A) at (\ss*1.15 + \angX*\arrowSize - 0.3cm, \ss*1.37 + \angY*\arrowSize) [text=\colorSurface, inner sep=0pt, font=\Large, minimum height=1em]{$\mathbf{n}$};
    
        \node[point, draw=\colorSqOneText] at (\ss,\ss) {};
        \node[point, draw=\colorSqTwoText] (sqtwo) at (\ss + \ss / 2,\ss + \ss / 2) {};
        \node[point, draw=\colorSqThreeText] (sqthree) at (\ss + \ss / 4,\ss + \ss / 4) {};

        \draw[line width= 0.7mm, -{Stealth[length=3mm]}, \colorSqOneText] (\ss,\ss) -- (\ss*1.25 , \ss);

        \draw[line width= 0.7mm, -{Stealth[length=3mm]}, \colorSqTwoText] (\ss + \ss / 2,\ss + \ss / 2) -- (\ss + \ss / 2 + 0.8cm, \ss + \ss / 2 + 0.8cm);

        \draw[line width= 0.7mm, -{Stealth[length=3mm]}, \colorSqThreeText] (\ss + \ss / 4,\ss + \ss / 4) -- (\ss + \ss / 4 + 0.38cm,\ss + \ss / 4 + 0.92cm);
        
        \coordinate (E) at (\ss + \ss / 4 + 0.5cm,\ss + \ss / 4);
        \coordinate (F) at (\ss + \ss / 4 + 0.19cm,\ss + \ss / 4 + 0.40cm);
        \draw [\colorSqThreeText, line width= 0.2mm, -{Stealth[length=1.3mm]}] (E) to[out=90,in=0] node[name=action,font=\Large, xshift=3mm] {$\theta^2$} (F);

        \draw [\colorSqThree, very thick, dashed] (sqthree) -- (\ss + \ss / 4 + 0.8cm,\ss + \ss / 4);

        \coordinate (G) at (\ss + \ss / 2 + 0.5cm,\ss + \ss / 2);
        \coordinate (H) at (\ss + \ss / 2 + 0.30cm,\ss + \ss / 2 + 0.30cm);
        \draw [\colorSqTwoText, line width= 0.2mm, -{Stealth[length=1.3mm]}] (G) to[out=90,in=-30] node[name=action,font=\Large, xshift=3mm, yshift=1mm] {$\theta^1$} (H);

        \draw [\colorSqTwoText, very thick, dashed] (sqtwo) -- (\ss + \ss / 2 + 0.8cm, \ss + \ss / 2) ;

        \node[\colorSqOneText] (C) at (\ss + 0.5cm,\ss - 3mm) [inner sep=0pt, font=\Large]{$\theta^0$};

    \end{tikzpicture}
        }
    \caption{Final aligned grids using the tree in \protect\subref{oriented_tree_rep2}.}
    \label{oriented_tree_rep3}
    \end{subfigure}\hfill
    \begin{subfigure}[T]{.61\linewidth}
    \centering
    \scalebox{0.57}{
        \begin{tikzpicture}
    [level 1/.style={sibling distance=40mm},
     level 2/.style={sibling distance=40mm},
     level distance = 4cm,
     every node/.style={circle, draw=black,thin},
     emph/.style={edge from parent/.style={black,very thick,draw}},
     norm/.style={edge from parent/.style={black, thin,draw}},]

    \node[fill=\colorSqOne, fill opacity=0.10, text opacity=1, circle, draw=\colorSqOne, text=black, line width=1.0mm, align=left] (lvl0) {$\delta^0 = [-90^\circ,90^\circ[$,\\$\theta^0=0^\circ$}
      child {node[fill=\colorSqTwo, fill opacity=0.05, text opacity=1, circle, draw=\colorSqTwo, align=left, line width=1.0mm, dashed] {$\delta^1 = [-90^\circ,0^\circ[$,\\$\theta^1=-45^\circ$} 
      edge from parent node[left=0mm, draw=none] {\Large$\alpha = \theta^-$}
      }
      child[emph] {node[fill=\colorSqTwo,fill opacity=0.35, text opacity=1, circle, draw=\colorSqTwo, line width=1.0mm, align=left] {$\delta^1 = [0^\circ,90^\circ[$,\\$\theta^1=45^\circ$}
        child[norm] {node[fill=\colorSqThree, fill opacity=0.05, text opacity=1, circle, draw=\colorSqThree, align=left, line width=1.0mm, dashed] {$\delta^2 = [0^\circ,45^\circ[$,\\$\theta^2=22.5^\circ$}
        edge from parent node[left=0mm, draw=none] {\Large $\alpha = \theta^-$}
        }
        child[emph] {node[fill=\colorSqThree,fill opacity=0.45, text opacity=1, circle, draw=\colorSqThree, align=left, line width=1.0mm] {$\delta^2 = [45^\circ,90^\circ[$,\\$\theta^2=67.5^\circ$}
        edge from parent node[right=0mm, draw=none] {\Large \bm{$\alpha = \theta^+$}}
        }
        edge from parent node[right=0mm, draw=none] {\Large \bm{$\alpha = \theta^+$}}
      };

\end{tikzpicture}
        }
    \caption{Orientation tree to align cell grids marked with solid edges colors (gray, blue and green) at \protect\subref{oriented_tree_rep1}, obtaining the results in \protect\subref{oriented_tree_rep3}.}
    \label{oriented_tree_rep2}
    \end{subfigure}
    \caption{\textbf{Oriented Grid Construction.} Taking a 2D example with only one DoF ($\theta$), we model the structure of the object by using an octree, as shown in \protect\subref{oriented_tree_rep1}. However, for each level, we use an orientation tree that searches for the appropriate rotation/anchor, as shown in \protect\subref{oriented_tree_rep2}. From the orientation tree, we obtain the $\theta^l$ for LOD $l$ (as the middle point of limits $\delta^l$) in a coarse-to-fine manner that fits the object's surface. At each level, for each possible action, the query's normal $\mathbf{n}$ produced angle is compared against the angle range of each child. According to the result, we take the appropriate action $\alpha$, shown in bold in \protect\subref{oriented_tree_rep2}, to obtain the next level anchor. The cells of the original octree in \protect\subref{oriented_tree_rep1} are rotated according to the chosen anchor per level, obtaining the results shown in \protect\subref{oriented_tree_rep3}.}
    \label{fig:trees}
\end{figure}
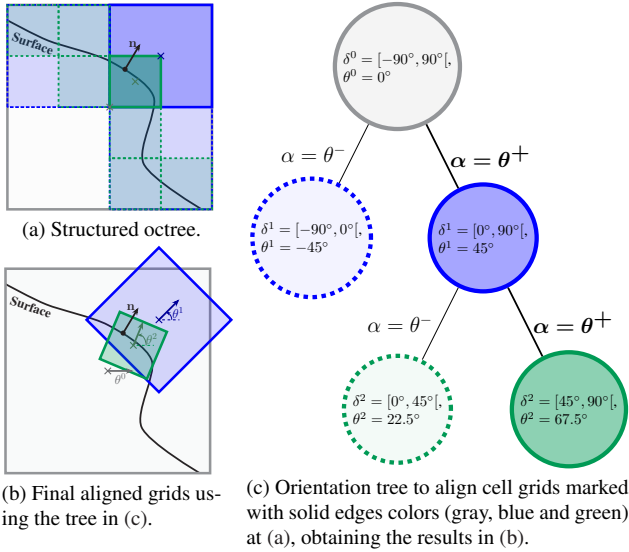

\paperpar{Interpolation} Aggregating features to capture local or global information about 3D objects is critical. Approaches that require a grid representation often use interpolation as an aggregating tool due to its simplicity and effectiveness. For instance, \cite{mescheder2019occupancy,lombardi2019neural,peng2020convolutional,sun2022direct,jiang2020local,liu2020neural,wang2021neus,Oechsle2021ICCV,fridovich2022plenoxels} use trilinear interpolation to obtain features per query inside a cell. \cite{fridovich2022plenoxels} discusses interpolation schemes, concluding the superior performance of trilinear interpolation over the nearest neighbor approach leading to optimal training convergence that is robust to different learning rate hyperparameters.

Similarly, multiscale grid features~\cite{yu2021plenoctrees,takikawa2021neural,muller2022instant} are obtained per LOD in an octree representation using trilinear interpolation, following the same principle. The encoder grid space can be trained in a coarse-to-fine fashion, \textit{i.e.} the features are trained using a per-level MLP for the same query point, where the corners for each cell are shared among levels.
The common factor in these approaches is that all exploit grids' regularity, making trilinear interpolation a natural choice. Due to the orientated cell property of our geometric encoder, we propose a new cylindrical interpolation scheme inspired by trilinear interpolation, along with a 3D sparse convolutional kernel shared among all LODs. The former warrants feature invariance between anchor rotations (since they are estimated up to a rotation) and the latter smoothness between neighboring cells and levels.

\paperpar{Decoders and Tasks} The oriented-grid encoder proposed in this paper can be applied to any type of decoder and representation task for implicit representations of 3D objects. 
In neural implicit methods focusing on per-object representation, the decoder typically consists of a small number of MLPs and employs a geometric grid-based encoder~\cite{sitzmann2020implicit,takikawa2021neural,martel2021acorn,wang2021neus,Oechsle2021ICCV,tancik2020fourier,lindell2022bacon,long2023neuraludf}.

In this work, the goal is to model the 3D representation to a surface mapping parameter. To this extent, the most commonly used ones are 
SDFs~\cite{hoppe1992surface,carr2001reconstruction,chen2019learning,park2019deepsdf,duan2020curriculum, wang2021neus,jiang2020local,liu2022learning}, occupancies~\cite{chen2019learning,mescheder2019occupancy,chabra2020deep,niemeyer2020differentiable,peng2020convolutional,lionar2021dynamic,tang2021sign,Oechsle2021ICCV}, or unsigned distance fields~\cite{chibane2020ndf,guillard2022udf,long2023neuraludf}. 
Once modeled, the mesh is reconstructed using the marching cubes algorithm~\cite{lorensen1987marching} or raytracing~\cite{amanatides1984ray,hart1996sphere} to render images from different viewpoints. 
Once the oriented octree grid is built, we generate a mesh representation from a point cloud. Our approach is agnostic to any output representation.

    \begin{figure*}[t]
    \resizebox{1\linewidth}{!}{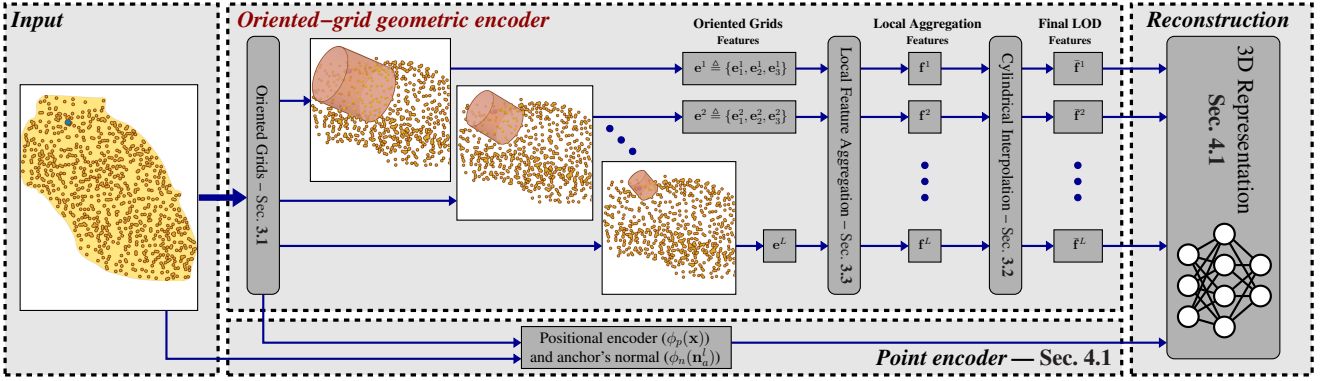}
    \caption{\textbf{3D reconstruction pipeline.} Graphical representation of a 3D surface reconstruction pipeline: grid-based and positional encoders are used, followed by a 3D reconstruction module. For each LOD $l$, the query point (in blue) is matched to a cell in the octree. The corresponding cell at each level aligns with the object's surface according to an anchor normal (\cref{subsec:irregular_grids}). A local aggregation 3DCNN computes the corresponding feature for each cell while considering its neighborhood (\cref{subsec:localfeatagg}). From the proposed cylindrical representation, a feature is interpolated (\cref{subsec:cylindrical_inter}) by evaluating the point's position inside the cylinder and the aggregated features. The final object is reconstructed from the interpolated feature, positional encoder, and normal encoder (\cref{subsec:3drecons}). Light gray boxes are the learnable layers in this figure, and the dark blocks represent the features.}
    \label{fig:networkarch}
\end{figure*}

\section{Oriented grid encoder}
\label{sec:oriented_grid_enc}

In the 3D implicit representation pipeline, a single 3D point passes through a geometric encoder, positional encoder, or both. The features are then injected into a decoder that models the object's surface, as in~\cref{fig:teaser}. Repeating the process for all point cloud points obtains a sparse output representation with respect to the modeled 3D surface.

The paper focuses mainly on the 3D grid encoder, specifically, the geometric grid encoder (orange block in~\cref{fig:networkarch}), detailed in this section. The input is a point and the pre-initialized trees that best fit the object. 
The output is a set of LOD tree features. 
We start by constructing the trees in~\cref{subsec:irregular_grids}. The features extracted from the trees are aggregated locally, crossing neighboring information (see~\cref{subsec:localfeatagg}). The aggregated features are used in the cylindrical interpolation scheme proposed in~\cref{subsec:cylindrical_inter}.

\subsection{Oriented grids construction}\label{subsec:irregular_grids}
    
Similar to previous works~\cite{chien1986volume, laine2010efficient, wang2017cnn, wang2020deep, takikawa2021neural,yu2021plenoctrees}, we use an octree representation to model the grid-based 3D encoder. However, in addition to the standard eight actions for splitting a grid into eight smaller ones in the subsequent LODs, we include rotation actions to model cell orientations, where at the higher levels a smaller (tighter) grid and a finer alignment better represent the object, as shown in \cref{fig:trees}. Instead of modeling each action individually, which would result in a branching factor of 56 --- $8 \ \text{(size)} \times 7 \ \text{(orientation)}$ -- per subsequent LOD, and since grid size and orientation are independent, we split them into two trees:
\begin{tree}
\item Structured octree for modeling the sizes of the grids; and\label{tree:octree}
\item Orientation tree for modeling the cell orientation.\label{tree:orientation}
\end{tree}
For the structured octree in \cref{tree:octree}, its representation consists of LODs $l \in \{1, \ldots, L\}$, bounded within $[-1, 1]^3$. We followed the typical octree modeling~\cite{takikawa2021neural, yu2021plenoctrees}.

\paperpar{Orientation tree}
For a normalized point $\mathbf{x} \in \mathbb{R}^3$ taken from the object's surface point cloud, we associate a normal to this query\footnote{We can either get the normal from a 3D oriented point cloud or estimate it from neighboring 3D points.}, denoted as $\mathbf{n} \in \mathbb{R}^3$.
The goal is to align the cells along the surface in a coarse-to-fine manner. To maintain consistency within the LODs, we construct a set of normal anchors representing the finite possible set of orientations per level. We then rotate the cells such that the $z$-axis matches the anchor, which is the closest anchor to the query normal $\mathbf{n}$.
To model a searching tree~\cite{russell2016artificial}, one needs to define: {\it i)} the node state, {\it ii)} the actions, {\it iii)} state transition, and {\it iv)} initial state:
\begin{itemize}
\item The node {\it state} denoted as $\delta$ is defined by a set of three 2-tuple elements:
\begin{equation}
\delta \triangleq \{ (r_{x^-}, r_{x^+}), (r_{y^-}, r_{y^+}), (r_{z^-}, r_{z^+}) \},
\end{equation}
where superscripts $-$ and $+$ indicate the lower and higher angle range limits.

\item We have seven possible {\it actions}:
\begin{equation}
    \mathcal{A} \triangleq \{ 0, x_-, x_+, y_-, y_+, z_-, z_+ \}.
\end{equation}

\item A state at LOD $l+1$ is obtained from a state at LOD $l$ after applying an action $\alpha\in\mathcal{A}$ using a {\it state transition} function $\delta_{\alpha}^{l+1} = \mathcal{T}(\delta^{l}, \alpha)$, such that:
{\footnotesize%
\begin{align}
    \mathcal{T}(\delta^l, x_-) & = \left\{ \left(r_{x^-}^{l}, \frac{r_{x^-}^{l} + r_{x^+}^{l}}{2}\right), (r_{y^-}^{l}, r_{y^+}^{l}), (r_{z^-}^{l}, r_{z^+}^{l}) \right\}
    \label{eq:action_x-} \\
    \mathcal{T}(\delta^l, x_+) & = \left\{ \left(\frac{r_{x^-}^{l} + r_{x^+}^{l}}{2}, r_{x^+}^{l}\right), (r_{y^-}^{l}, r_{y^+}^{l}), (r_{z^-}^{l}, r_{z^+}^{l}) \right\}
\label{eq:action_x+}
\end{align}}%
for actions $x_-$ and $x_+$. Action $0$ means no rotation; $\delta^{l+1}_{0} = \delta^{l}$. The remaining actions $\{y_-, y_+, z_-, z_+\}$ can be derived from \cref{eq:action_x-,eq:action_x+}.

\item The {\it initial state} is set as $\{(-\pi, \pi), (-\pi, \pi), (-\pi, \pi)\}$.
\end{itemize}

We conclude orientation tree modeling by defining rotations. A rotation is obtained per LOD and cell from state $\delta$. The {\it rotation anchor} $\mathbf{n}_a$ is computed from the states' Euler angles range $(r_x, r_y, r_z)$, where $r_i = (r_{i^+} + r_{i^-})/2, \forall i \in \{x, y, z\}$.
To compute the state $\delta$ for each cell, we allow the rotation anchors to align the $z$--axis of the cell with the surface normal, using cosine similarity, up to a rotation\footnote{Multiple solutions from the search can align to the surface normal. Since we restrict the alignment of the normal to the $z$--axis, the $x$ and $y$ axes can move freely. An example of this can be seen in the supplementary material.}.

\paperpar{Grid to query point association}
Each cell in~\cref{tree:octree} has a fixed orientation computed from searching~\cref{tree:orientation}. During training and evaluation, a query point is associated with a cell in~\cref{tree:octree} (structured tree) on a particular LOD. Then, the cell is rotated using the corresponding rotation anchor.
We note that a point may be outside all octree cells. In this case, we discard the query.

We note that \textit{the normal of the input points is not used in the model}; the normals are only used to construct the oriented grid.
Also, the orientation tree, from low-to-high LODs, will incorporate a coarse-to-fine approximation to the input normals.
This means that the orientation tree will be mostly invariant to small noise levels.
At lower LODs, rotations obtained by noisy and noiseless normals might differ slightly.
In addition, we highlight that this work aims to build an implicit representation of an object. The normals used to construct the encoder are typically accurate.
During training and evaluation, we use the cell's anchor normal.

\subsection{Cylindrical interpolation}\label{subsec:cylindrical_inter}

While trilinear interpolation has been the typical way of obtaining the features for regular grids~\cite{wang2021neus,Oechsle2021ICCV,yu2021plenoctrees,takikawa2021neural}, for oriented grids, using the same approach as regular grids is not appropriate since the cells do not share corners between neighbors (cells are rotated). There is rotation ambiguity in the $z$--axis. So, we propose to use oriented cylinders, as shown in~\cref{fig:teaser,fig:networkarch}, that can exploit the alignment of the cells and mitigate the lack of invariance in defining the grid orientation, as discussed above. An example can be seen in the supplementary material. This rotation invariance adds an explicit smoothness constraint to points inside the grid.

\begin{figure}[t]
    \resizebox{1\linewidth}{!}{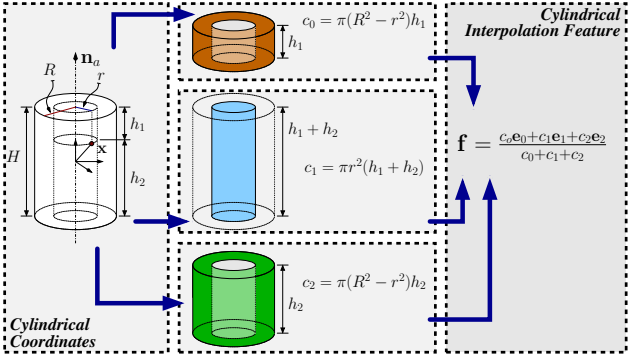}
    \caption{\textbf{Cylindrical interpolation.} The input cell grid has a corresponding anchor's normal $\mathbf{n}_a$ obtained from \cref{subsec:irregular_grids}. The cylinder is aligned with the grid normal anchor, with a radius $R$ and height $H$. The interpolation scheme is of volumetric interpolation type. It depends on the distance of the query point $\mathbf{x}$ to the cylinder's height boundaries $h_1$ and $h_2$, and the distance between $\mathbf{x}$ and the cylindrical axis of symmetry, denoted as $r$. The first coefficient $c_0$ is computed from the distance of the point to the top plane $h_1$ and the difference in volumes considering $R$ and the point's distance to the axis of symmetry $r$ (orange). The coefficient $c_2$ is computed from the distance to the bottom plane $h_2$ and the difference in volumes considering $R$ and the point's distance to the axis of symmetry $r$ (green). Finally, $c_1$ is the remainder cylinder (blue). Each coefficient as an associate learnable feature $\mathbf{e}_k$ for $k = \{0, 1, 2\}$. The interpolated feature $\mathbf{f}$ is the weighted average of $\mathbf{e}_k$ with $c_k$ weights.}
    \label{fig:cylinterpol}
\end{figure}

Given the query point, the objective is to compute relative spatial volumes for feature coefficients, considering the point's relative position within the cylinder. Cylindrical interpolation coefficients measure the closeness of the point to the extremities of the cylinder cell representation, as depicted in~\cref{fig:cylinterpol}. A point closer to the top and the border of the cylinder will produce a lower volume for that boundary (yellow volume). Therefore, its distance from the opposite face will be high, thus, a higher volume coefficient (green volume). The highest coefficient in this example will be the opposite according to the center axis (blue volume). 
Finally, the volumetric interpolation gives us three coefficients, $c_0$, $c_1$, and $c_2$ (see \cref{fig:cylinterpol} for derivation). With these coefficients, learnable features $\mathbf{e}^l \triangleq \{ \mathbf{e}_0^l, \mathbf{e}_1^l, \mathbf{e}_2^l \}$, where $\mathbf{e}_k^l \in \mathbb{R}^F$, per cell\footnote{For simplicity, we omit the cell index.} and per LOD $l$, are used to interpolate the query point features (linear combination): $\mathbf{f}^l = \frac{c_0 \mathbf{e}_0^l + c_1 \mathbf{e}_1^l + c_2\mathbf{e}_2^l}{c_0 + c_1 + c_2}$. 

\subsection{Local Feature Aggregation}\label{subsec:localfeatagg}

The features obtained solely by the proposed interpolation scheme lack neighborhood and LOD information. In contrast to trilinear interpolation, where the features from the corners of the interpolation cell are shared among neighbors and levels, cylindrical interpolation does not inherently incorporate knowledge between its neighborhood and other LODs. Thus, we propose the usage of a shared across levels 3DCNN for local feature aggregation, here denoted as $g_{\theta_l}(.)$. The convolutional kernel $\mathcal{K}_k$ per feature level $k = \{0, 1, 2\}$ is shared across the tree. 
For each feature $\mathbf{e}^{l}_k$ at each cell, there is an associated kernel $\mathcal{K}_k$:
\begin{equation}
    \bar{\mathbf{e}}_k^l = g_{\theta_l}\left(\mathbf{e}_k^l, \mathcal{N}^l(\mathbf{x}) ,\mathcal{K}_k \right), \ \forall l,
    \label{eq:3dcnn}
\end{equation}
where $\mathcal{N}^l(\cdot)$ is the set of neighborhood cells. 
We utilize a 3D sparse convolution, \textit{i.e.} the neighbor is ignored if it is not present. The neighborhood comprises neighbor cells within the kernel, with the current cell as the center. 

At last, the interpolated features from the geometric encoder in~\cref{fig:networkarch} are the linear interpolation of the coefficients and the local feature computed from the 3DCNN (\cref{eq:3dcnn}):
\begin{equation}
    \bar{\mathbf{f}} = \frac{c_0 \bar{\mathbf{e}}_0 + c_1 \bar{\mathbf{e}}_1 + c_2 \bar{\mathbf{e}}_2}{c_0 + c_1 + c_2}.
\end{equation}

    \section{Implementation}\label{subsec:implementation}

\cref{subsec:3drecons} presents the encoder creation, used decoder architectures, and output representations. \Cref{subsec:losses} describes the training and evaluation procedures, and \cref{subsec:experimental_setup} details the experiments.

\subsection{Architecture encoder and decoders}\label{subsec:3drecons}
\paperpar{Encoder}
The dual-tree construction used to encode an object is a pre-processing operation. Different noise levels are added to the input points, following the standard protocol~\cite{takikawa2021neural}. %
This representation retains the LOD cells to which the object belongs and the respective anchor orientation. Each cell has its corresponding features, as described in \cref{subsec:cylindrical_inter}. During training and evaluation, a point is queried, and the features are extracted and passed to the decoder.

\paperpar{Decoders and output representations}
We use an MLP-based architecture as a decoder for the implicit representation. The decoder is trained at each level and shared across all LODs. Besides the input interpolated feature, we add a positional encoder~\cite{mildenhall2020nerf} $\phi_p(\cdot)$ on the point with $L_p$ frequencies, as well as $\phi_n(\cdot)$ on the anchor's normal with $L_n$ frequencies. The point and the normal are attached to each positional encoder, with the size of $P = 3 \times 2 \times L_p + 3$ and $N = 3 \times 2 \times L_n + 3$. We show our method for SDF and occupancy as output representation.

\subsection{Training and Evaluation}\label{subsec:losses}

During the training stage, we sample $N_q$ points from the input point cloud, determine where they lie in each LOD, and interpolate their features accordingly to the chosen voxels. We compute the sum of the squared errors of the predicted samples or the cross-entropy from the active LODs for SDF and occupancy, respectively. Additionally, we estimate the normals from the gradient~\cite{Oechsle2021ICCV, niemeyer2020differentiable}. Then, we compute the L2-norm between the computed normals and the anchor normals as a regularization term, weighted by $\alpha_{n}$. The two terms are added to obtain the final loss.

During the evaluation, we obtain uniformly distributed input samples from a unit cube with resolution $Q = 512^3$. We show results for the last LOD $L$, corresponding to finer LOD. The input query gets discarded if it doesn't match an existing octree cell. Finally, we obtain the mesh from the output using marching cubes~\cite{lorensen1987marching}.

\subsection{Experimental Setup}\label{subsec:experimental_setup}

Following prior works, we evaluate our 3D reconstruction quality using Chamfer Distance (CD), Normal Consistency (NC), and Intersection over Union (IoU) for each object.
The CD is computed as the reciprocal minimum distance for the query point and its ground truth match. We compute CD five times and show the mean. NC is the corresponding normals computed from the cosine similarity between the query normal (corresponding to the query point obtained during CD calculation) and its corresponding ground truth normal. We report the NC as a residual of the cosine similarity between both normals. IoU quantifies the overlap between two grid sets. %

\paperpar{Datasets}
We evaluate our method on three different datasets, ABC~\cite{koch2019abc}, Thingi10k~\cite{Zhou2016Thingi10KAD}, and ShapeNet~\cite{shapenet2015}. We sample 32 meshes each from Thingi10k and ABC and 150 from ShapeNet. We watertight ShapeNet meshes using~\cite{point-cloud-utils}.
This work is implemented in PyTorch~\cite{paszke2017automatic}. 

\paperpar{Model}
Our decoder architecture has one hidden layer of dimension $128$, with ReLU. %
Each voxel feature is represented as a $F = 32$ dimensional feature vector. The positional encoding for the query point and normal are represented with $L_p = L_n = 6$ frequencies. We consider the sparse 3D convolutions for local feature aggregation, with a kernel size $\mathcal{K}_k, \forall l$, $k=5$. The cylinder radius $R$ was empirically set to $R = (h_1+h_2) / \sqrt{2}$. Further parameter investigations are shown in the supplementary material.

Using the Adam optimizer~\cite{diederik2015adam}, we train our model for up to $100$ epochs, with a learning rate of $0.001$ and $\alpha_{n} = 0.1$. An initial sample size of $5 \times 10^6$ points is considered, with a batch size of $512$. Resampling is done after every epoch. The points are sampled from the surface and around its vicinity in equal proportions. It is also ensured that each voxel has at least 32 samples before surface sampling. We consider the LODs $\mathcal{L} = \{ 3, \ldots, 7 \}$ for all the datasets.

\paperpar{Baselines}
We compare our approach against state-of-the-art approaches frequency-based encoders BACON~\cite{lindell2022bacon}, SIREN~\cite{sitzmann2020implicit} and Fourier Features (FF)~\cite{tancik2020fourier}, and 3D sparse network NDF~\cite{chibane2020ndf}, which were trained with the supplied settings. We simplify NDF~\cite{chibane2020ndf} by voxelizing the point cloud and obtain the surface using marching cubes, using the same settings from~\cref{subsec:losses}, instead of their ball pivoting algorithm~\cite{bernardini1999ball}.
We also evaluate a regular grid direct approach against our method, where we adapt~\cite{takikawa2021neural}, which required smaller changes in our pipeline for a fair comparison against the oriented grids (see the supplementary materials for details). 
The surface reconstruction setup was the same for all methods, described in~\cref{subsec:losses}.

    \begin{table}[t]
    \caption{\textbf{Ablation studies.} The table shows different stages leading to the final encoder. Starting with oriented grids with trilinear interpolation, proposed cylindrical interpolation, and finally, local feature aggregation with 3DCNNs. The CD is multiplied with $10^{-5}$ and NC by $10^{-4}$.%
    }
    \label{tab:ablations}
    \resizebox{1\linewidth}{!}{\setlength{\tabcolsep}{1.5pt}\begin{NiceTabular}{ccccc ccc}[code-before =%
    \rectanglecolor{Gray!20}{1-1}{2-8}%
    ]%
    \toprule
    \multicolumn{5}{c}{\thead{Network options}} & \multicolumn{3}{c}{\thead{Results}} \\%
    \cmidrule(lr){1-5} \cmidrule(lr){6-8}
    \makecell{Trilinear\\Interpolation} & \makecell{Cylindrical\\Interpolation} &  \makecell{3DCNN\\3x3x3} & \makecell{3DCNN\\5x5x5} & \makecell{Normals\\Regularization} &
    {CD$\downarrow$} & {NC$\downarrow$} & {IoU$\uparrow$} \\ \midrule 
    {\large \checkmark} & & & & & 1.278 & 4.410 & 0.881 \\
    & {\large \checkmark} & & & & 0.570 & 4.348 & 0.990 \\
    & {\large \checkmark} & {\large \checkmark} & & & 0.434 & 4.094 & \textbf{0.998} \\
    & {\large \checkmark} & & {\large \checkmark} & & \textbf{0.431} & 4.093 & \textbf{0.998} \\
    & {\large \checkmark} & & {\large \checkmark} & {\large \checkmark} & 0.443 & \textbf{4.058} & \textbf{0.998} \\\bottomrule
    \end{NiceTabular}}
\end{table}
\begin{figure}
    \centering
    \begin{subfigure}[t]{0.15\linewidth}
        \includegraphics[width=1\linewidth]{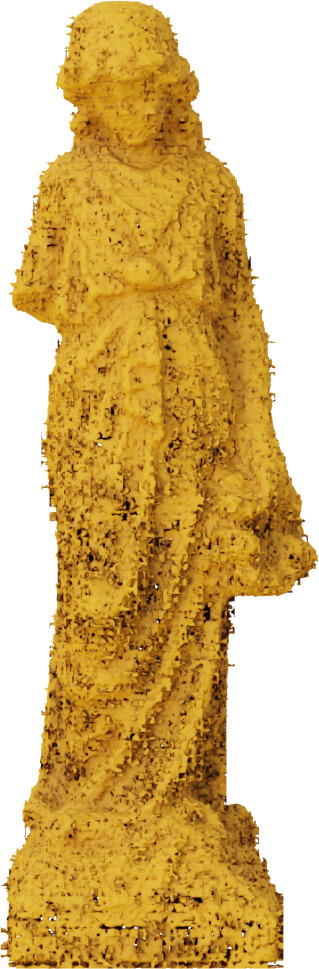}
        \caption{}\label{subfig:trilinear}
    \end{subfigure}%
    \begin{subfigure}[t]{0.15\linewidth}
        \includegraphics[width=1\linewidth]{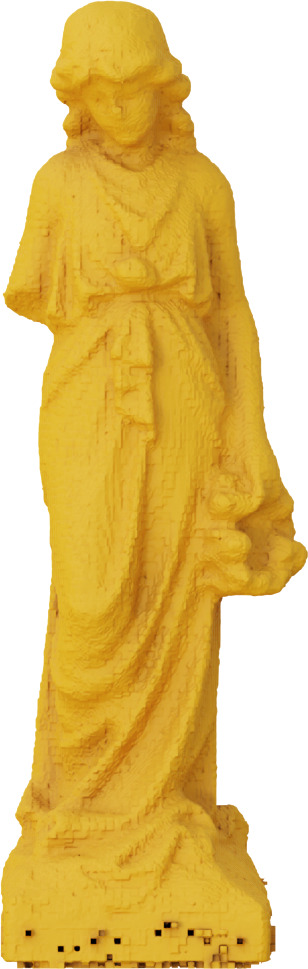}
        \caption{}\label{subfig:ci}
    \end{subfigure}%
    \begin{subfigure}[t]{0.15\linewidth}
    \includegraphics[width=1\linewidth]{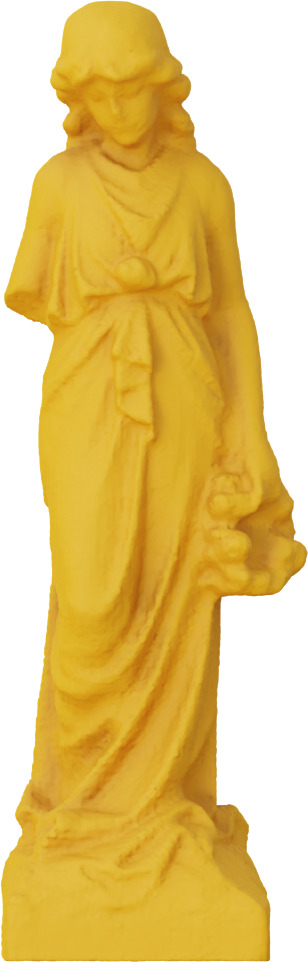}
        \caption{}\label{subfig:333cnn}
    \end{subfigure}%
    \begin{subfigure}[t]{0.15\linewidth}
        \includegraphics[width=1\linewidth]{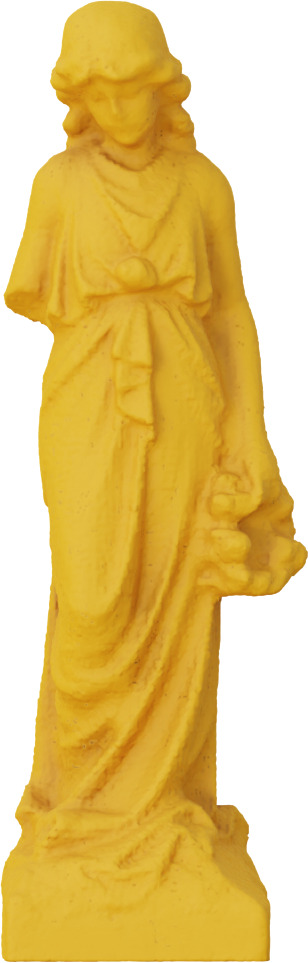}
        \caption{}\label{subfig:555cnn}
    \end{subfigure}%
    \begin{subfigure}[t]{0.15\linewidth}
        \includegraphics[width=1\linewidth]{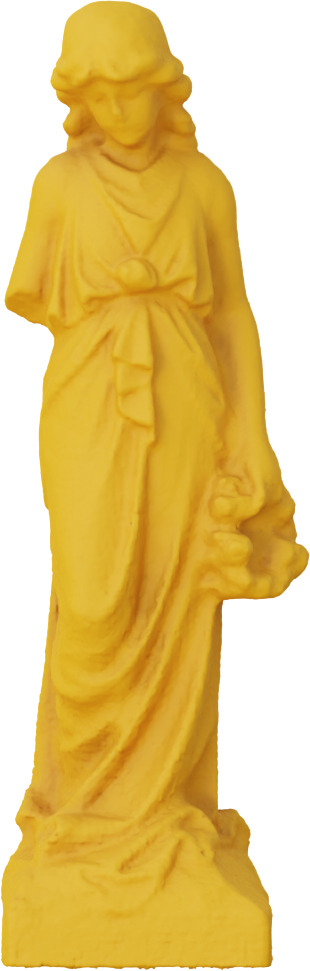} 
        \caption{}\label{subfig:norm}
    \end{subfigure}\hfill \vrule \hfill
    \begin{subfigure}[t]{0.15\linewidth}
        \includegraphics[width=1\linewidth]{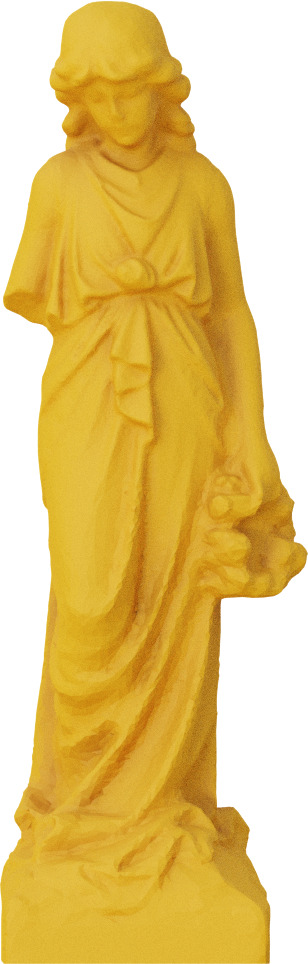}
        \caption{}\label{subfig:GT}
    \end{subfigure}%
    \caption{\textbf{Ablation example.} Ablation effects in rendering (numbers in \cref{tab:ablations}). \protect\subref{subfig:trilinear} represent the oriented encoder with trilinear interpolation; \protect\subref{subfig:ci} adds cylindrical interpolation; \protect\subref{subfig:333cnn} and \protect\subref{subfig:555cnn} use $3 \times 3 \times 3$ and $5 \times 5 \times 5$ 3DCNN kernels for feature aggregation, respectively; and \protect\subref{subfig:norm} adds normal regularization to \protect\subref{subfig:555cnn}. \protect\subref{subfig:GT} shows the ground-truth.}
    \label{fig:ablations_tab1}
\end{figure}

\section{Experiments}\label{sec:experiments}

Ablations are presented in~\cref{subsec:ablations}, where we evaluate the different stages of our method. \Cref{subsec:regular_oriented_experiments} shows the direct comparison between regular and oriented grids. Finally, we evaluate our method against methods using different encoder strategies in~\cref{subsec:baselines}.

\subsection{Ablations}\label{subsec:ablations}

We gradually add changes to different pipeline blocks to analyze each component's relevance. Ten meshes from ABC and Thingi10k datasets are randomly sampled for training and testing. \Cref{fig:ablations_tab1} shows different cases listed in \cref{tab:ablations}, based on the changes made to encoder.
We notice that using oriented grids with trilinear interpolation results in many holes. Since cells are rotated per the anchor normal, we use a rotation-invariant cylindrical representation for interpolation. Though rougher, this yields a more adapted representation (significant improvement in CD). 

We see a significant improvement in the mesh smoothness (reflected in NC) with the addition of 3DCNN (\cref{fig:ablations_tab1}~\subref{subfig:333cnn} and~\subref{subfig:555cnn}), effectively contributing to the local feature aggregation step. Our experiments show that the kernel of $5 \times 5 \times 5$ achieves better performance, preferred for subsequent experiments.
The proposed normal regularization enhances smoothness but sacrifices accuracy.
Additional ablations are discussed in the supplementary material.

\begin{table}[t]
    \caption{\textbf{Regular {\it vs.}~oriented grids.} Results for both types for SDF and occupancy decoders. The CD is multiplied with $10^{-5}$ and NC by $10^{-4}$.}
    \label{tab:orientedvsregular}
    \resizebox{1\linewidth}{!}{\setlength{\tabcolsep}{8.5pt}\begin{NiceTabular}{ccccccc}[code-before =%
    \rectanglecolor{Gray!20}{1-1}{2-7}%
    ]
    \toprule
    \multicolumn{4}{c}{\thead{Network options}} & \multicolumn{3}{c}{\thead{Results}} \\ %
    \cmidrule(lr){1-4} \cmidrule(lr){5-7}
    \makecell{Regular\\Grids} & \makecell{Oriented\\Grids} & \makecell{SDF\\Decoder} & \makecell{Occupancy\\Decoder} &
    {CD$\downarrow$} & {NC$\downarrow$} & {IoU$\uparrow$} \\ \midrule 
    {\large \checkmark} & & {\large \checkmark} & & 0.445 & 4.257 & 0.997 \\
    {\large \checkmark} & & & {\large \checkmark} & 0.687 & 4.117 & 0.989 \\
    & {\large\checkmark} & & {\large \checkmark} & 0.788 & 4.140 & 0.995 \\
    & {\large\checkmark} & {\large \checkmark} & & \textbf{0.443} & \textbf{4.058 } & \textbf{0.998} \\
    \bottomrule
    \end{NiceTabular}}
\end{table}

\begin{figure}
    {%
    \begin{subfigure}[t]{.49\linewidth}
    \captionsetup{justification=raggedleft,singlelinecheck=false,margin={14pt}}%
        \ZoomPictureA{scene_ours.jpg}{0.68}{0.9}{}{3}
        \vspace{-15pt}\hspace{-40pt}\caption{Oriented grids}\label{subfig:scene_oriented}
    \end{subfigure}\hfill%
    \begin{subfigure}[t]{.49\linewidth}
        \captionsetup{justification=raggedleft,singlelinecheck=false,margin={14pt}}%
        \ZoomPictureA{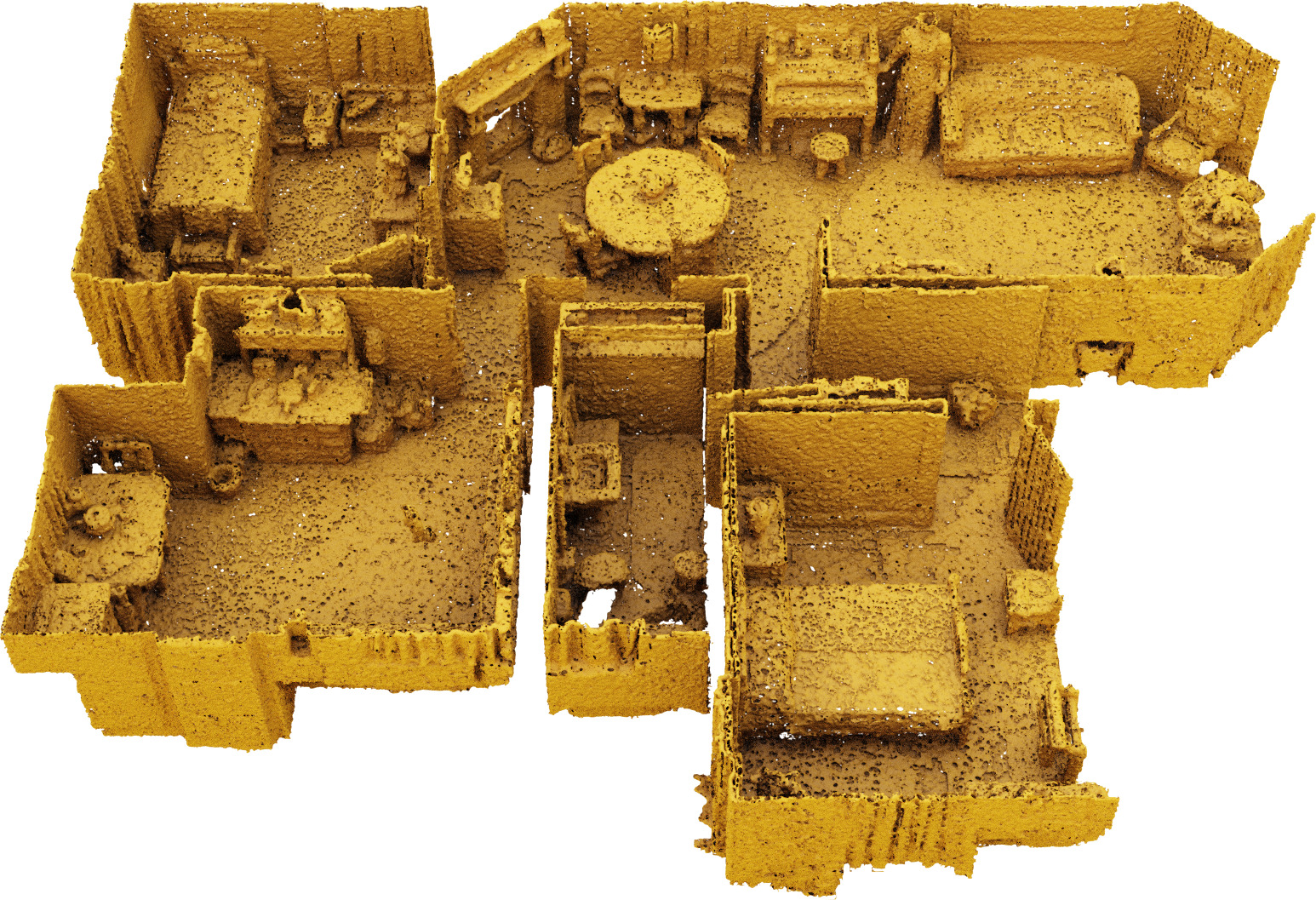}{0.68}{0.9}{}{3}
        \vspace{-15pt}\caption{Regular grids}\label{subfig:scene_regular}
    \end{subfigure}
    }
    \caption{\textbf{3DMatterport oriented {\it vs.}~regular grids.} 3D scene rendering from 3DMatterport dataset using oriented encoder in \protect\subref{subfig:scene_oriented} {\it vs.} regular encoder in \protect\subref{subfig:scene_regular}. The oriented grid encoder adapts well to different types of 3D representations. The occupancy decoder from regular grids cannot produce smooth and accurate results due to an unconstrained 3D representation. The objects presented are mostly planar and have thin surfaces, where our method excels due to the nature of the orientation grid and cylindrical interpolation. The result is a less rough and detailed 3D render. (Visible stitching effects from the scene splitting.)}
    \label{fig:experiments_sdf_regular_regular}
\end{figure}%

\begin{figure}
    {%
    \begin{subfigure}[t]{0.49\linewidth}
        \captionsetup{justification=raggedright,singlelinecheck=false,margin={5pt}}%
        \ZoomPictureB{ours_object.jpg}{0.22}{0.37}{}{3}
        \vspace{-15pt}\caption{Oriented grids}
    \end{subfigure}\hfill
    \begin{subfigure}[t]{0.49\linewidth}
        \captionsetup{justification=raggedright,singlelinecheck=false,margin={5pt}}%
        \ZoomPictureB{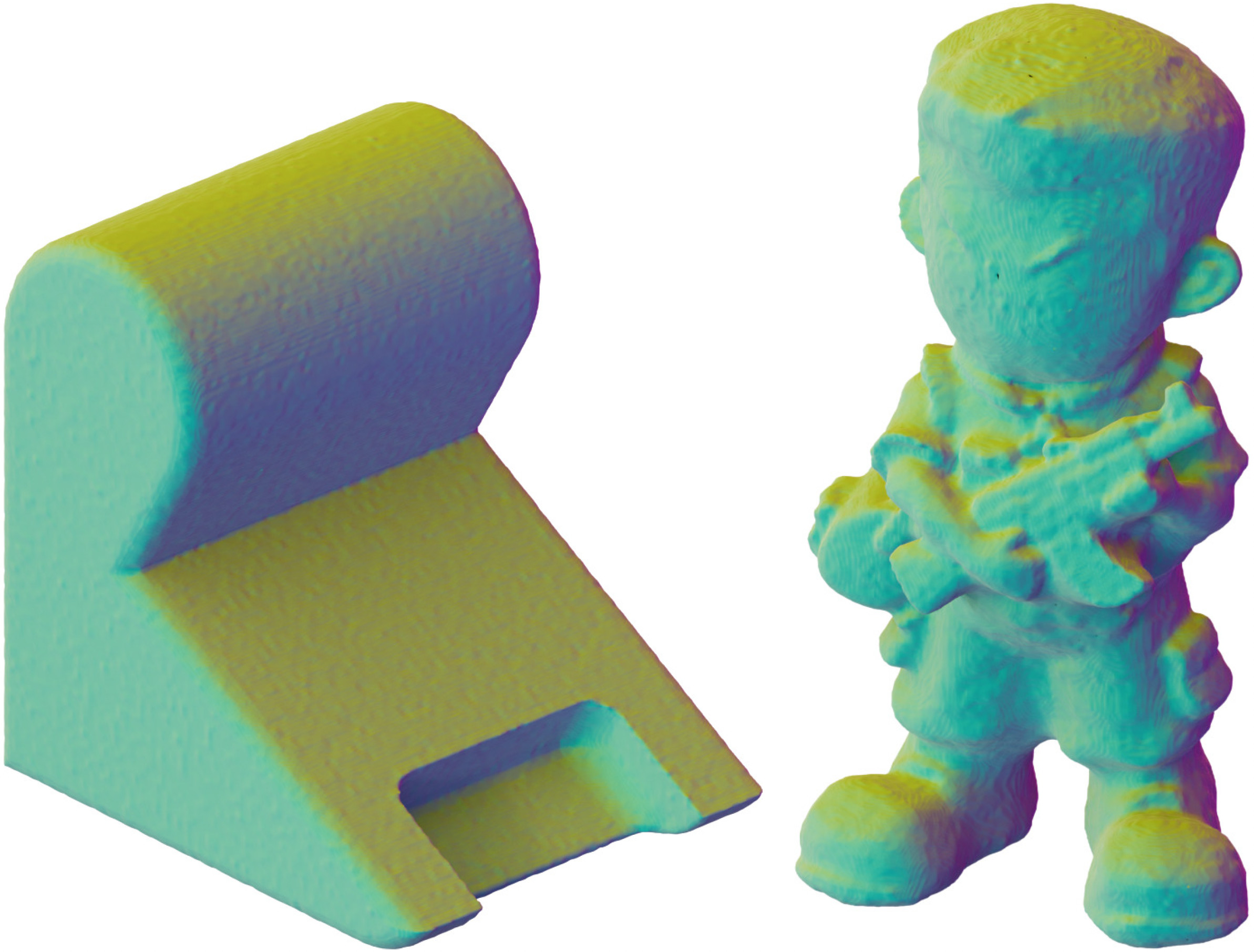}{0.22}{0.37}{}{3}
        \vspace{-15pt}\caption{Regular grids}
    \end{subfigure}
    }
    \caption{\textbf{Normal vizualization regular {\it vs.}~oriented grids.} Reconstruction of two objects -- left ABC, right Thingi -- after one epoch, with SDF decoder (without regularization). The caustic-like effect/surface noise reduction on the object's surface is noticeably visible, even in more detailed objects, after just one epoch in the oriented grids.}
    \label{fig:experiments_sdf_ours_regular}
\end{figure}%

\begin{figure*}[t]
    \resizebox{1\textwidth}{!}{\setlength{\tabcolsep}{0pt}\begin{NiceTabular}
    {@{}c c c c c c@{}}%
    \includegraphics[width=0.18\linewidth,keepaspectratio]{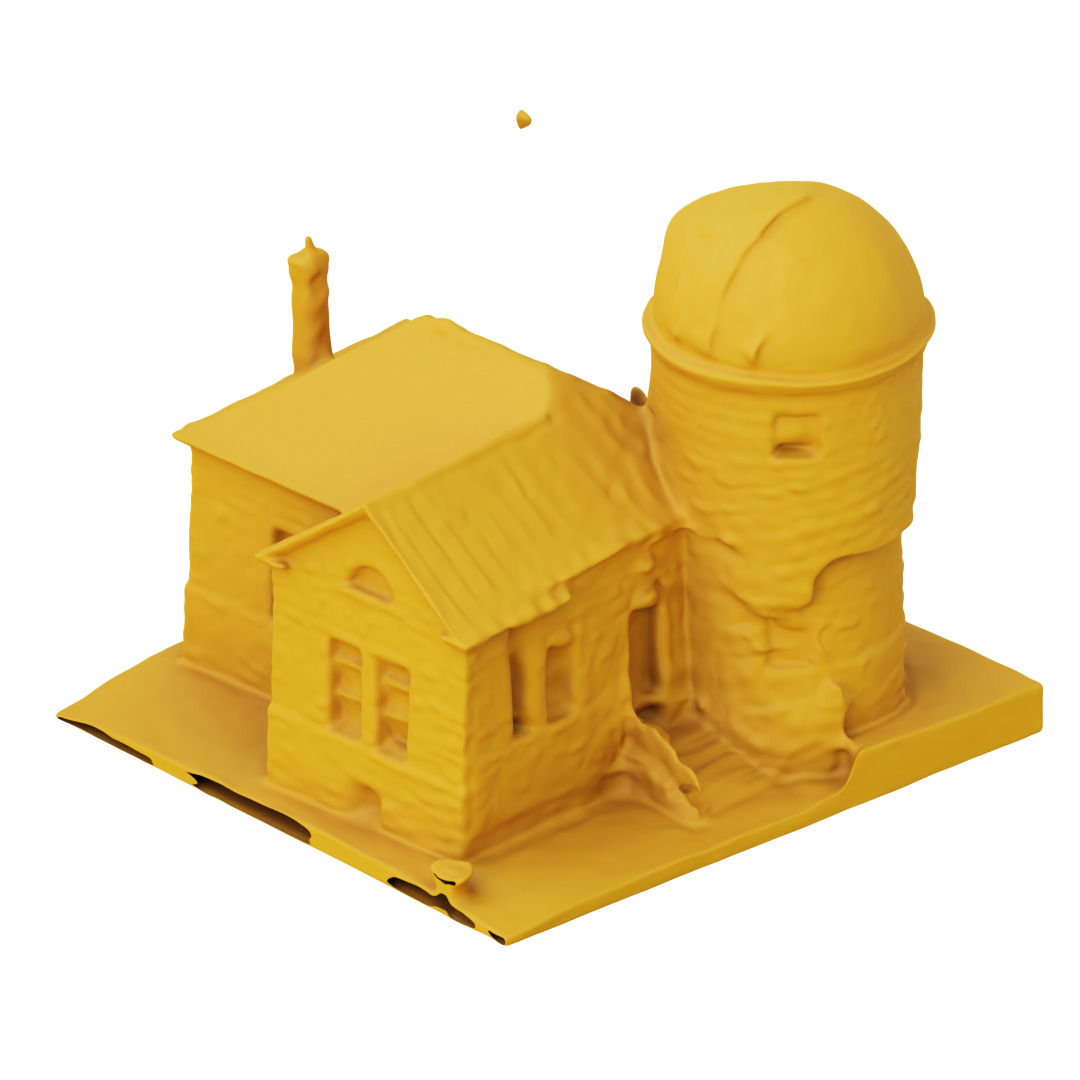}
    & \includegraphics[width=0.18\linewidth,keepaspectratio]{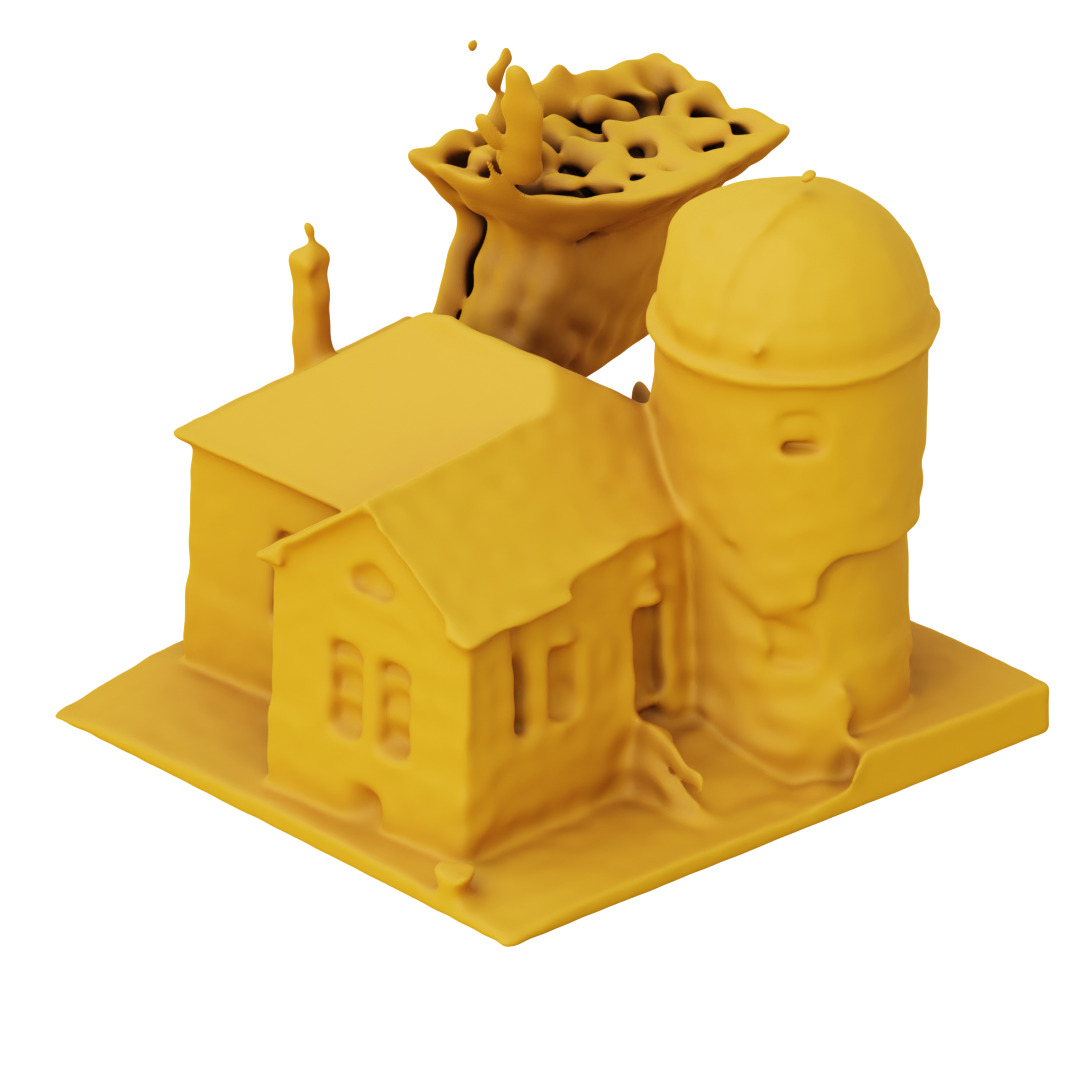}
    & \includegraphics[width=0.18\linewidth,keepaspectratio]{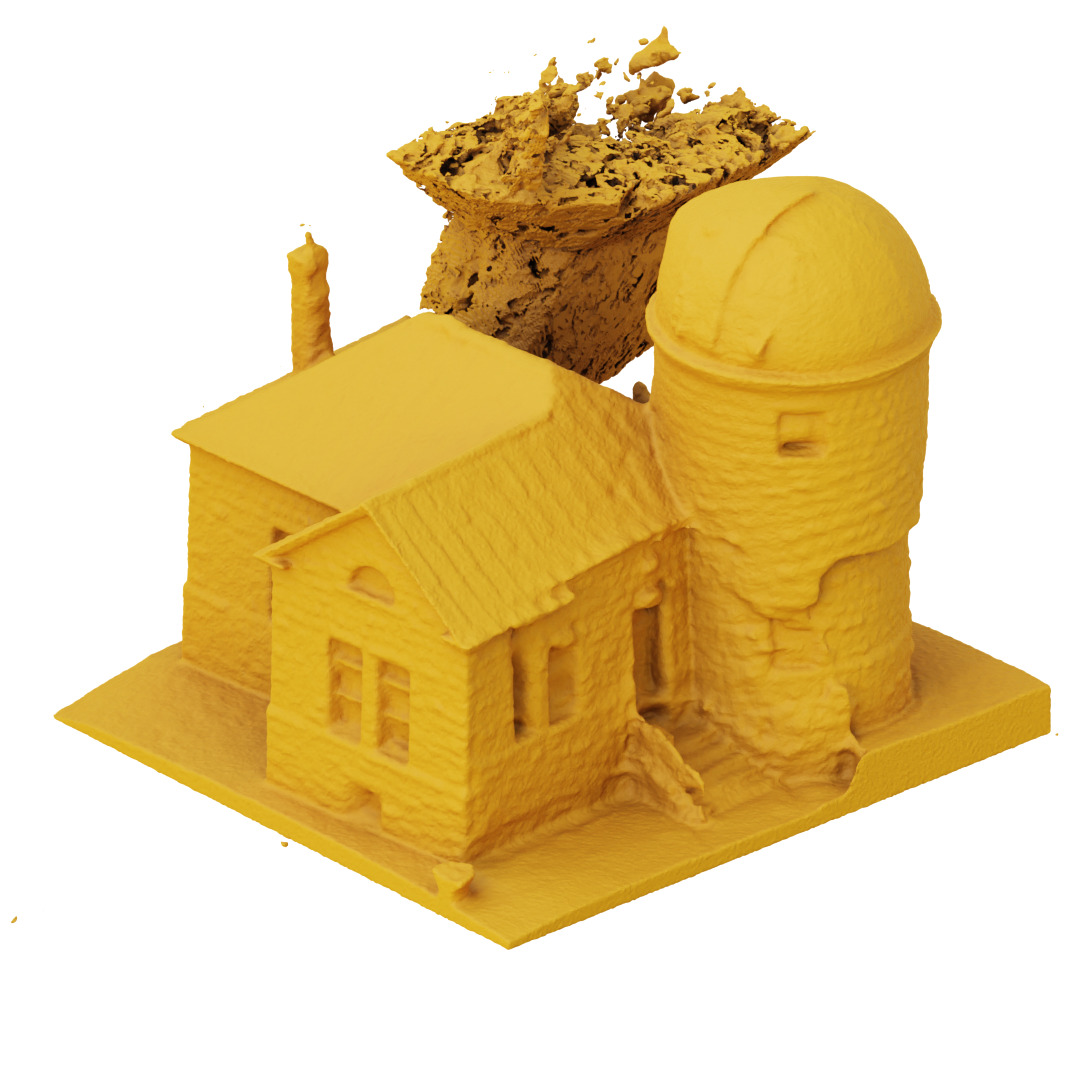} %
    & \includegraphics[width=0.18\linewidth,keepaspectratio]{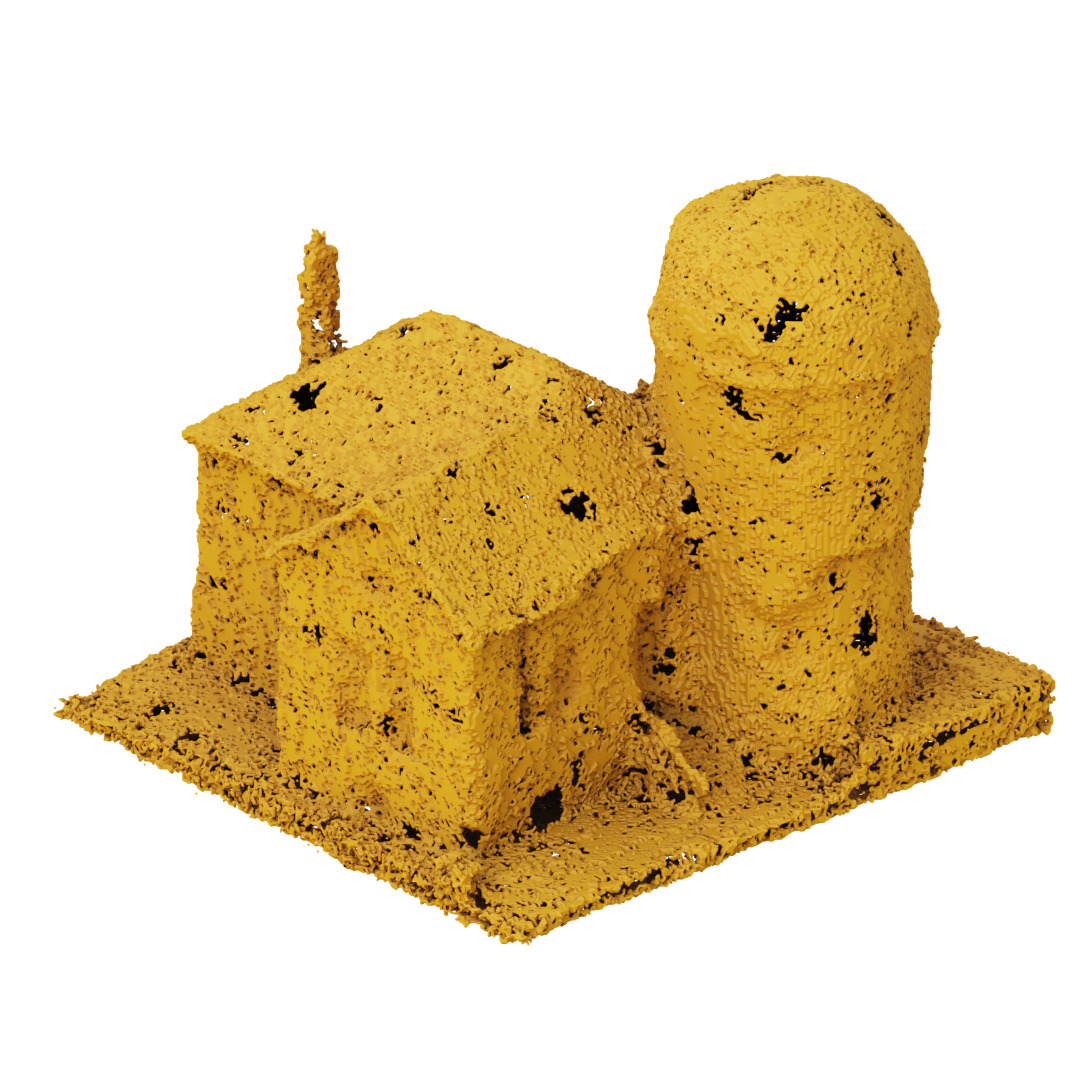}
    & \includegraphics[width=0.18\linewidth,keepaspectratio]{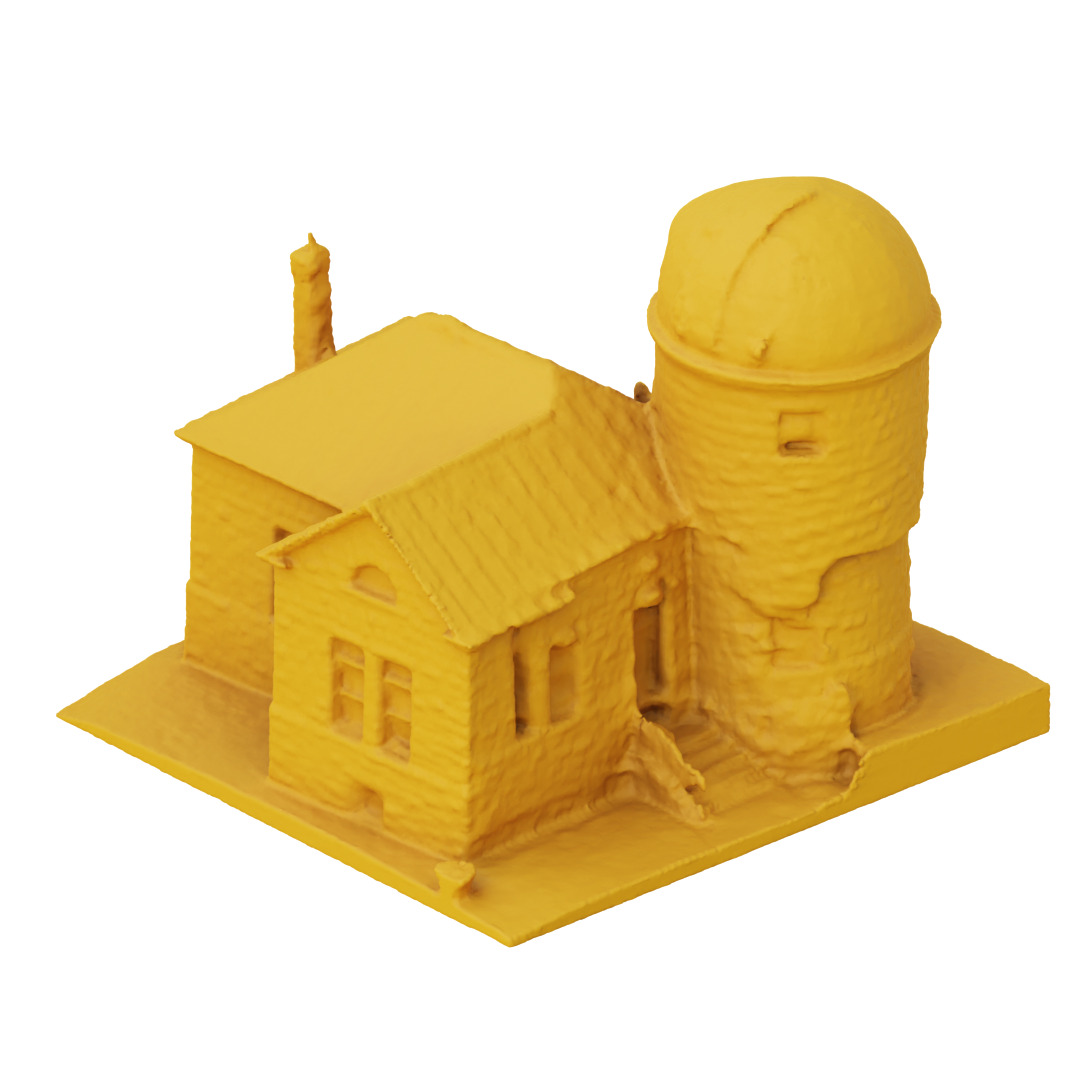}
    & \includegraphics[width=0.18\linewidth,keepaspectratio]{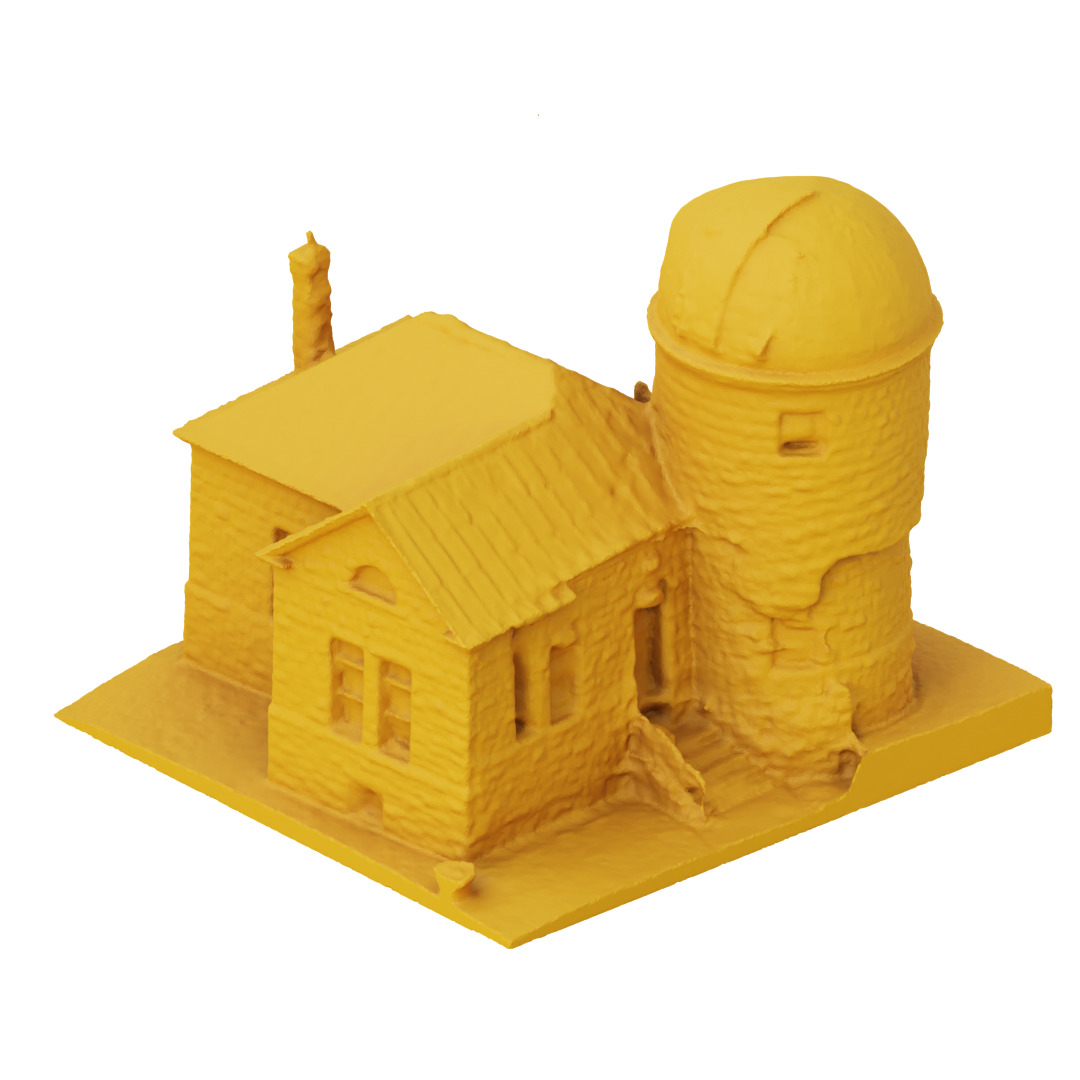} %
    \\
    \includegraphics[width=0.18\linewidth,keepaspectratio]{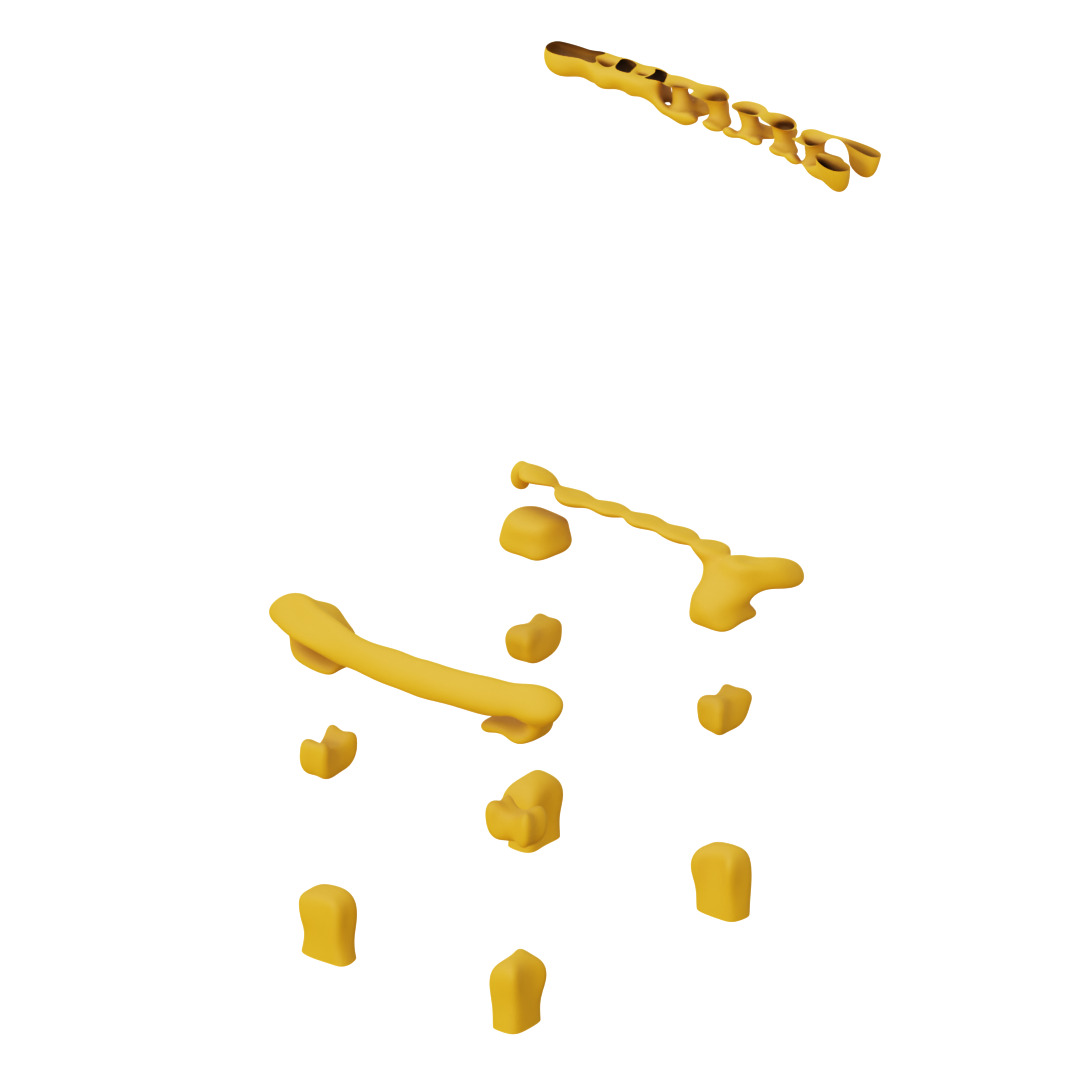}
    & \includegraphics[width=0.18\linewidth,keepaspectratio]{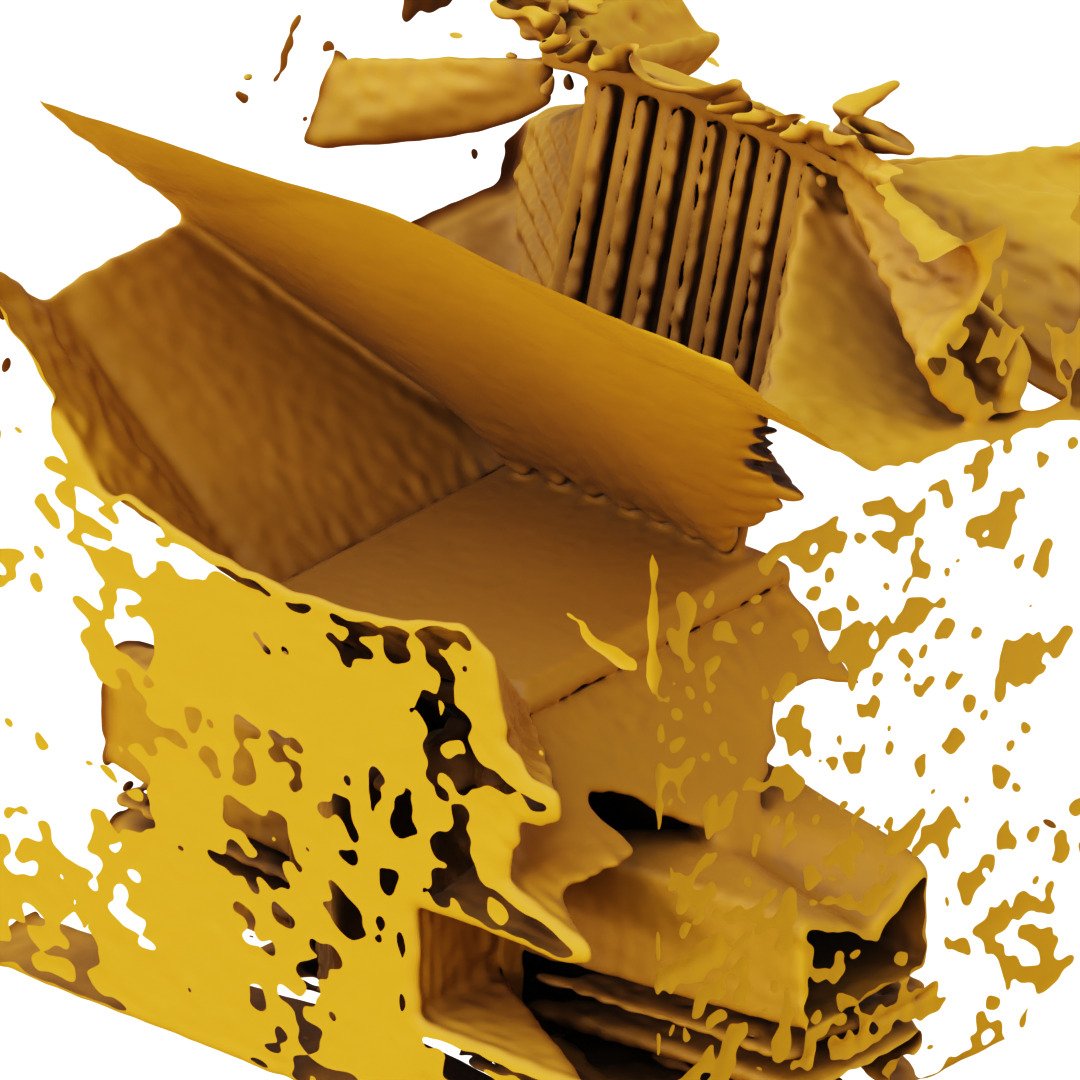}
    & \includegraphics[width=0.18\linewidth,keepaspectratio]{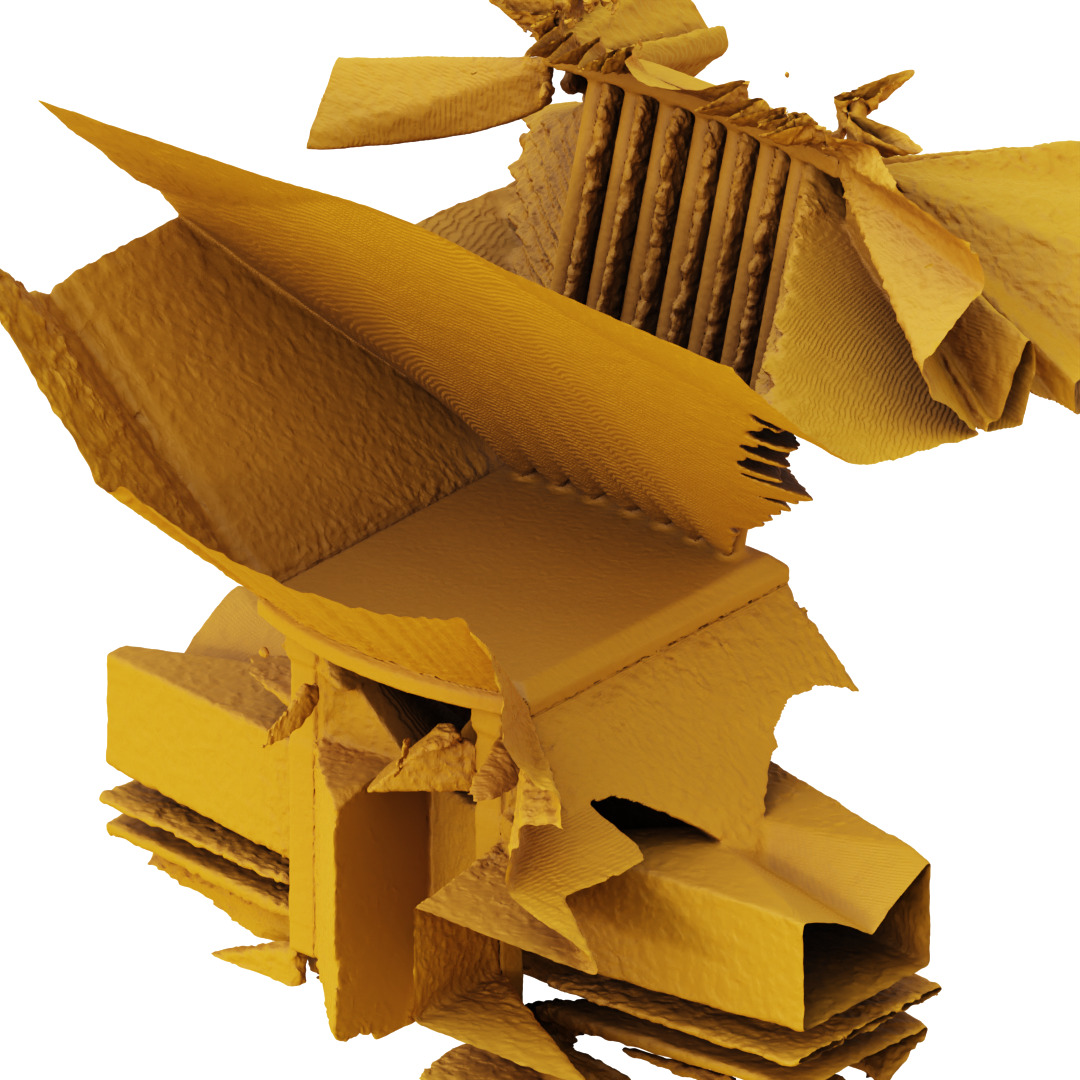}
    & \includegraphics[width=0.18\linewidth,keepaspectratio]{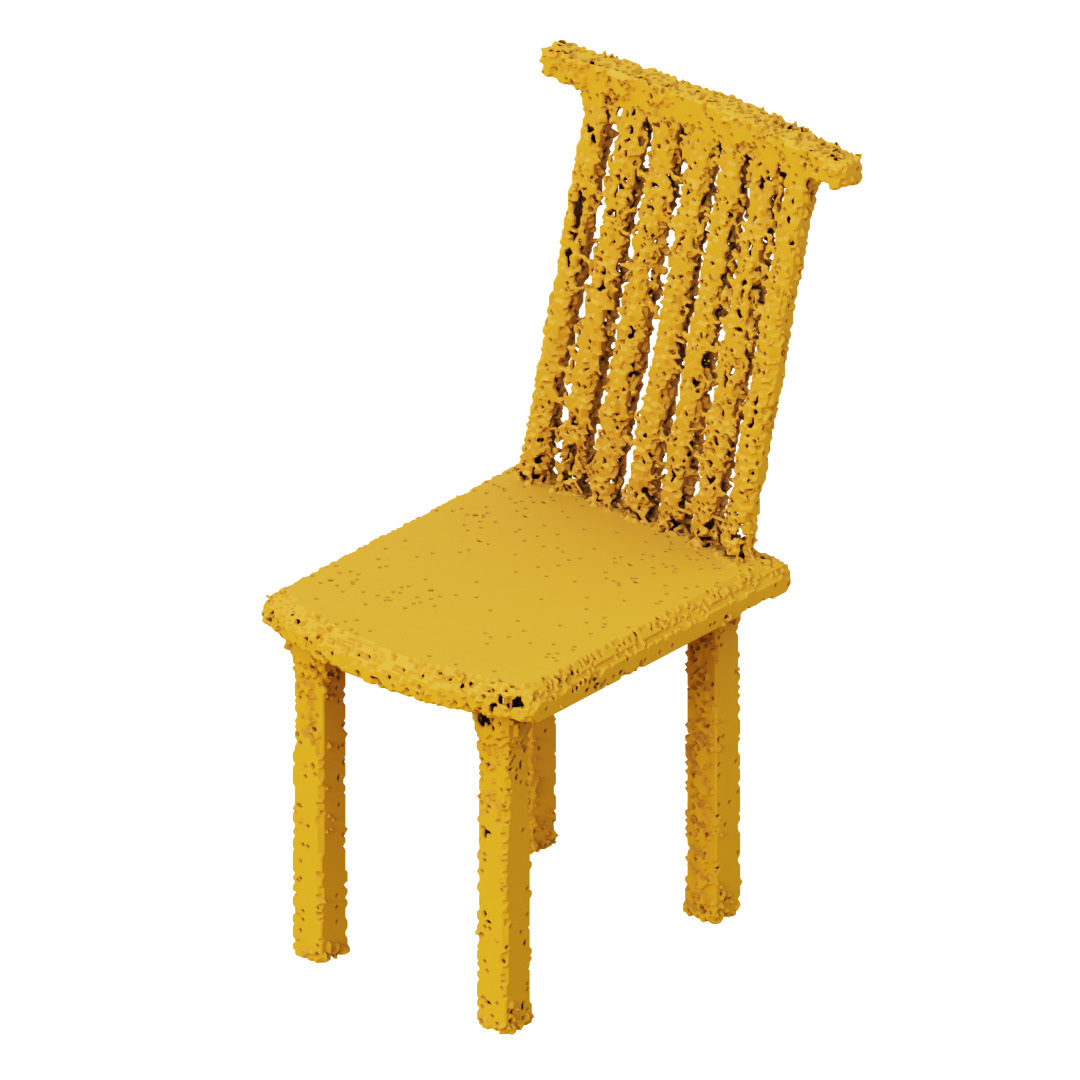}
    & \includegraphics[width=0.18\linewidth,keepaspectratio]{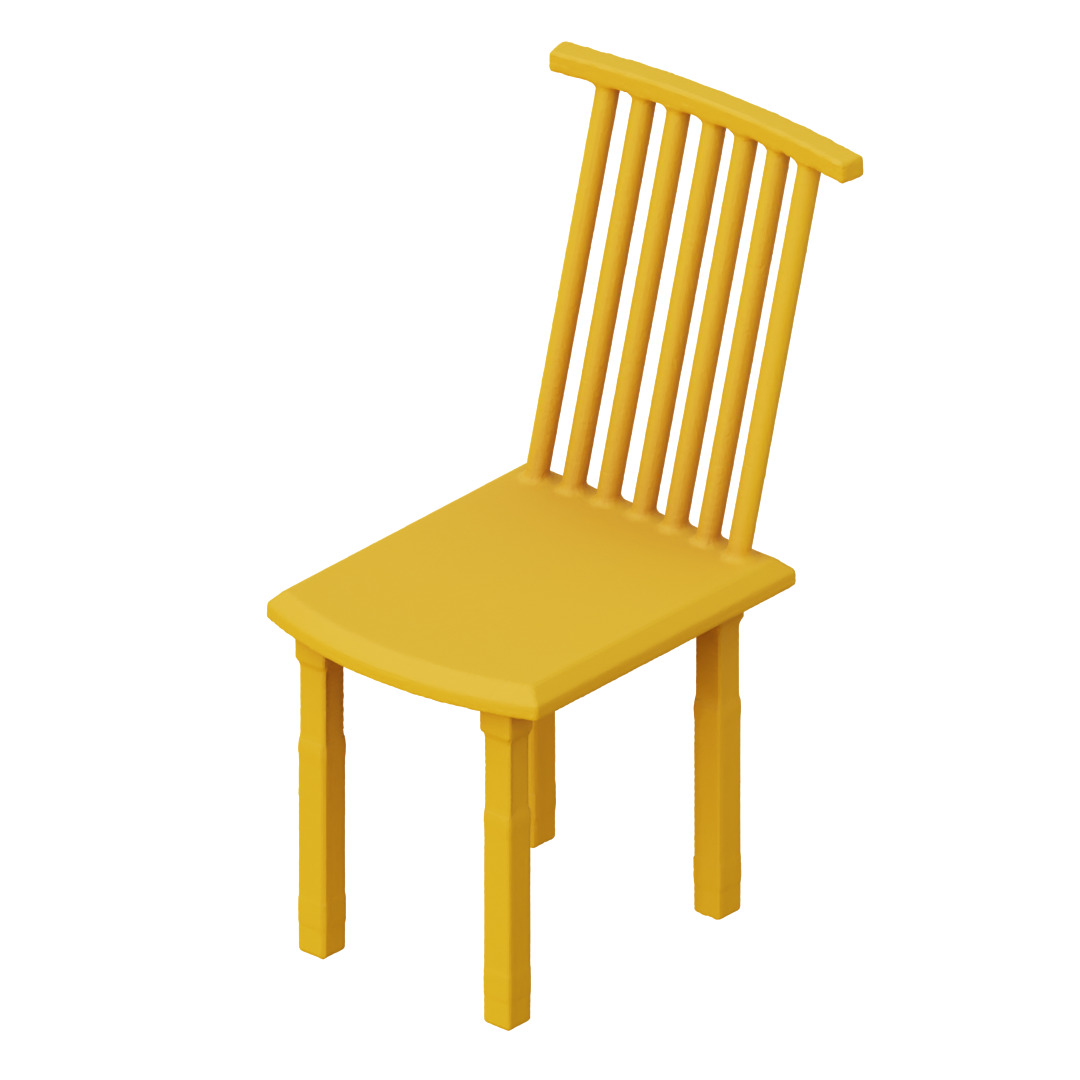}
    & \includegraphics[width=0.18\linewidth,keepaspectratio]{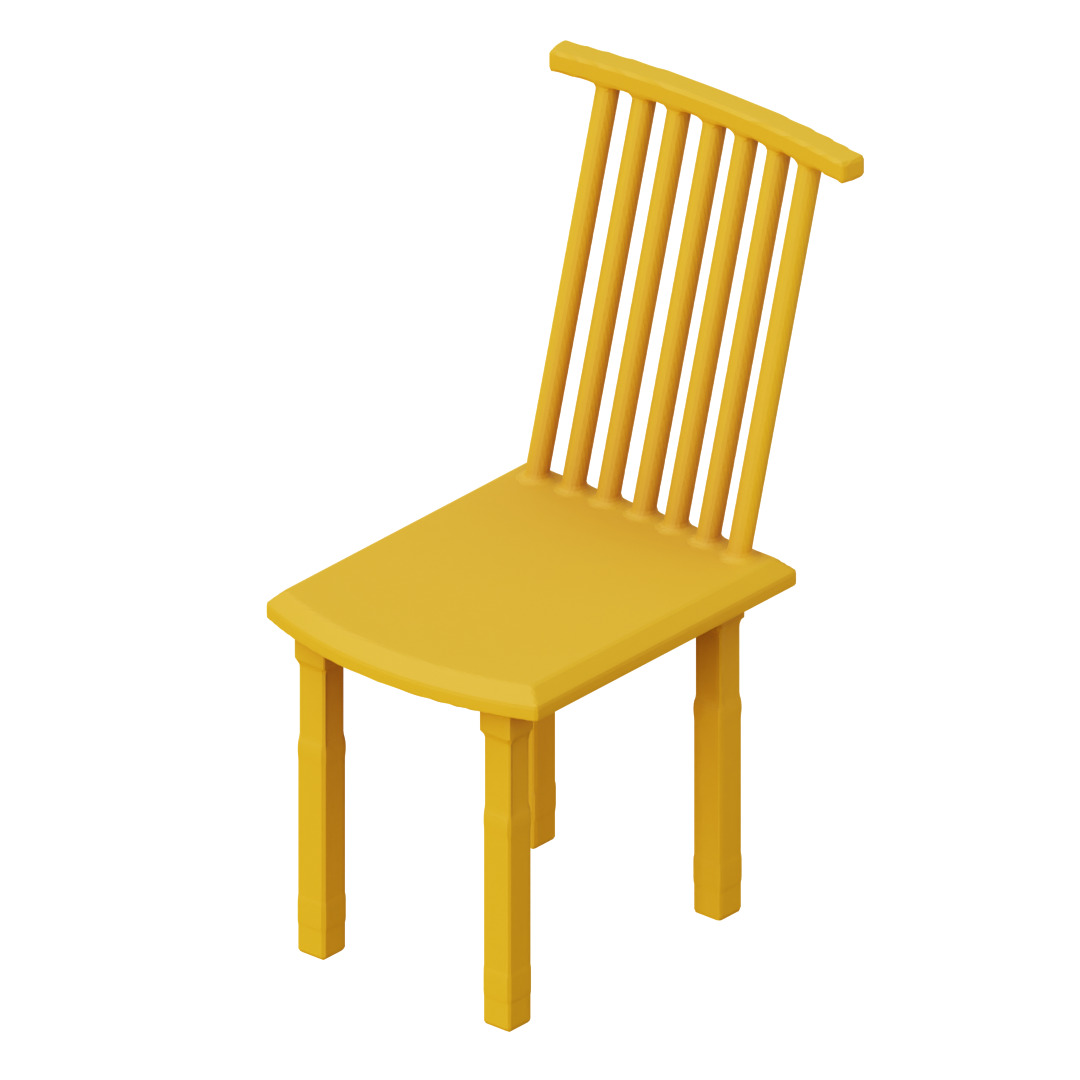}
    \\
    \textbf{SIREN}~\cite{sitzmann2020implicit} & \textbf{BACON}~\cite{lindell2022bacon} & \textbf{FF}~\cite{tancik2020fourier} & \textbf{NDF}~\cite{chibane2020ndf} & \textbf{Ours} & \textbf{GT}
    \end{NiceTabular}}
    \caption{\textbf{Baselines reconstruction.} This figure shows some examples of reconstructed meshes in the Thingi10k (above) and Shapenet (below) datasets for the baselines.}
    \label{fig:qual_comp}
\end{figure*}

\subsection{Regular vs. Oriented grids}\label{subsec:regular_oriented_experiments}

\Cref{tab:orientedvsregular} compares the performance of our method with regular grids on SDF and occupancy decoders. 
SDF and occupancies decoders are trained as explained in~\cref{subsec:implementation}. 
Normal regularization remains the same for both cases. Our method outperforms regular grids with an SDF decoder on all fronts, yielding smoother results on structured surfaces. Despite the underperformance of the occupancy framework, we observe fewer holes and dents (with the latter having a significant impact on the IoU), as shown in the supplementary material. These results show the adaptability of our method to different decoder output representations.

To open possible extensions of our work to large-scale scene representation, we show the render of our method on a scene from Matterport3D~\cite{chang2017matterport3d}. 
We divide the scene into $4 \times 4$ crops (with ground plane included) and train a model with an occupancy decoder for each crop. During inference, the mesh crops are rendered using marching cubes and finally fused to yield the scene, as shown in~\cref{fig:experiments_sdf_regular_regular}. 
Regular grids render a rougher and muddled 3D representation. Our method adapts well to thin surfaces and renders the scene with less roughness and sharper quality. Additional scenes are shown in the supplementary material.

As a consequence of the oriented grids, we observe that the proposed encoder renders planar surfaces more effectively in fewer training steps. Especially in more structural regular objects, the regular grids produce a caustic-like effect (surface noise). We observe a noise reduction in the surface reconstruction for the oriented grids, as seen in~\cref{fig:experiments_sdf_ours_regular} as soon as the first epoch. More results per epoch can be seen in the supplementary material.

\subsection{Baselines}\label{subsec:baselines}

\paperpar{Quantitative Results}
\Cref{tab:quat_results} details the experimental results of our method against the baselines.
We infer that grid-based methods outperform the baselines with significant improvement on all fronts. In a simple dataset composed of planar objects, like ABC, our encoder reconstructs smoother planar surfaces due to the alignment of the oriented grids. While rendering holistic details, most baselines often have over-smoothed surfaces. We also observe a higher IoU for our method due to fewer holes in our mesh and negligible splatting (many small mesh traces around the sampling region). Overall, the oriented grids produce robust 3D representations with higher fidelity across all datasets.

We also provide the number of parameters required for rendering the mesh. The advantage of multi-resolution grid representations is that the decoder size can be reduced to just an MLP with one hidden layer. This results in our approach getting meshes faster than other methods.

\paperpar{Qualitative Results}
Illustrative examples from Thingi10k and ShapeNet datasets are shown in~\cref{fig:qual_comp}. While BACON and FF can model the object with reasonable accuracy, it registers a lot of splattering, giving rise to unwanted noisy surfaces and artifacts. SIREN and BACON produce over-smoothed surfaces, losing intricate details on the mesh. NDF produces a lot of holes but manages to get compact meshes without splattering. BACON, SIREN, and FF collapse on the ShapeNet dataset. We note that we tried both watertight and not watertight versions of ShapeNet, but we obtained similar results for the failing baselines. We suspect these methods fail on ShapeNet due to their reliance on frequency-based encoding, which fails in high-frequency shape settings, \eg the small gaps in the chair.

\begin{table}[t]
    \caption{\textbf{Baselines} Experimental results for the oriented grid encoder against the baselines in three different datasets: ABC~\cite{koch2019abc}, Thingi10k \cite{Zhou2016Thingi10KAD}, and ShapeNet~\cite{shapenet2015}. The CD is multiplied by $10^{-5}$. The NC is multiplied by $10^{-4}$.}
    \label{tab:quat_results}
    \resizebox{1\linewidth}{!}{\setlength{\tabcolsep}{9.5pt}\begin{NiceTabular}{@{}lccccc}[colortbl-like] \toprule
     \rowcolor{Gray!40} & \thead{SIREN}~\cite{sitzmann2020implicit} & \thead{BACON}~\cite{lindell2022bacon} & \thead{FF}~\cite{tancik2020fourier} & \thead{NDF}~\cite{chibane2020ndf} & \thead{Ours} \\\midrule
    \rowcolor{Gray!10} & \multicolumn{6}{c}{\it Results for the ABC dataset \cite{koch2019abc}.} \\
    \rowcolor{Gray!10}    CD$\downarrow$ & 5.837 & 2.229 & 18.98 & 5.020 & \textbf{0.603} \\
    \rowcolor{Gray!10}    NC$\downarrow$ & 5.150 & 4.658 & 5.170 & 4.732 & \textbf{3.987} \\
    \rowcolor{Gray!10}    IoU $\uparrow$ & 0.879 & 0.987 & 0.916 & 0.950 & \textbf{0.998} \\\midrule
         & \multicolumn{6}{c}{\it Results for the Thingi10k dataset \cite{Zhou2016Thingi10KAD}.} \\
        CD$\downarrow$ & 62.85 & 60.72 & 67.93 & 4.421 & \textbf{0.608} \\
        NC$\downarrow$ & 49.11 & 5.052 & 58.31 & 4.636 & \textbf{4.413} \\
        IoU $\uparrow$ & 0.716 & 0.928 & 0.877 & 0.920 & \textbf{0.995}\\\midrule
        \rowcolor{Gray!10} & \multicolumn{6}{c}{\it Results for the ShapeNet dataset \cite{shapenet2015}.} \\
        \rowcolor{Gray!10} CD$\downarrow$ & -- & -- & -- & 4.125 & \textbf{0.392} \\
        \rowcolor{Gray!10} NC$\downarrow$ & -- & -- & -- & 6.186 & \textbf{5.425} \\
        \rowcolor{Gray!10} IoU $\uparrow$ & -- & -- & -- & 0.984 & \textbf{0.999} \\\midrule
        & \multicolumn{6}{c}{\it Number of Inference Parameters} \\
        \# Params. & 199K & 537K & 527K & 4.62M & 388K \\\bottomrule
        
    \end{NiceTabular}}
\end{table}

\paperpar{Limitations} Small holes can arise from non-watertight planar surfaces that affect both oriented and regular grids. The oriented grid, however, fills holes more adequately than the regular counterpart. 
A more general multi-resolution grid representation issue is the difficulty of modeling thin surfaces. Despite the limitation, our method substantially improves from the regular grid. We provide meshes for such cases in the supplementary material.
We hope to study and mitigate the above limitations in subsequent works.

    \section{Conclusion}\label{sec:conclusion}

This paper proposes a novel approach for a 3D grid-based encoder for 3D representation. The encoder considers the inherent structural regularities in objects by aligning the grids with the object surface normal and aggregating the cell features from a newly developed cylindrical interpolation technique and local aggregation scheme that mitigates the issues caused by the alignment. Oriented grids yielded state-of-the-art results while being more robust and accurate to decoder representation changes, answering the paper's research question. 
Future work lies in extending the work for neural radiance fields and scene and object reconstruction where the object implicit representation is unknown.

\vspace{.5cm}
\paperpar{Acknowledgements}
Gon\c{c}alo was partially supported by the LARSyS funding (DOI: 10.54499/LA/P/0083/2020, 10.54499/UIDP/50009/2020, and 10.54499/UIDB/50009/2020) and grant PD/BD/150630/2020, from the Portuguese ``Funda\c{c}\~{a}o para a Ci\^{e}ncia e a Tecnologia''.

    \clearpage
    {\small
    \bibliographystyle{ieee_fullname}
    \bibliography{egbib}

\begin{thebibliography}{10}\itemsep=-1pt

\bibitem{amanatides1984ray}
John Amanatides.
\newblock Ray tracing with cones.
\newblock {\em SIGGRAPH}, 18(3):129--135, 1984.

\bibitem{attal2020matryodshka}
Benjamin Attal, Selena Ling, Aaron Gokaslan, Christian Richardt, and James
  Tompkin.
\newblock Matryodshka: Real-time 6dof video view synthesis using multi-sphere
  images.
\newblock In {\em European Conf. Computer Vision (ECCV)}, pages 441--459, 2020.

\bibitem{bernardini1999ball}
Fausto Bernardini, Joshua Mittleman, Holly Rushmeier, Cl{\'a}udio Silva, and
  Gabriel Taubin.
\newblock The ball-pivoting algorithm for surface reconstruction.
\newblock {\em IEEE Trans. on Visualization and Computer Graphics},
  5(4):349--359, 1999.

\bibitem{boulch2022poco}
Alexandre Boulch and Renaud Marlet.
\newblock Poco: Point convolution for surface reconstruction.
\newblock In {\em IEEE/CVF Conf. Computer Vision and Pattern Recognition
  (CVPR)}, pages 6302--6314, 2022.

\bibitem{broxton2020immersive}
Michael Broxton, John Flynn, Ryan Overbeck, Daniel Erickson, Peter Hedman,
  Matthew Duvall, Jason Dourgarian, Jay Busch, Matt Whalen, and Paul Debevec.
\newblock Immersive light field video with a layered mesh representation.
\newblock {\em ACM Transactions on Graphics (TOG)}, 39(4):86--1, 2020.

\bibitem{carr2001reconstruction}
Jonathan~C Carr, Richard~K Beatson, Jon~B Cherrie, Tim~J Mitchell, W~Richard
  Fright, Bruce~C McCallum, and Tim~R Evans.
\newblock Reconstruction and representation of 3d objects with radial basis
  functions.
\newblock In {\em ACM on Computer Graphics and Interactive Techniques}, pages
  67--76, 2001.

\bibitem{chabra2020deep}
Rohan Chabra, Jan~E Lenssen, Eddy Ilg, Tanner Schmidt, Julian Straub, Steven
  Lovegrove, and Richard Newcombe.
\newblock Deep local shapes: Learning local sdf priors for detailed 3d
  reconstruction.
\newblock In {\em European Conf. Computer Vision (ECCV)}, pages 608--625.
  Springer, 2020.

\bibitem{chang2017matterport3d}
Angel Chang, Angela Dai, Thomas Funkhouser, Maciej Halber, Matthias Niebner,
  Manolis Savva, Shuran Song, Andy Zeng, and Yinda Zhang.
\newblock Matterport3d: Learning from rgb-d data in indoor environments.
\newblock In {\em Int'l Conf. 3D Vision (3DV)}, pages 667--676, 2017.

\bibitem{shapenet2015}
Angel~X Chang, Thomas Funkhouser, Leonidas Guibas, Pat Hanrahan, Qixing Huang,
  Zimo Li, Silvio Savarese, Manolis Savva, Shuran Song, Hao Su, et~al.
\newblock Shapenet: An information-rich 3d model repository.
\newblock \url{https://shapenet.org/}, 2015.

\bibitem{chen2022tensorf}
Anpei Chen, Zexiang Xu, Andreas Geiger, Jingyi Yu, and Hao Su.
\newblock Tensorf: Tensorial radiance fields.
\newblock In {\em European Conf. Computer Vision (ECCV)}, pages 333--350, 2022.

\bibitem{chen2019learning}
Zhiqin Chen and Hao Zhang.
\newblock Learning implicit fields for generative shape modeling.
\newblock In {\em IEEE/CVF Conf. Computer Vision and Pattern Recognition
  (CVPR)}, pages 5939--5948, 2019.

\bibitem{chibane20ifnet}
Julian Chibane, Thiemo Alldieck, and Gerard Pons-Moll.
\newblock Implicit functions in feature space for 3d shape reconstruction and
  completion.
\newblock In {\em IEEE/CVF Conf. Computer Vision and Pattern Recognition
  (CVPR)}, pages 6970--6981, 2020.

\bibitem{chibane2020ndf}
Julian Chibane, Aymen Mir, and Gerard Pons-Moll.
\newblock Neural unsigned distance fields for implicit function learning.
\newblock In {\em Advances in Neural Information Processing Systems (NeurIPS)},
  2020.

\bibitem{chien1986volume}
Chiun-Hong Chien and Jake~K Aggarwal.
\newblock Volume/surface octrees for the representation of three-dimensional
  objects.
\newblock {\em Computer Vision, Graphics, and Image Processing},
  36(1):100--113, 1986.

\bibitem{dai2020neural}
Peng Dai, Yinda Zhang, Zhuwen Li, Shuaicheng Liu, and Bing Zeng.
\newblock Neural point cloud rendering via multi-plane projection.
\newblock In {\em IEEE/CVF Conf. Computer Vision and Pattern Recognition
  (CVPR)}, pages 7830--7839, 2020.

\bibitem{duan2020curriculum}
Yueqi Duan, Haidong Zhu, He Wang, Li Yi, Ram Nevatia, and Leonidas~J Guibas.
\newblock Curriculum deepsdf.
\newblock In {\em European Conf. Computer Vision (ECCV)}, pages 51--67, 2020.

\bibitem{fridovich2022plenoxels}
Sara Fridovich-Keil, Alex Yu, Matthew Tancik, Qinhong Chen, Benjamin Recht, and
  Angjoo Kanazawa.
\newblock Plenoxels: Radiance fields without neural networks.
\newblock In {\em IEEE/CVF Conf. Computer Vision and Pattern Recognition
  (CVPR)}, pages 5501--5510, 2022.

\bibitem{groueix2018papier}
Thibault Groueix, Matthew Fisher, Vladimir~G Kim, Bryan~C Russell, and Mathieu
  Aubry.
\newblock A papier-m{\^a}ch{\'e} approach to learning 3d surface generation.
\newblock In {\em Proceedings of the IEEE conference on computer vision and
  pattern recognition}, pages 216--224, 2018.

\bibitem{guillard2022udf}
Benoit Guillard, Federico Stella, and Pascal Fua.
\newblock Meshudf: Fast and differentiable meshing of unsigned distance field
  networks.
\newblock In {\em European Conf. Computer Vision (ECCV)}, 2022.

\bibitem{hane2017hierarchical}
Christian H{\"a}ne, Shubham Tulsiani, and Jitendra Malik.
\newblock Hierarchical surface prediction for 3d object reconstruction.
\newblock In {\em Int'l Conf. 3D Vision (3DV)}, pages 412--420, 2017.

\bibitem{hart1996sphere}
John~C Hart.
\newblock Sphere tracing: A geometric method for the antialiased ray tracing of
  implicit surfaces.
\newblock {\em The Visual Computer}, 12(10):527--545, 1996.

\bibitem{hoppe1992surface}
Hugues Hoppe, Tony DeRose, Tom Duchamp, John McDonald, and Werner Stuetzle.
\newblock Surface reconstruction from unorganized points.
\newblock In {\em ACM on Computer Graphics and Interactive Techniques}, pages
  71--78, 1992.

\bibitem{huang2023neural}
Jiahui Huang, Zan Gojcic, Matan Atzmon, Or Litany, Sanja Fidler, and Francis
  Williams.
\newblock Neural kernel surface reconstruction.
\newblock In {\em IEEE/CVF Conf. Computer Vision and Pattern Recognition
  (CVPR)}, pages 4369--4379, 2023.

\bibitem{jiang2020local}
Chiyu Jiang, Avneesh Sud, Ameesh Makadia, Jingwei Huang, Matthias Nie{\ss}ner,
  Thomas Funkhouser, et~al.
\newblock Local implicit grid representations for 3d scenes.
\newblock In {\em IEEE/CVF Conf. Computer Vision and Pattern Recognition
  (CVPR)}, pages 6001--6010, 2020.

\bibitem{diederik2015adam}
Diederik~P. Kingma and Jimmy Ba.
\newblock Adam: {A} method for stochastic optimization.
\newblock In {\em Int'l Conf. Learning Representations (ICLR)}, 2015.

\bibitem{koch2019abc}
Sebastian Koch, Albert Matveev, Zhongshi Jiang, Francis Williams, Alexey
  Artemov, Evgeny Burnaev, Marc Alexa, Denis Zorin, and Daniele Panozzo.
\newblock Abc: A big cad model dataset for geometric deep learning.
\newblock In {\em IEEE/CVF Conf. Computer Vision and Pattern Recognition
  (CVPR)}, pages 9601--9611, 2019.

\bibitem{laine2010efficient}
Samuli Laine and Tero Karras.
\newblock Efficient sparse voxel octrees--analysis, extensions, and
  implementation.
\newblock {\em NVIDIA Corporation}, 2(6), 2010.

\bibitem{lindell2022bacon}
David~B Lindell, Dave Van~Veen, Jeong~Joon Park, and Gordon Wetzstein.
\newblock Bacon: Band-limited coordinate networks for multiscale scene
  representation.
\newblock In {\em IEEE/CVF Conf. Computer Vision and Pattern Recognition
  (CVPR)}, pages 16252--16262, 2022.

\bibitem{lionar2021dynamic}
Stefan Lionar, Daniil Emtsev, Dusan Svilarkovic, and Songyou Peng.
\newblock Dynamic plane convolutional occupancy networks.
\newblock In {\em IEEE/CVF Winter Conference on Applications of Computer
  Vision}, pages 1829--1838, 2021.

\bibitem{liu2022learning}
Hsueh-Ti~Derek Liu, Francis Williams, Alec Jacobson, Sanja Fidler, and Or
  Litany.
\newblock Learning smooth neural functions via lipschitz regularization.
\newblock In {\em SIGGRAPH}, 2022.

\bibitem{liu2020neural}
Lingjie Liu, Jiatao Gu, Kyaw Zaw~Lin, Tat-Seng Chua, and Christian Theobalt.
\newblock Neural sparse voxel fields.
\newblock {\em Advances in Neural Information Processing Systems (NeurIPS)},
  33:15651--15663, 2020.

\bibitem{lombardi2019neural}
Stephen Lombardi, Tomas Simon, Jason Saragih, Gabriel Schwartz, Andreas
  Lehrmann, and Yaser Sheikh.
\newblock Neural volumes: learning dynamic renderable volumes from images.
\newblock {\em ACM Transactions on Graphics (TOG)}, 38(4):1--14, 2019.

\bibitem{long2023neuraludf}
Xiaoxiao Long, Cheng Lin, Lingjie Liu, Yuan Liu, Peng Wang, Christian Theobalt,
  Taku Komura, and Wenping Wang.
\newblock Neuraludf: Learning unsigned distance fields for multi-view
  reconstruction of surfaces with arbitrary topologies.
\newblock In {\em IEEE/CVF Conf. Computer Vision and Pattern Recognition
  (CVPR)}, 2023.

\bibitem{lorensen1987marching}
William~E Lorensen and Harvey~E Cline.
\newblock Marching cubes: A high resolution 3d surface construction algorithm.
\newblock {\em SIGGRAPH}, 21(4):163--169, 1987.

\bibitem{martel2021acorn}
Julien~NP Martel, David~B Lindell, Connor~Z Lin, Eric~R Chan, Marco Monteiro,
  and Gordon Wetzstein.
\newblock Acorn: adaptive coordinate networks for neural scene representation.
\newblock {\em ACM Transactions on Graphics (TOG)}, 40(4):1--13, 2021.

\bibitem{mescheder2019occupancy}
Lars Mescheder, Michael Oechsle, Michael Niemeyer, Sebastian Nowozin, and
  Andreas Geiger.
\newblock Occupancy networks: Learning 3d reconstruction in function space.
\newblock In {\em IEEE/CVF Conf. Computer Vision and Pattern Recognition
  (CVPR)}, pages 4460--4470, 2019.

\bibitem{mildenhall2020nerf}
Ben Mildenhall, Pratul~P. Srinivasan, Matthew Tancik, Jonathan~T. Barron, Ravi
  Ramamoorthi, and Ren Ng.
\newblock Nerf: Representing scenes as neural radiance fields for view
  synthesis.
\newblock In {\em European Conf. Computer Vision (ECCV)}, 2020.

\bibitem{muller2022instant}
Thomas M{\"u}ller, Alex Evans, Christoph Schied, and Alexander Keller.
\newblock Instant neural graphics primitives with a multiresolution hash
  encoding.
\newblock {\em SIGGRAPH}, 2022.

\bibitem{niemeyer2020differentiable}
Michael Niemeyer, Lars Mescheder, Michael Oechsle, and Andreas Geiger.
\newblock Differentiable volumetric rendering: Learning implicit 3d
  representations without 3d supervision.
\newblock In {\em IEEE/CVF Conf. Computer Vision and Pattern Recognition
  (CVPR)}, pages 3504--3515, 2020.

\bibitem{Oechsle2021ICCV}
Michael Oechsle, Songyou Peng, and Andreas Geiger.
\newblock Unisurf: Unifying neural implicit surfaces and radiance fields for
  multi-view reconstruction.
\newblock In {\em IEEE/CVF Int'l Conf. Computer Vision (ICCV)}, 2021.

\bibitem{park2019deepsdf}
Jeong~Joon Park, Peter Florence, Julian Straub, Richard Newcombe, and Steven
  Lovegrove.
\newblock Deepsdf: Learning continuous signed distance functions for shape
  representation.
\newblock In {\em IEEE/CVF Conf. Computer Vision and Pattern Recognition
  (CVPR)}, pages 165--174, 2019.

\bibitem{paszke2017automatic}
Adam Paszke, Sam Gross, Soumith Chintala, Gregory Chanan, Edward Yang, Zachary
  DeVito, Zeming Lin, Alban Desmaison, Luca Antiga, and Adam Lerer.
\newblock Automatic differentiation in pytorch.
\newblock In {\em Advances in Neural Information Processing Systems - Workshop
  (NeurIPS-W)}, 2017.

\bibitem{peng2021shape}
Songyou Peng, Chiyu Jiang, Yiyi Liao, Michael Niemeyer, Marc Pollefeys, and
  Andreas Geiger.
\newblock Shape as points: A differentiable poisson solver.
\newblock {\em Advances in Neural Information Processing Systems (NeurIPS)},
  34:13032--13044, 2021.

\bibitem{peng2020convolutional}
Songyou Peng, Michael Niemeyer, Lars Mescheder, Marc Pollefeys, and Andreas
  Geiger.
\newblock Convolutional occupancy networks.
\newblock In {\em European Conf. Computer Vision (ECCV)}, pages 523--540, 2020.

\bibitem{qi2017pointnet}
Charles~R Qi, Hao Su, Kaichun Mo, and Leonidas~J Guibas.
\newblock Pointnet: Deep learning on point sets for 3d classification and
  segmentation.
\newblock In {\em IEEE Conf. Computer Vision and Pattern Recognition (CVPR)},
  pages 652--660, 2017.

\bibitem{qi2017pointnet++}
Charles~Ruizhongtai Qi, Li Yi, Hao Su, and Leonidas~J Guibas.
\newblock Pointnet++: Deep hierarchical feature learning on point sets in a
  metric space.
\newblock {\em Advances in Neural Information Processing Systems (NIPS)}, 30,
  2017.

\bibitem{riegler2020free}
Gernot Riegler and Vladlen Koltun.
\newblock Free view synthesis.
\newblock In {\em European Conf. Computer Vision (ECCV)}, pages 623--640.
  Springer, 2020.

\bibitem{riegler2017octnet}
Gernot Riegler, Ali Osman~Ulusoy, and Andreas Geiger.
\newblock Octnet: Learning deep 3d representations at high resolutions.
\newblock In {\em IEEE Conf. Computer Vision and Pattern Recognition (CVPR)},
  pages 3577--3586, 2017.

\bibitem{russell2016artificial}
Stuart~J. Russell and Peter Norvig.
\newblock {\em Artificial Intelligence: a modern approach}.
\newblock Pearson, 3 edition, 2009.

\bibitem{sitzmann2020implicit}
Vincent Sitzmann, Julien Martel, Alexander Bergman, David Lindell, and Gordon
  Wetzstein.
\newblock Implicit neural representations with periodic activation functions.
\newblock {\em Advances in Neural Information Processing Systems (NeurIPS)},
  33:7462--7473, 2020.

\bibitem{sun2022direct}
Cheng Sun, Min Sun, and Hwann-Tzong Chen.
\newblock Direct voxel grid optimization: Super-fast convergence for radiance
  fields reconstruction.
\newblock In {\em IEEE/CVF Conf. Computer Vision and Pattern Recognition
  (CVPR)}, pages 5459--5469, 2022.

\bibitem{takikawa2021neural}
Towaki Takikawa, Joey Litalien, Kangxue Yin, Karsten Kreis, Charles Loop, Derek
  Nowrouzezahrai, Alec Jacobson, Morgan McGuire, and Sanja Fidler.
\newblock Neural geometric level of detail: Real-time rendering with implicit
  3d shapes.
\newblock In {\em IEEE/CVF Conf. Computer Vision and Pattern Recognition
  (CVPR)}, pages 11358--11367, 2021.

\bibitem{tancik2020fourier}
Matthew Tancik, Pratul Srinivasan, Ben Mildenhall, Sara Fridovich-Keil, Nithin
  Raghavan, Utkarsh Singhal, Ravi Ramamoorthi, Jonathan Barron, and Ren Ng.
\newblock Fourier features let networks learn high frequency functions in low
  dimensional domains.
\newblock {\em Advances in Neural Information Processing Systems (NeurIPS)},
  33:7537--7547, 2020.

\bibitem{tang2021sign}
Jiapeng Tang, Jiabao Lei, Dan Xu, Feiying Ma, Kui Jia, and Lei Zhang.
\newblock Sa-convonet: Sign-agnostic optimization of convolutional occupancy
  networks.
\newblock In {\em IEEE/CVF Int'l Conf. Computer Vision (ICCV)}, 2021.

\bibitem{wang2021neus}
Peng Wang, Lingjie Liu, Yuan Liu, Christian Theobalt, Taku Komura, and Wenping
  Wang.
\newblock Neus: Learning neural implicit surfaces by volume rendering for
  multi-view reconstruction.
\newblock {\em Advances in Neural Information Processing Systems (NeurIPS)},
  34:27171--27183, 2021.

\bibitem{wang2017cnn}
Peng-Shuai Wang, Yang Liu, Yu-Xiao Guo, Chun-Yu Sun, and Xin Tong.
\newblock O-cnn: Octree-based convolutional neural networks for 3d shape
  analysis.
\newblock {\em ACM Transactions on Graphics (TOG)}, 36(4):1--11, 2017.

\bibitem{wang2020deep}
Peng-Shuai Wang, Yang Liu, and Xin Tong.
\newblock Deep octree-based cnns with output-guided skip connections for 3d
  shape and scene completion.
\newblock In {\em IEEE/CVF Conf. on Computer Vision and Pattern Recognition
  Workshops (CVPRW)}, pages 266--267, 2020.

\bibitem{wang2018adaptive}
Peng-Shuai Wang, Chun-Yu Sun, Yang Liu, and Xin Tong.
\newblock Adaptive o-cnn: A patch-based deep representation of 3d shapes.
\newblock {\em ACM Transactions on Graphics (TOG)}, 37(6):1--11, 2018.

\bibitem{wang2021unsupervised}
Peng-Shuai Wang, Yu-Qi Yang, Qian-Fang Zou, Zhirong Wu, Yang Liu, and Xin Tong.
\newblock Unsupervised 3d learning for shape analysis via multiresolution
  instance discrimination.
\newblock In {\em AAAI Conference on Artificial Intelligence}, pages
  2773--2781, 2021.

\bibitem{point-cloud-utils}
Francis Williams.
\newblock Point cloud utils, 2022.
\newblock https://www.github.com/fwilliams/point-cloud-utils.

\bibitem{williams2022neural}
Francis Williams, Zan Gojcic, Sameh Khamis, Denis Zorin, Joan Bruna, Sanja
  Fidler, and Or Litany.
\newblock Neural fields as learnable kernels for 3d reconstruction.
\newblock In {\em IEEE/CVF Conf. Computer Vision and Pattern Recognition
  (CVPR)}, pages 18500--18510, 2022.

\bibitem{yifan2019differentiable}
Wang Yifan, Felice Serena, Shihao Wu, Cengiz {\"O}ztireli, and Olga
  Sorkine-Hornung.
\newblock Differentiable surface splatting for point-based geometry processing.
\newblock {\em ACM Transactions on Graphics (TOG)}, 38(6):1--14, 2019.

\bibitem{yifan2021iso}
Wang Yifan, Shihao Wu, Cengiz Oztireli, and Olga Sorkine-Hornung.
\newblock Iso-points: Optimizing neural implicit surfaces with hybrid
  representations.
\newblock In {\em IEEE/CVF Conf. Computer Vision and Pattern Recognition
  (CVPR)}, pages 374--383, 2021.

\bibitem{yu2021plenoctrees}
Alex Yu, Ruilong Li, Matthew Tancik, Hao Li, Ren Ng, and Angjoo Kanazawa.
\newblock Plenoctrees for real-time rendering of neural radiance fields.
\newblock In {\em IEEE/CVF Int'l Conf. Computer Vision (ICCV)}, pages
  5752--5761, 2021.

\bibitem{Zhou2016Thingi10KAD}
Qingnan Zhou and Alec Jacobson.
\newblock Thingi10k: A dataset of 10, 000 3d-printing models.
\newblock \url{https://ten-thousand-models.appspot.com/}, 2016.

\end{thebibliography}
    }
}{}

\ifthenelse{\boolean{suppmat}}{

    \appendix
    \renewcommand\thefigure{\thesection.\arabic{figure}}    
    \renewcommand\thetable{\thesection.\arabic{table}}    

    \title{Oriented-grid Encoder for 3D Implicit Representations\\{\large \sc Supplementary materials}}

    \author{Arihant Gaur\textsuperscript{1}\footnotemark[1] \hspace{50pt} G. Dias Pais\textsuperscript{1,2}\footnotemark[1] \hspace{50pt}  Pedro Miraldo\textsuperscript{1}\\[5pt]
    \begin{minipage}{0.5\textwidth}
        \centering
        \textsuperscript{1}Mitsubishi Electric Research Labs (MERL)\\
    \end{minipage}\hfill
    \begin{minipage}{0.47\textwidth}
        \centering
        \textsuperscript{2}Instituto Superior T\'ecnico, Lisboa\\
    \end{minipage}
    }

    \maketitle

    {\it \noindent
    These supplementary materials present additional ablations (\cref{app:ablations}) and experiments against regular grids and baselines (\cref{app:experiments}). The code will be made available upon acceptance.
    }
    \tableofcontents

\section{Cylindrical interpolation}
This section shows how the cylindrical interpolation mitigates the issue caused by the alignment of the oriented cell to the $z$--axis. While searching through the Orientation tree, since no other constraints are set, consecutive cells can have different rotations pointing aligned to the $z$--axis but with different $x$ and $y$-axis rotations since the tree estimation is up to a rotation. Therefore, trilinear interpolation ($f_{tri}(\cdot)$) is not well-suited to our oriented cells. This issue is mitigated by the proposed cylindrical interpolation ($f_{cyl}(\cdot)$), where the same point has the same feature, no matter what the rotation around the $z$--axis, as shown in~\cref{fig:comp_interp}.

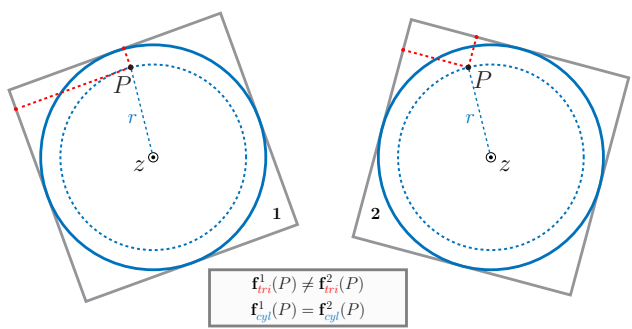
\begin{figure}[b]
    \resizebox{1\linewidth}{!}{\begin{tikzpicture}

    \def\s{8cm}
    \def\ss{4cm}
    \def\angX{0.6}
    \def\angY{1}
    \def\arrowSize{1cm}
    \def\offset{12cm}
    \def\ptx{\ss*0.8}
    \def\pty{\ss*1.8}
    \def\rotationOne{20}
    \def\rotationTwo{20}
    
    \draw[name path =rectangle, line width= 1.0mm, \colorSqOne, rotate around={\rotationOne:(\ss, \ss)}] (0,0) rectangle (\s,\s) node (SqOne) {};
    
    \node[circle, draw, \colorAxis] (az) at (\ss,\ss){};
    \fill [\colorAxis] (\ss,\ss) circle [x radius=0.7mm, y radius=0.7mm];
    \node[\colorAxis] (axisz) at (\ss - 5mm,\ss - 3mm) [inner sep=0pt, font=\Huge]{$z$};
    
    \coordinate (pt) at (\ptx, \pty);
    \node at (pt) [below = 2mm of pt, xshift = -3mm, font=\Huge, \colorSurface] {$P$};
    
    \draw [\colorCircle, line width= 1.0mm] (\ss,\ss) circle [x radius=\ss, y radius=\ss];
    
    \draw [\colorCircle, line width= 0.7mm, dashed] (\ss,\ss) circle [x radius=\ss*0.825, y radius=\ss*0.825];
    
    \draw [name path=radi, line width= 0.5mm, dashed, \colorCircle, font=\huge] (az) to [edge label=$r$] (pt);
    
    \begin{scope}[rotate around={\rotationOne:(\ptx, \pty)}]
        \path [name path=thrAy] (\ptx, \pty) -- ($ (\ptx, \ss) ! 1.3 ! (\ptx, \pty) $);
        \path [name path=thrAx, red] (\ptx, \pty) -- ($ (\ptx, \pty) ! 1.4 ! (0, \pty) $);
    
        \fill [red, name intersections={of=thrAy and rectangle}]
        (intersection-1) circle (2pt) node (ix) {};
    
        \fill [red, name intersections={of=thrAx and rectangle}]
        (intersection-1) circle (2pt) node (iy) {};
    
        \draw [red, line width= 0.7mm, dashed] (pt) -- (ix);
        \draw [red, line width= 0.7mm, dashed] (pt) -- (iy);
    \end{scope}
    
    \fill [point, \colorSurface] (pt) circle [x radius=1mm, y radius=1mm];
    
    \def\rotationTwo{-15}
    \begin{scope}[xshift = \offset]
        \draw[name path =rectangle, line width= 1.0mm, \colorSqOne, rotate around={\rotationTwo:(\ss, \ss)}] (0,0) rectangle (\s,\s) node (SqOne) {};
    
        \node[circle, draw, \colorAxis] (az) at (\ss,\ss){};
        \fill [\colorAxis] (\ss,\ss) circle [x radius=0.7mm, y radius=0.7mm];
        \node[\colorAxis] (axisz) at (\ss + 5mm,\ss - 3mm) [inner sep=0pt, font=\Huge]{$z$};
    
        \coordinate (pt) at (\ptx, \pty);
        \node at (pt) [below = 0mm of pt, xshift = 5mm, font=\Huge, \colorSurface] {$P$};
    
        \draw [\colorCircle, line width= 1.0mm] (\ss,\ss) circle [x radius=\ss, y radius=\ss];
    
        \draw [\colorCircle, line width= 0.7mm, dashed] (\ss,\ss) circle [x radius=\ss*0.825, y radius=\ss*0.825];
    
        \begin{scope}[rotate around={\rotationTwo:(\ptx, \pty)}]
            \path [name path=thrAy] (\ptx, \pty) -- ($ (\ptx, \ss) ! 1.4 ! (\ptx, \pty) $);
            \path [name path=thrAx, red] (\ptx, \pty) -- ($ (\ptx, \pty) ! 0.9 ! (0, \pty) $);
    
            \fill [red, name intersections={of=thrAy and rectangle}]
            (intersection-1) circle (2pt) node (ix) {};
    
            \fill [red, name intersections={of=thrAx and rectangle}]
            (intersection-1) circle (2pt) node (iy) {};
    
            \draw [red, line width= 0.7mm, dashed] (pt) -- (ix);
            \draw [red, line width= 0.7mm, dashed] (pt) -- (iy);
        \end{scope}
    
        \draw [name path=radi, line width= 0.5mm, dashed, \colorCircle, font=\huge] (az) to [edge label=$r$] (pt);
        \fill [point, \colorSurface] (pt) circle [x radius=1mm, y radius=1mm];
        
    \end{scope}
    
    \draw[fill=lightgray,fill opacity=0.03, line width= 1.0mm, gray] (\s - 20mm, 0) rectangle (\s + 50mm, -20mm) node (textRec) {};
    
    \node at (\s + 5mm, 20mm) [xshift = -1mm, font=\LARGE,] {$\bm{1}$};
    \node at (\offset, 20mm) [xshift = -1mm, font=\LARGE,] {$\bm{2}$};
    
    \node [anchor=center, font=\LARGE] at (\s + 15mm, -5mm) {$\mathbf{f}_{{\color{red}tri}}^1({\color{\colorSurface}P}) \neq \mathbf{f}_{{\color{red}tri}}^2({\color{\colorSurface}P})$};
    \node [anchor=center, font=\LARGE] at (\s + 15mm, -15mm) {$\mathbf{f}_{{\color{RoyalBlue}cyl}}^1({\color{\colorSurface}P}) = \mathbf{f}_{{\color{RoyalBlue}cyl}}^2({\color{\colorSurface}P})$};

\end{tikzpicture}}
    \caption{\textbf{Invariant Features} This figure shows, from the $z$--axis (top view of a cylinder), two possible rotation anchors that can be obtained from the orientation tree. Suppose {\color{red}trilinear interpolation}, marked as dashed red, is used for the query point $P$. In that case, we obtain two different interpolated features for the same point in space, which causes discontinuity issues, as shown in Fig. 5 in the paper. However, using the proposed {\color{RoyalBlue} cylindrical interpolation}, marked as dashed blue, the same point will have the same feature, regardless of the rotation around the $z$--axis.}
    \label{fig:comp_interp}
\end{figure}

\section{Ablations}
\label{app:ablations}
This section focuses on providing further understanding of the construction of an oriented-grid encoder. \Cref{app:add_ablations} investigates the different hyperparameters and an additional output representation, while~\cref{app:ablation_radius} the development of the interpolation scheme and the radii chosen. 

\subsection{Hyperparameter Tuning}
\label{app:add_ablations}

\Cref{tab:add_ablations} shows additional ablations relating to the training batch size, sparse convolution kernel size, decoder hidden dimension size, normal regularization coefficient, and 3PSDF output representation~\cite{chibane2020ndf}. These ablations have the same experimental setup as the ones presented in the main paper.
Different hyperparameters do not make much difference in the stability of the proposed encoder. However, as we saw in the paper, the output representation is an essential part of modeling objects' surfaces. In the case of 3PSDF, we implement the output representation developed in~\cite{chibane2020ndf} since the authors do not provide the code for their approach. We note that 3PSDF does not rely solely on the output representation, which explains the differences between the output we obtained with our encoder and their result. Nevertheless, the representation obtains object meshes with more surface noise and extra blobs around the object than the ordinary occupancy solution.

\begin{table}[t]
    \caption{\textbf{Ablations} Experimental results of Oriented Grids for ABC and Thingi10k. The CD is multiplied by $10^{-5}$. The NC is multiplied by $10^{-4}$.}
    \label{tab:add_ablations}
    \resizebox{1\linewidth}{!}{%
    \setlength{\tabcolsep}{2.5pt}\begin{NiceTabular}{@{}lccccccccc@{}}[colortbl-like]
    \CodeBefore
    \rectanglecolor{Gray!20}{1-1}{2-10}
    \Body
    \toprule
     & \multicolumn{2}{c}{\thead{Batch Size}} & \multicolumn{2}{c}{\thead{Kernel Size}} & \multicolumn{2}{c}{\thead{Hidden \\ Dimension}} & \multicolumn{2}{c}{\thead{Normal \\ Weights}} & \thead{3PSDF \\ Represent.} \\ 
     \cmidrule(lr){2-3} \cmidrule(lr){4-5} \cmidrule(lr){6-7} \cmidrule(lr){8-9} 
     & $256$ & $1024$ & $5^3$ & $7^3$ & $64$ & $256$ & $1$ & $0.01$ & \cite{chibane2020ndf} \\\midrule
    CD$\downarrow$ & 0.485 & 0.497 & 0.431 & 0.476 & 0.484 & 0.464 & 0.452 & 0.452 & 175.7 \\
    NC$\downarrow$ & 4.116 & 4.106 & 4.093 & 4.113 & 4.103 & 4.098 & 4.092 & 4.057 & 37.5\\
    IoU $\uparrow$ & 0.998 & 0.998 & 0.998 & 0.998 & 0.998 & 0.998 & 0.997 & 0.998 & 0.845\\\bottomrule
    \end{NiceTabular}}
\end{table}

\begin{figure*}[t]
    \centering
    \setlength{\tabcolsep}{2pt}
    \renewcommand{\arraystretch}{1}
    \resizebox{1\linewidth}{!}{\begin{tabular}{c c c c c}
    \includegraphics[width=0.25\textwidth,keepaspectratio]{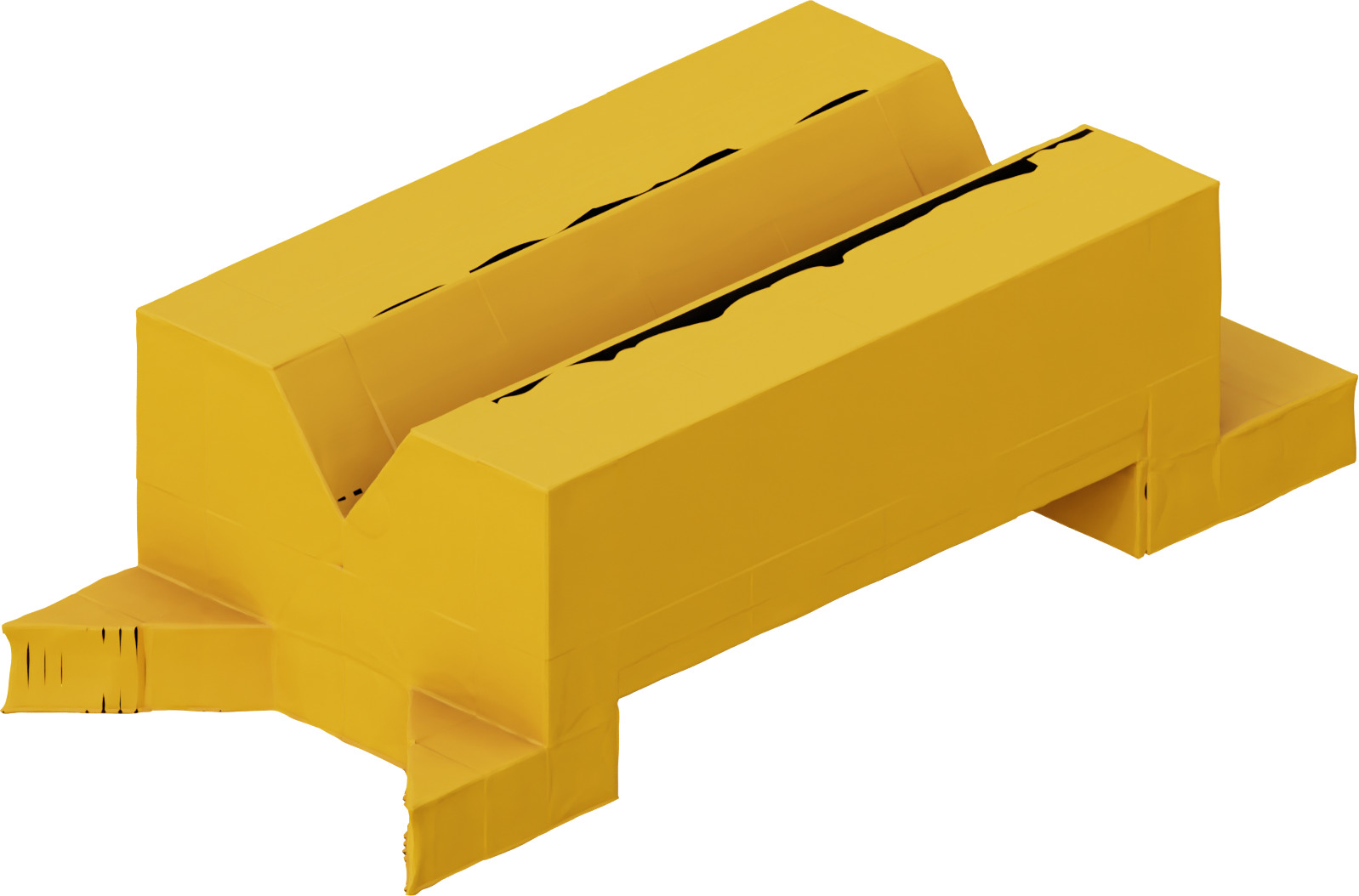}
    & \includegraphics[width=0.25\textwidth,keepaspectratio]{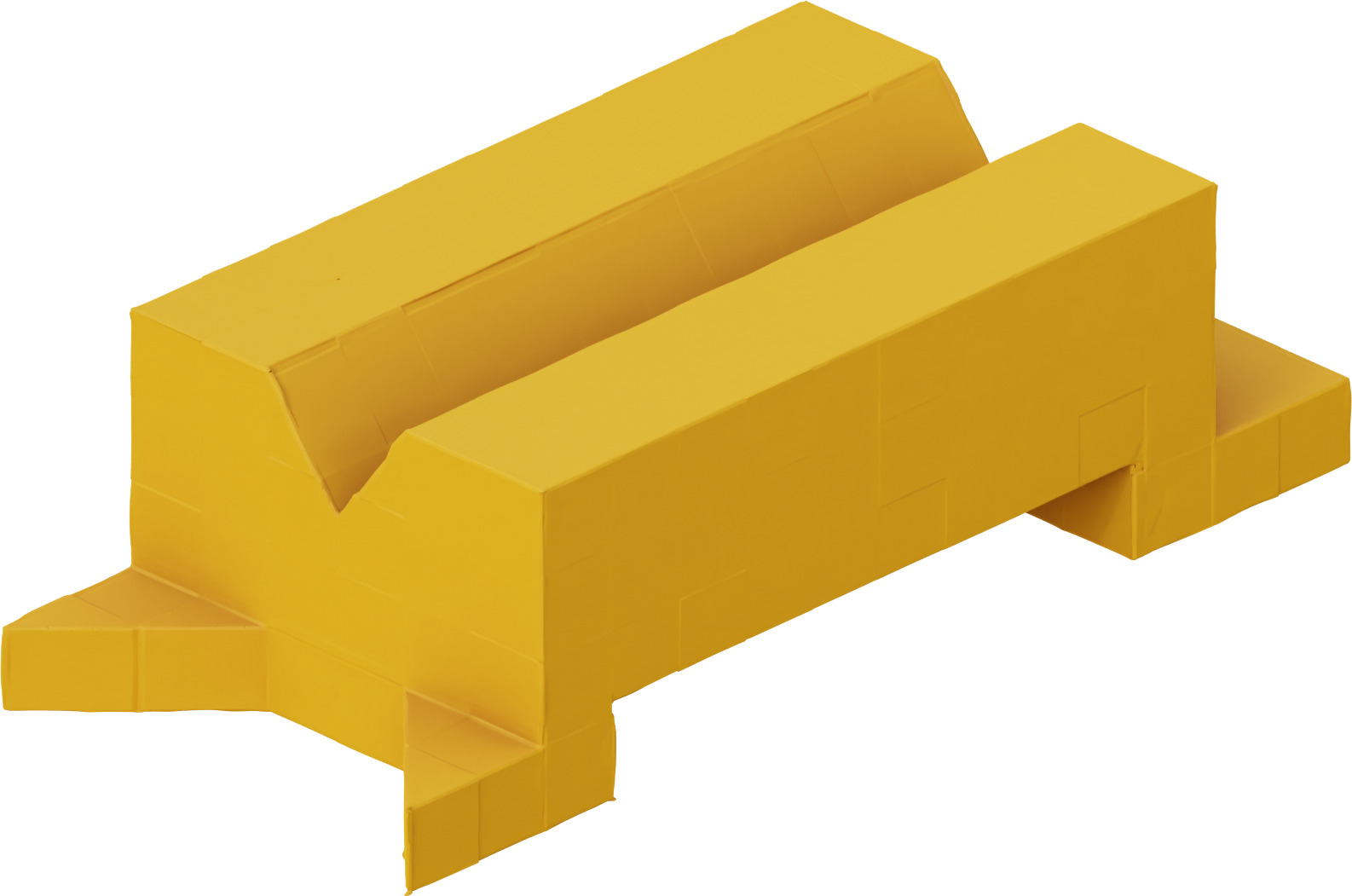}
    & \includegraphics[width=0.25\textwidth,keepaspectratio]{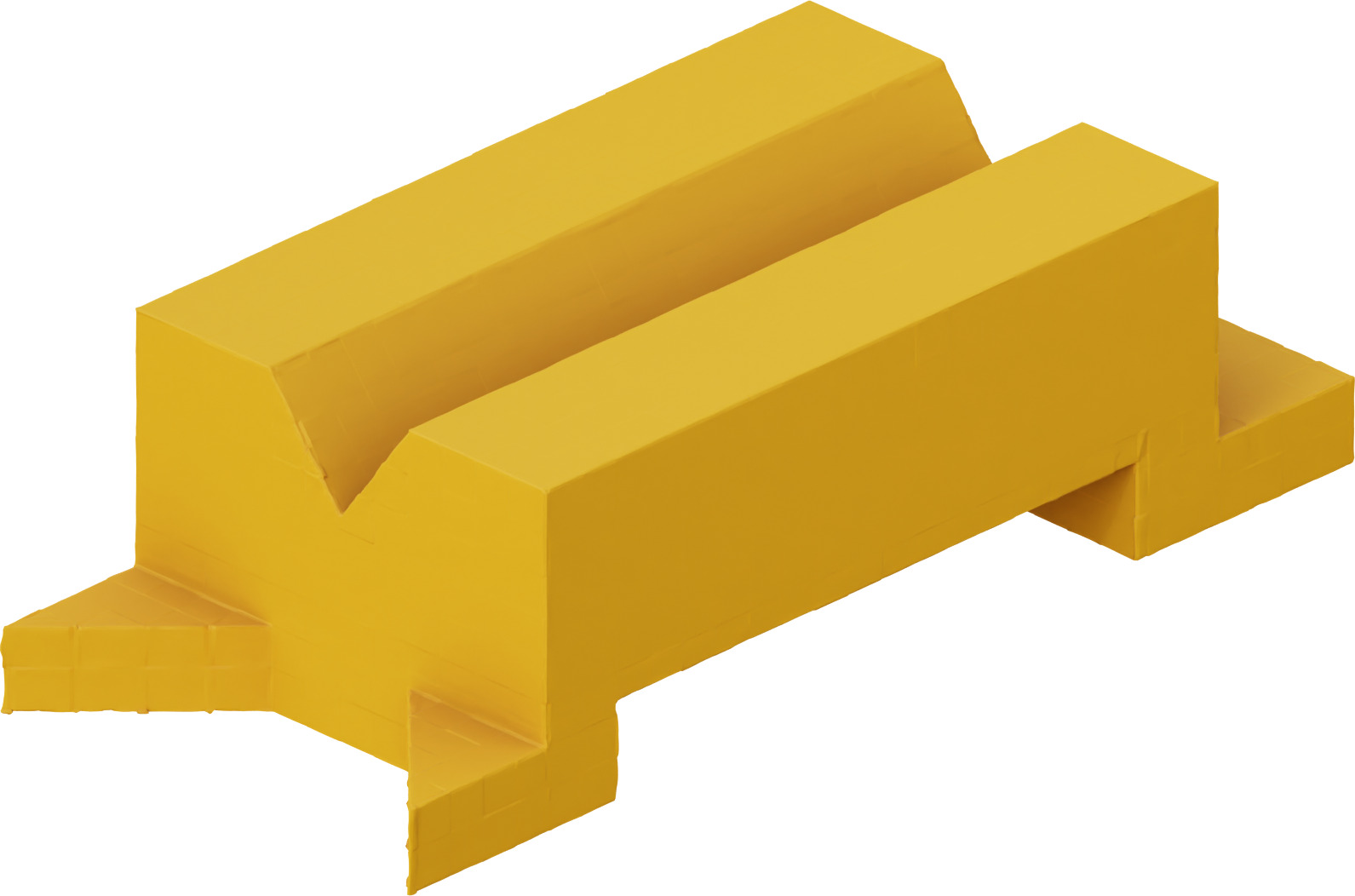} %
    & \includegraphics[width=0.25\textwidth,keepaspectratio]{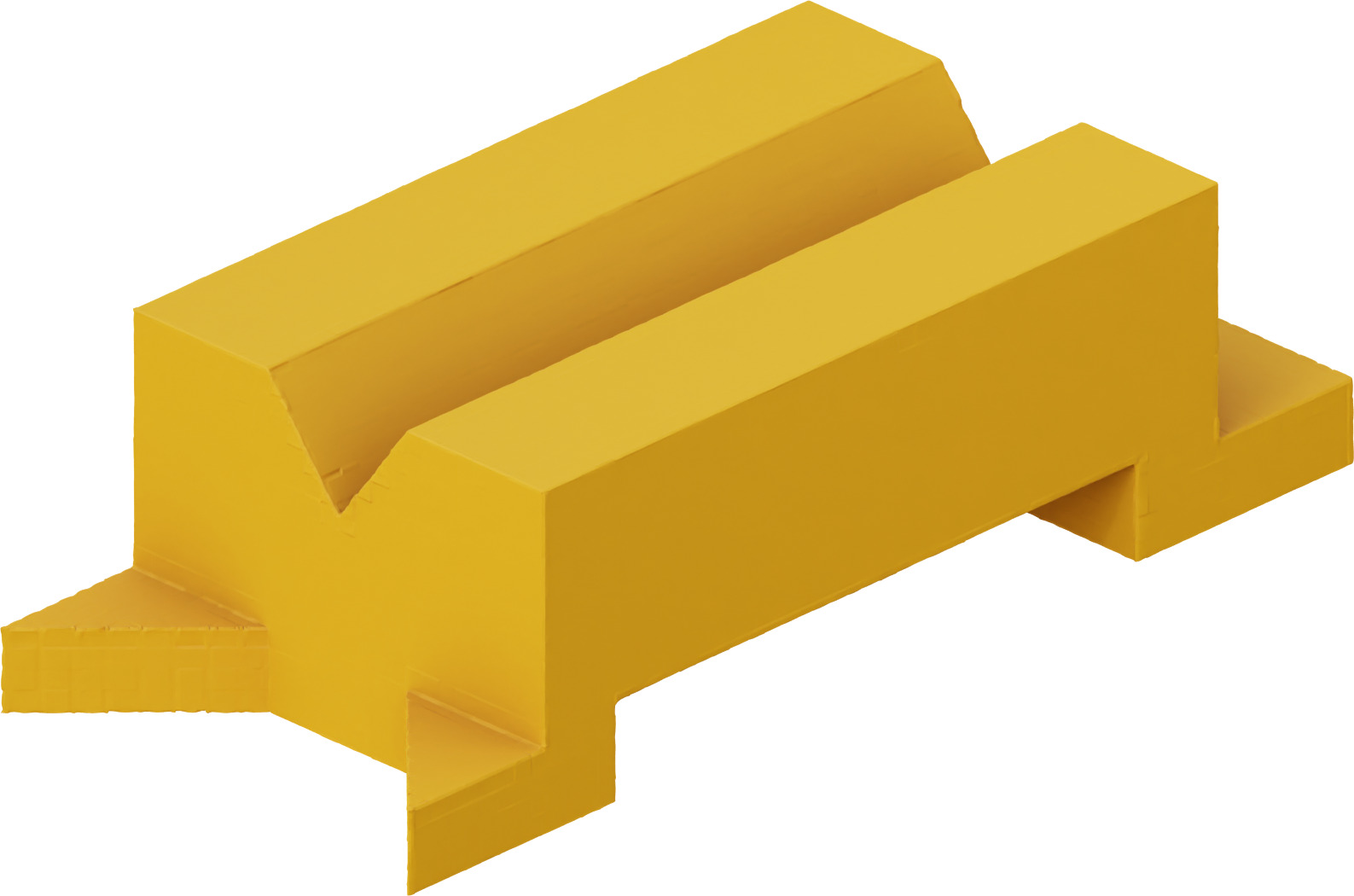}
    & \includegraphics[width=0.25\textwidth,keepaspectratio]{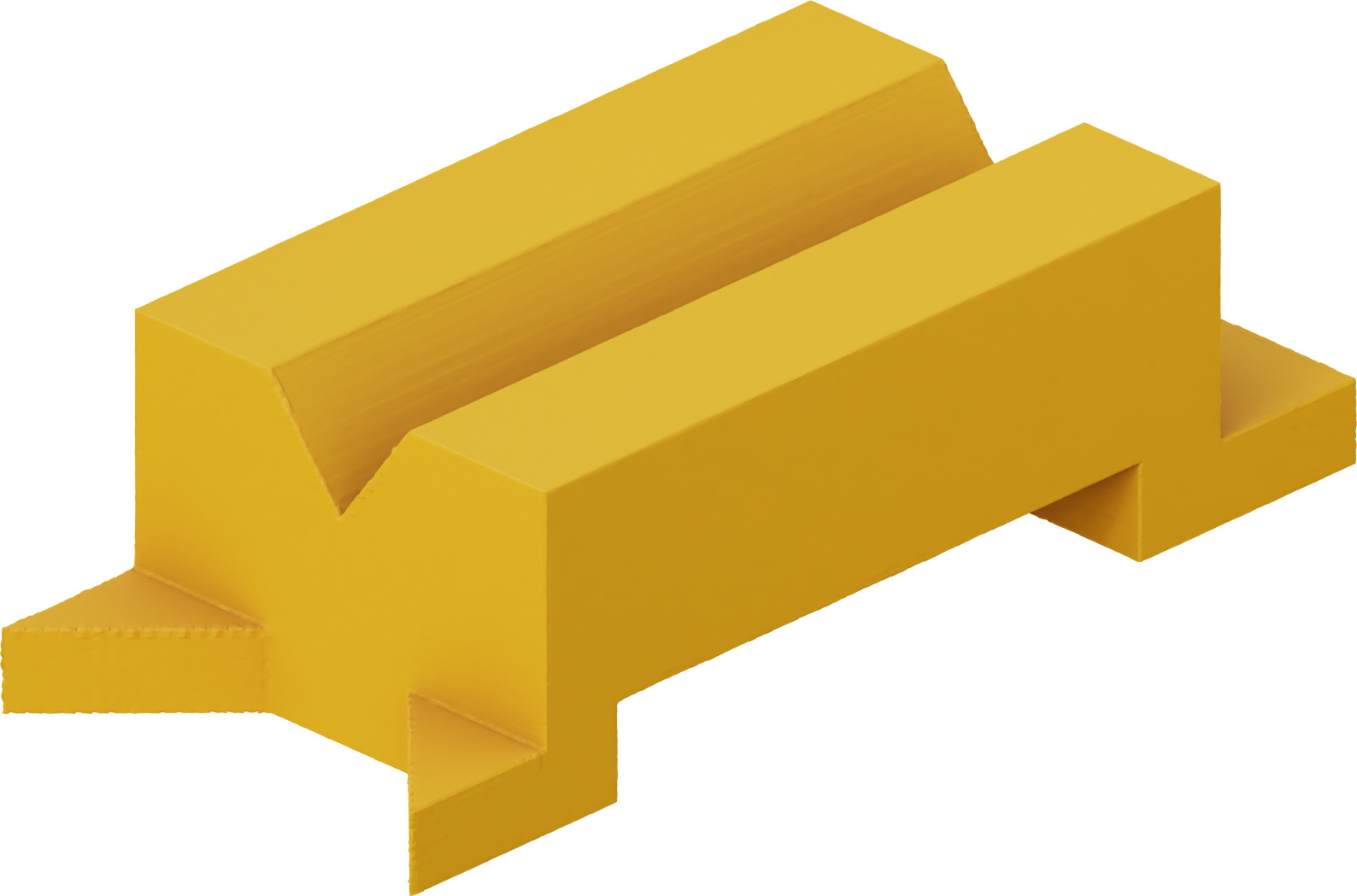}
    \\
    \textbf{LOD 3} & \textbf{LOD 4} & \textbf{LOD 5} & \textbf{LOD 6} & \textbf{LOD 7}
    \end{tabular}}
    \caption{\textbf{LOD results} This figure shows some examples of reconstructed meshes in the ABC (above) and Thingi10k (below) datasets for the baselines.}
    \label{fig:lods_sup}
\end{figure*}

\begin{figure}[t]
    \begin{subfigure}[t]{0.5\linewidth}
        \includegraphics[width=0.45\linewidth]{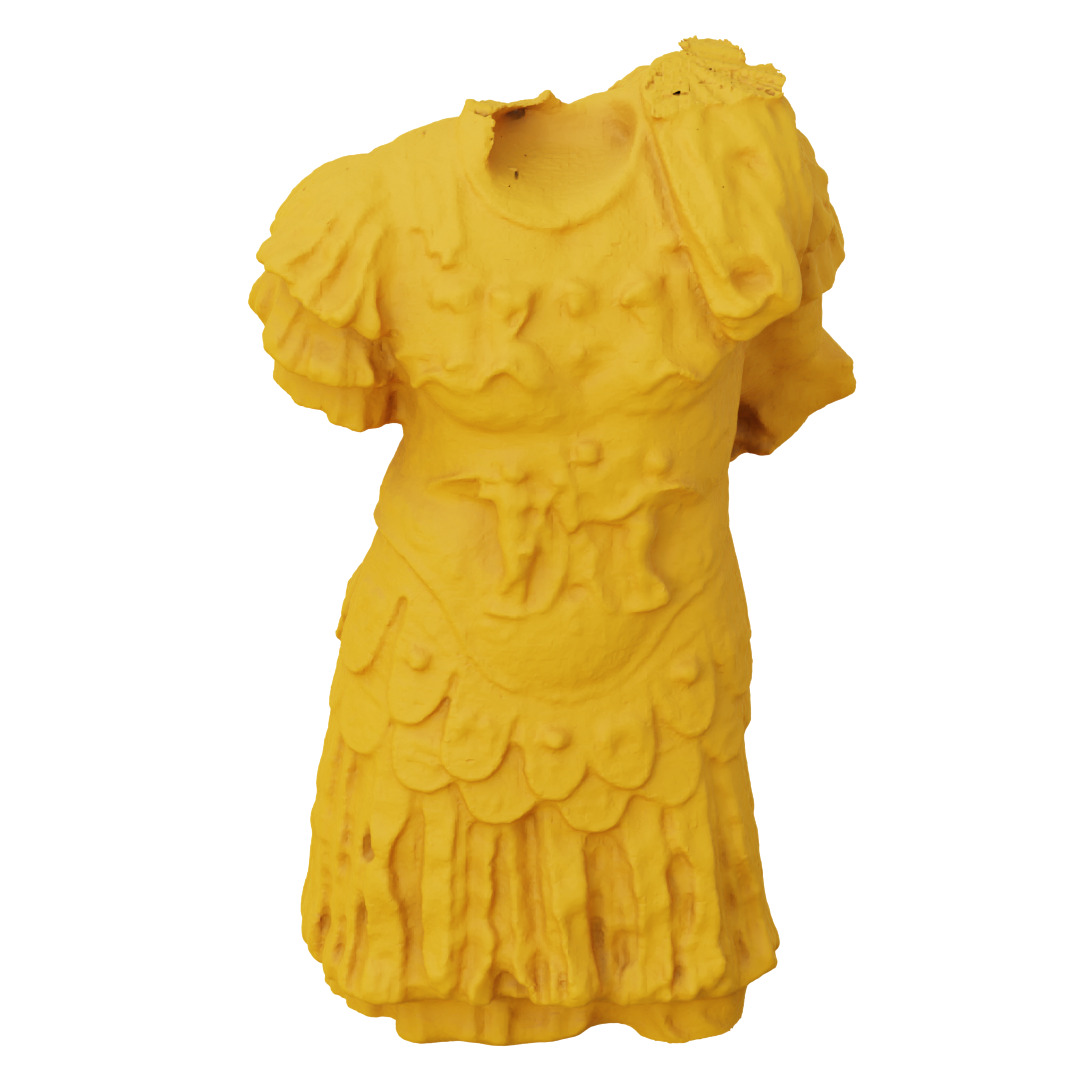}
        \includegraphics[width=0.45\linewidth]{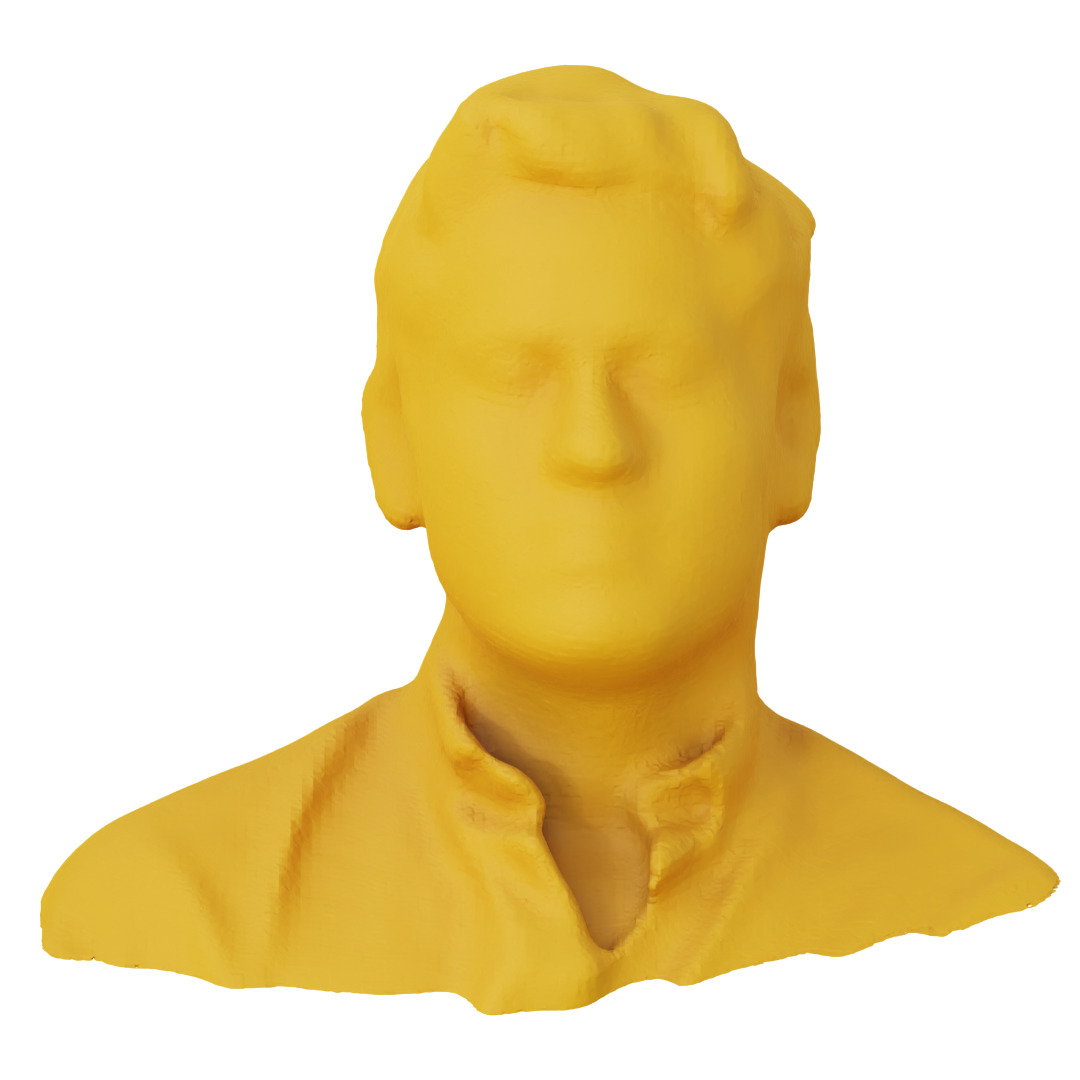}
        \caption{Circumscription}\label{subfig:circum_sup}
    \end{subfigure}%
    \begin{subfigure}[t]{0.5\linewidth}\hfill
        \includegraphics[width=0.45\linewidth]{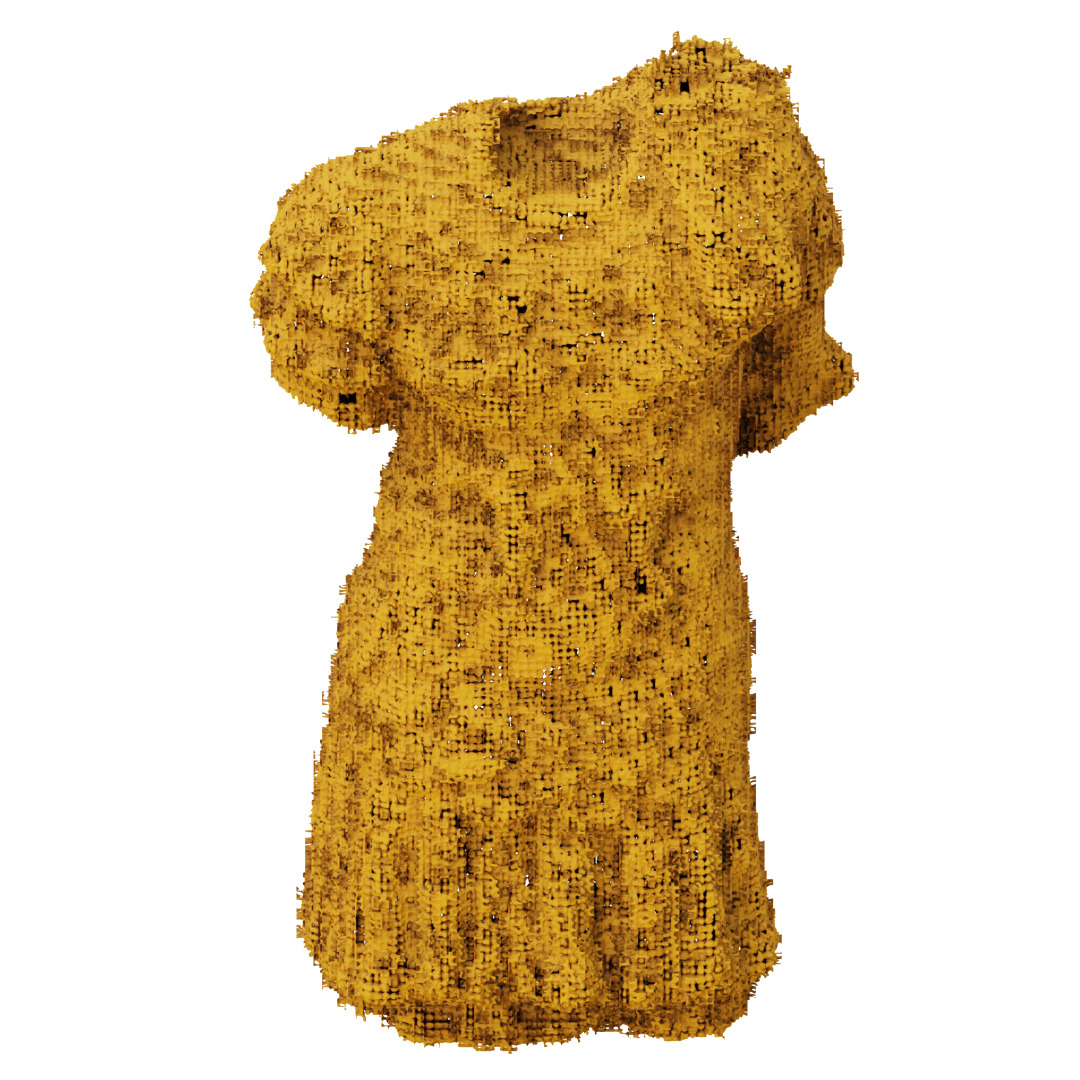}
        \includegraphics[width=0.45\linewidth]{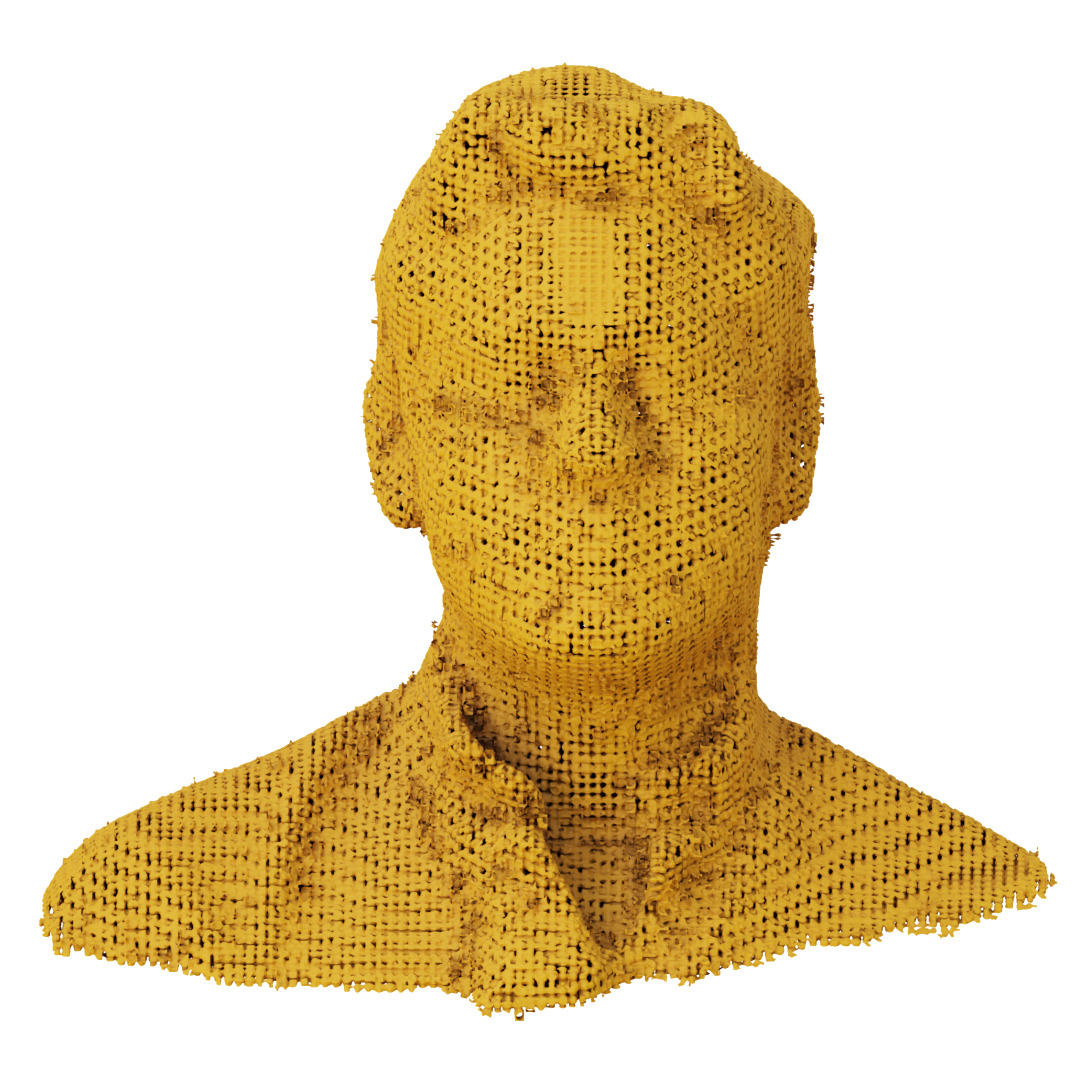}
        \caption{Inscription}\label{subfig:insc_sup}
    \end{subfigure}
    \caption{\textbf{Circumscription \textit{vs} Inscription} This figure shows one mesh from Thingi10K while using circumscription~\protect\subref{subfig:circum_sup} and inscription~\protect\subref{subfig:insc_sup}.}
    \label{fig:circumvsinsc}
\end{figure}

\subsection{Cylinder radius}
\label{app:ablation_radius}

Since cylindrical interpolation depends on the chosen cell radius $R$, we provide insights into choosing the optimal cylinder radius for the irregular grids. 
Specifically, we consider the following cases for radius: Inscription -- $R = \frac{(h_1 + h_2)}{2}$; and Circumscription -- $R = \frac{(h_1 + h_2)}{\sqrt{2}}$.

Using a smaller radius that does not circumscribe the whole cell (interpreting it as a cube) results in holes, as shown in~\cref{fig:circumvsinsc}. 
If full circumscription is not met, some query points that lie near the voxel boundaries but are not enclosed by the cylinder, \textit{i.e.} the projection to the axis of symmetry will be higher than the radius, $r > R$, thus $c_0 = 0$, and $c_2 = 0$. If multiple queries are sampled in these conditions in the same voxel, their interpolated feature $\bar{\mathbf{f}} = \bar{\mathbf{e}}_1^l$ is the same, which leads to classification inconsistencies, resulting in holes.

Circumscription minimizes this issue since it is the minimum distance where all query points are within the cylinder. Further, an increase in radius doesn't provide any benefit during training and evaluation. Since with a higher radius, overlap can occur between cylinders during interpolation. Since feature aggregation between cells and LODs occurs before that step, the overlap does not affect the interpolation scheme. Thus it does not affect neighboring cells' interpolation.

\subsection{Per LOD results}
\label{app:ablation_multires}

In consensus with coarse-to-fine approach for training, we show outputs from LODs $3$ -- $7$ in~\cref{fig:lods_sup} (with $\mathcal{L} = \{3, \ldots, 7 \}$. As expected, with an increase in the LOD, greater details emerge. Lower LODs show grid artifacts corresponding to octree voxels. However, as the LOD level increases, the artifacts disappear, sharpening the edges and smoothing the object's surface.

\section{Experiments}
\label{app:experiments}

\begin{figure}[t]
    \begin{subfigure}[t]{0.24\linewidth}
        \includegraphics[width=1\linewidth]{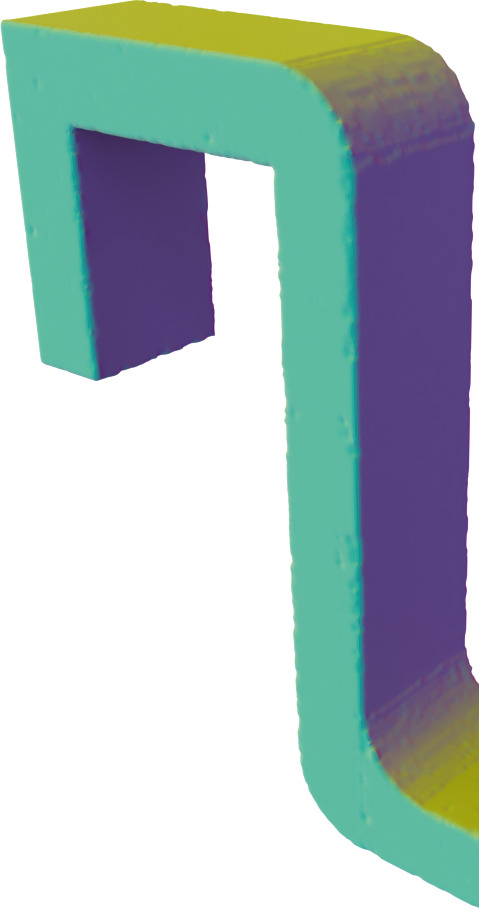}
        \caption{Oriented}\label{subfig:sdf_occ_occ_ours}
    \end{subfigure}%
    \begin{subfigure}[t]{0.24\linewidth}
        \includegraphics[width=1\linewidth]{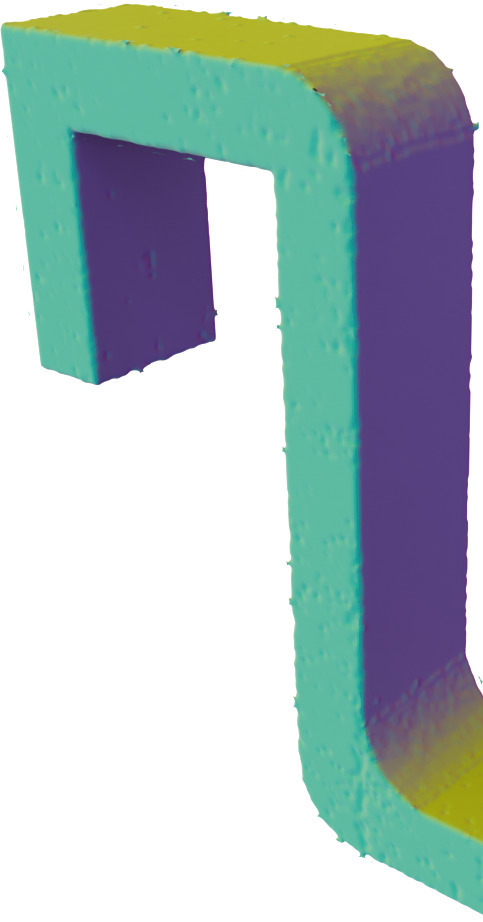}
        \caption{Regular}\label{subfig:sdf_occ_occ_reg}
    \end{subfigure}\hfill \vrule \hfill%
    \begin{subfigure}[t]{0.24\linewidth}
        \includegraphics[width=1\linewidth]{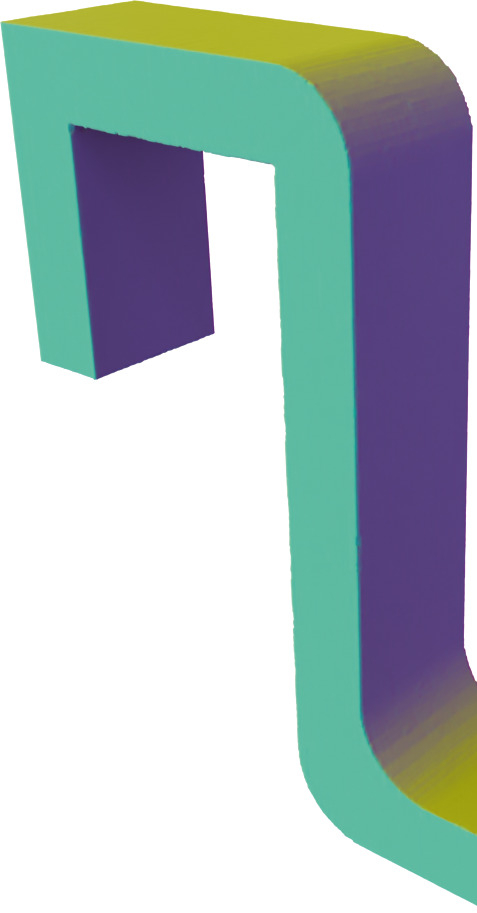}
        \caption{Oriented}\label{subfig:sdf_occ_sdf_ours}
    \end{subfigure}%
    \begin{subfigure}[t]{0.24\linewidth}
        \includegraphics[width=1\linewidth]{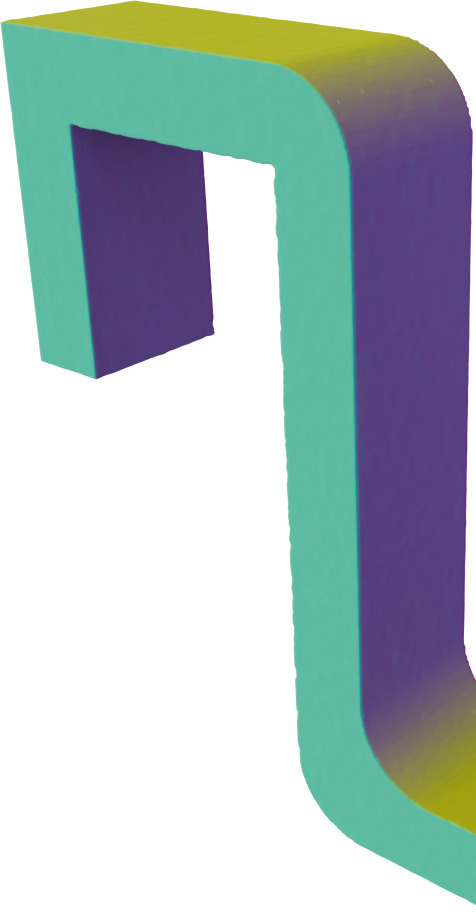}
        \caption{Regular}\label{subfig:sdf_occ_sdf_reg}
    \end{subfigure}%
    \caption{Oriented vs. regular grids in 3D object rendering with normal visualization using Occupancy networks in \protect\subref{subfig:sdf_occ_occ_ours} and \protect\subref{subfig:sdf_occ_occ_reg}, and using SDFs in \protect\subref{subfig:sdf_occ_sdf_ours} and \protect\subref{subfig:sdf_occ_sdf_reg}. SDFs grant a more accurate object surface than occupancy. The oriented encoder produces fewer holes and smoother surfaces than the regular one for occupancy. For SDFs, oriented produces less roughness and sharper edges over the regular grid.}
    \label{fig:experiments_sdf_occ}
\end{figure}

\begin{table}[t]
    \caption{\textbf{Regular \textit{vs} Oriented: Convergence} The CD is multiplied by $10^{-5}$. The NC is multiplied by $10^{-4}$.}
    \label{tab:epochs}
    \resizebox{1\linewidth}{!}{\setlength{\tabcolsep}{2.5pt}\begin{NiceTabular}{@{}lcccccccc}[code-before =%
    \rectanglecolor{Gray!20}{1-1}{2-9}%
    ]
    \toprule 
     & \multicolumn{2}{c}{\thead{Epoch 1}} & \multicolumn{2}{c}{\thead{Epoch 5}} & \multicolumn{2}{c}{\thead{Epoch 10}} & \multicolumn{2}{c}{\thead{Epoch 30}} \\ 
      \cmidrule(lr){2-3} \cmidrule(lr){4-5} \cmidrule(lr){6-7} \cmidrule(lr){8-9}
     & Oriented & Regular & Oriented & Regular & Oriented & Regular & Oriented & Regular \\\midrule
    CD$\downarrow$ & 0.481 & 0.494 & 0.468 & 0.453 & 0.459 & 0.452 & 0.443 & 0.445 \\
    NC$\downarrow$ & 4.100 & 4.118 & 4.100 & 4.259 & 4.097 & 4.258 & 4.058 & 4.256\\
    IoU $\uparrow$ & 0.996 & 0.996 & 0.998 & 0.996 & 0.998 & 0.997 & 0.998 & 0.997 \\ \bottomrule
    \end{NiceTabular}
}
\end{table}

\begin{figure}[t]
    \centering
    \begin{subfigure}[t]{0.45\linewidth}
        \includegraphics[width=0.8\linewidth]{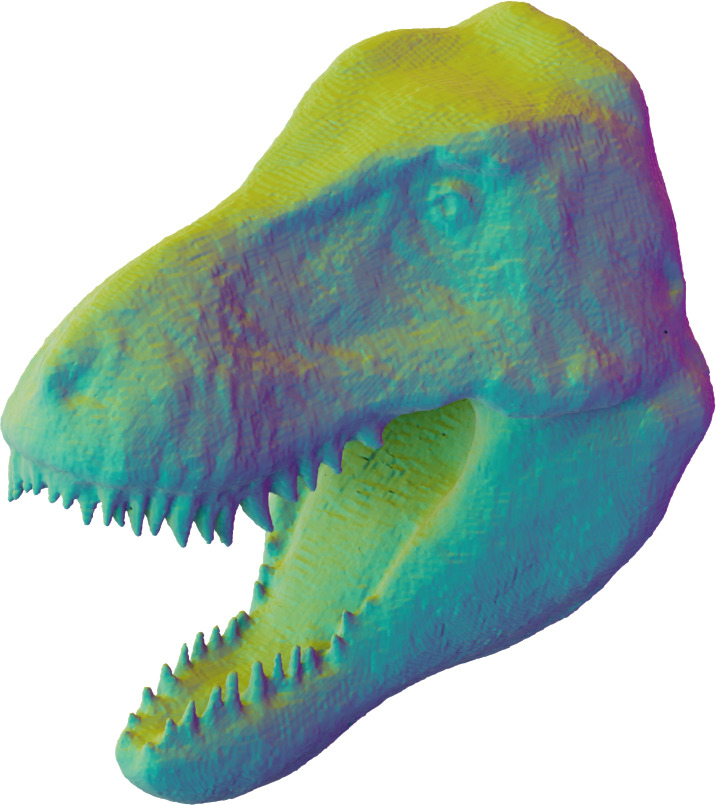}
        \caption{}\label{subfig:ep1_or}
    \end{subfigure} \quad
    \begin{subfigure}[t]{0.45\linewidth}
        \includegraphics[width=0.8\linewidth]{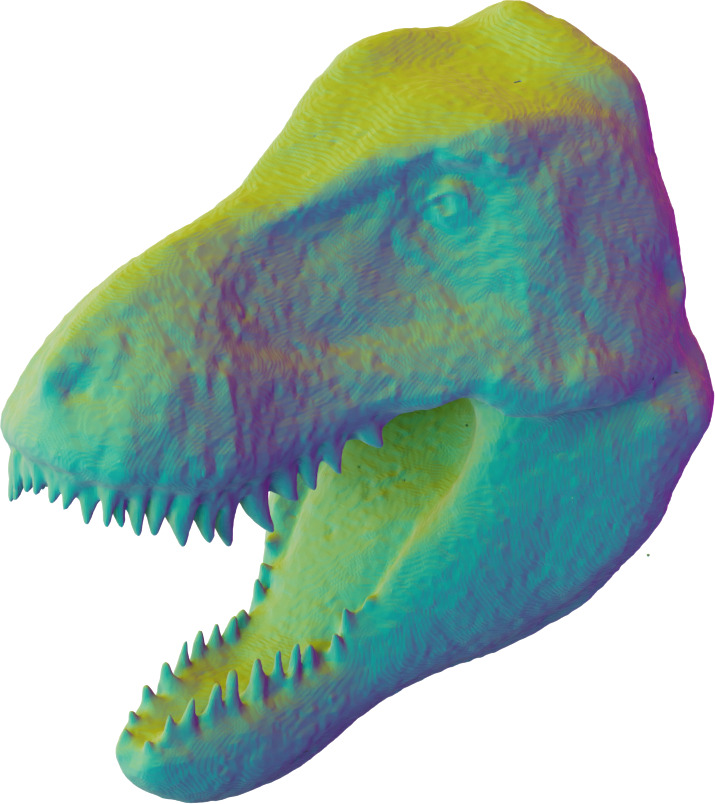}
        \caption{}\label{subfig:ep1_reg}
    \end{subfigure}
    \\
    \begin{subfigure}[t]{0.45\linewidth}
        \includegraphics[width=0.8\linewidth]{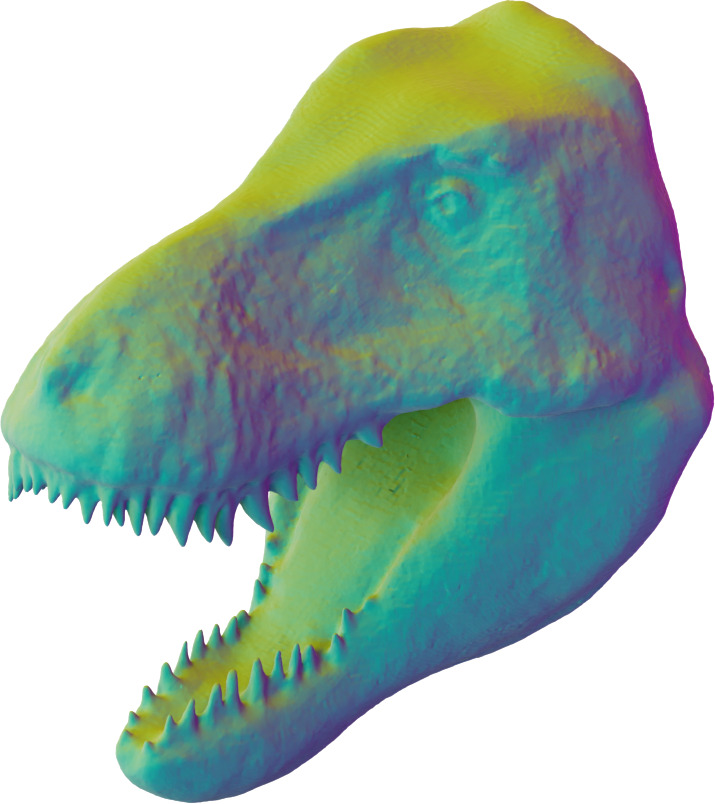}
        \caption{}\label{subfig:ep30_or}
    \end{subfigure} \quad
    \begin{subfigure}[t]{0.45\linewidth}
        \includegraphics[width=0.8\linewidth]{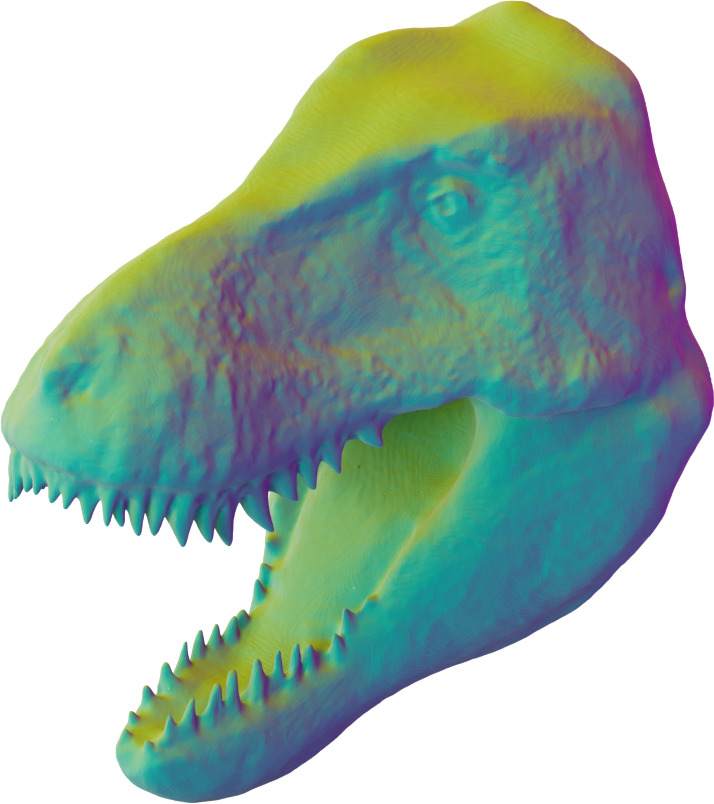}
        \caption{}\label{subfig:ep30_reg}
    \end{subfigure}    
    \caption{\textbf{Epochs} This figure shows one mesh from Thingi10K taken at the first epoch from oriented grids~\protect\subref{subfig:ep1_or} and regular grids~\protect\subref{subfig:ep1_reg} and at 30th epoch from oriented grids~\protect\subref{subfig:ep30_or} and regular grids~\protect\subref{subfig:ep30_reg}.}
    \label{fig:orivsreg_epochs}
\end{figure}

\begin{figure}[t]
    \begin{subfigure}[t]{0.5\linewidth}
        \includegraphics[width=0.45\linewidth]{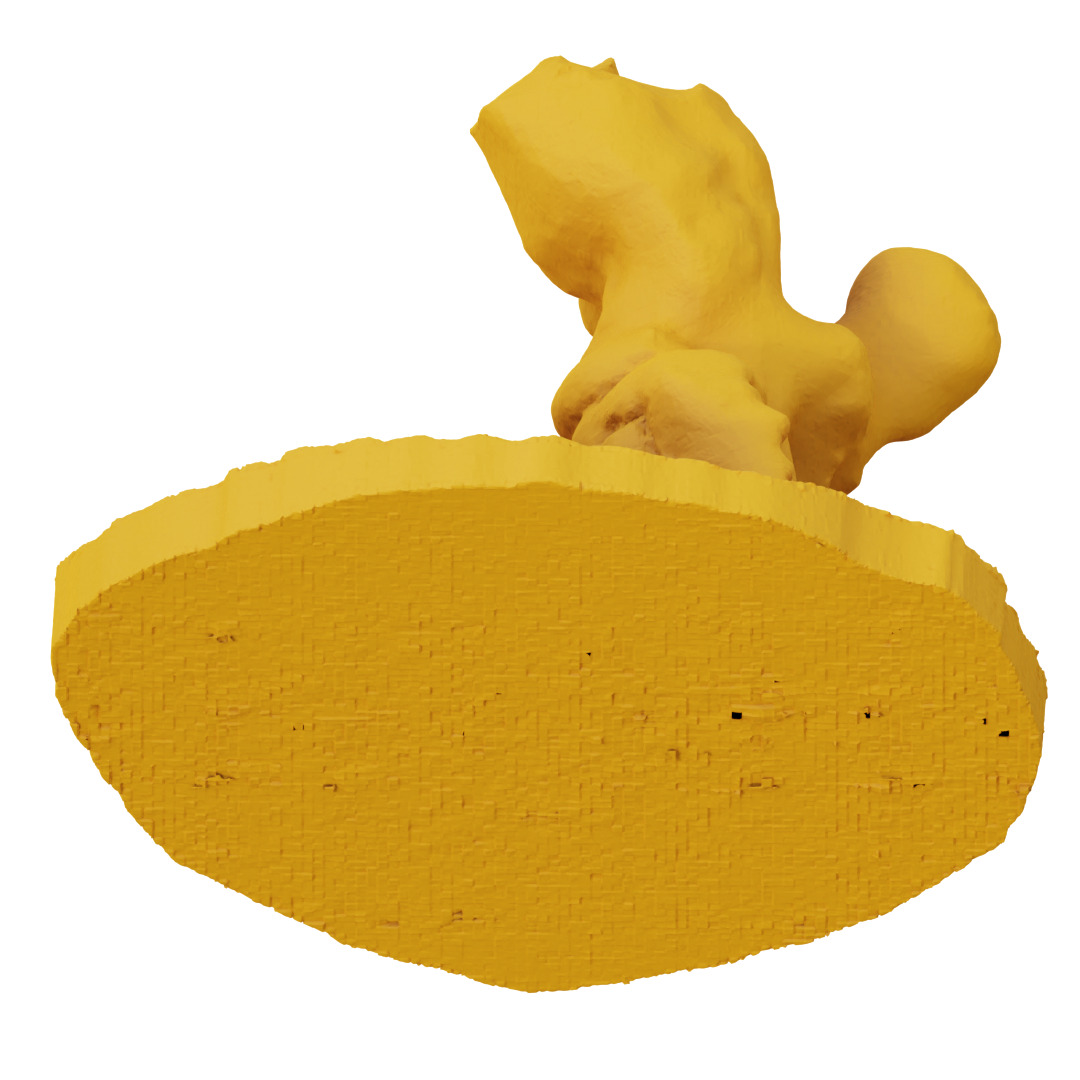}
        \includegraphics[width=0.45\linewidth]{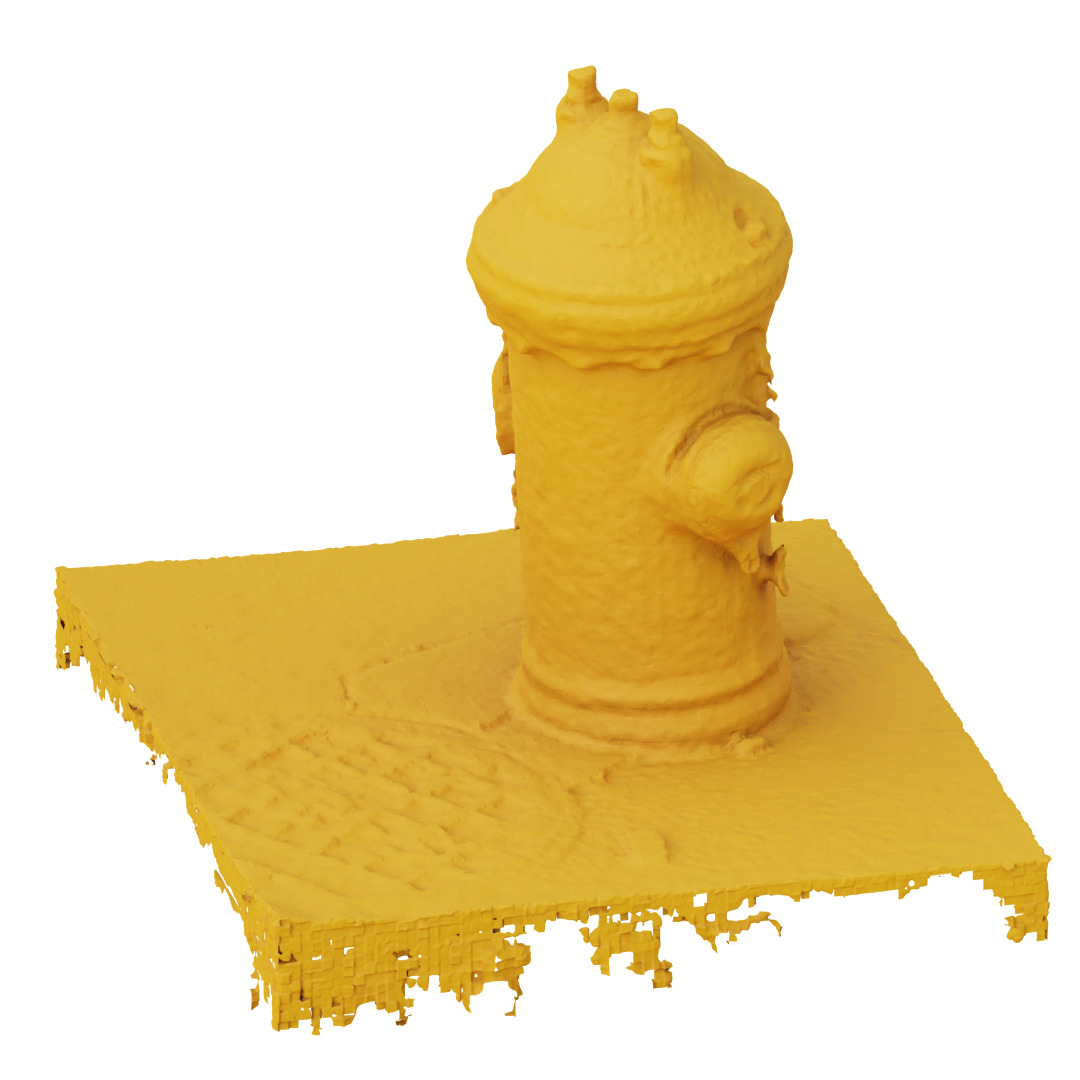}
        \caption{Oriented}\label{subfig:lim_oriented}
    \end{subfigure}%
    \begin{subfigure}[t]{0.5\linewidth}
        \includegraphics[width=0.45\linewidth]{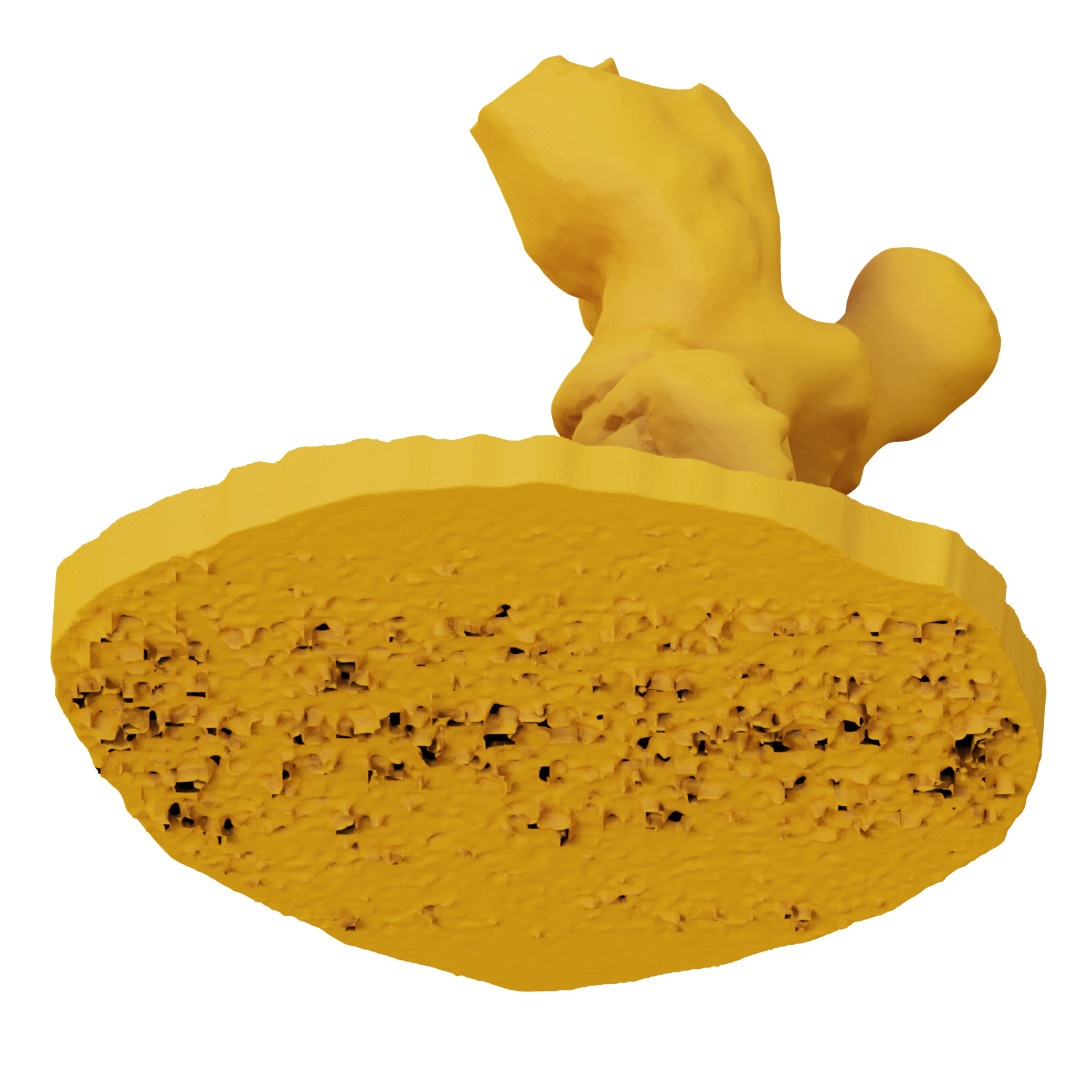}
        \includegraphics[width=0.45\linewidth]{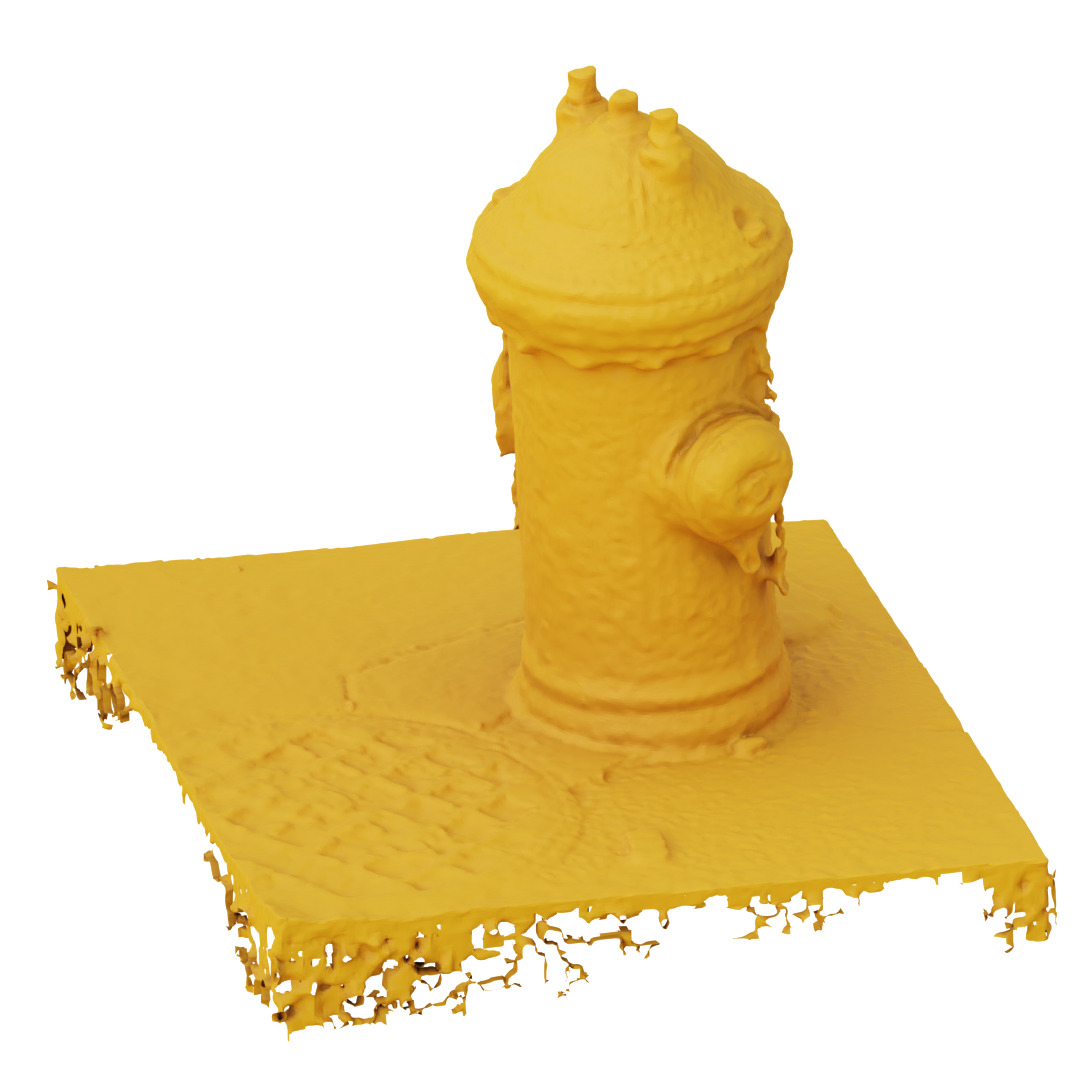}
        \caption{Regular}\label{subfig:lim_regular}
    \end{subfigure}
    \caption{\textbf{Limitations} This figure shows some of the issues we might encounter using a similar decoding strategy. Nonetheless, the proposed encoder~\protect\subref{subfig:lim_oriented} improves qualitatively on the regular grid solution~\protect\subref{subfig:lim_regular}.}
    \label{fig:lim}
\end{figure}

\begin{figure*}[t]
    \centering
    \setlength{\tabcolsep}{2pt}
    \renewcommand{\arraystretch}{1}
    \resizebox{1\linewidth}{!}{\begin{tabular}{c c c c c c}
    \includegraphics[width=0.2\textwidth,keepaspectratio]{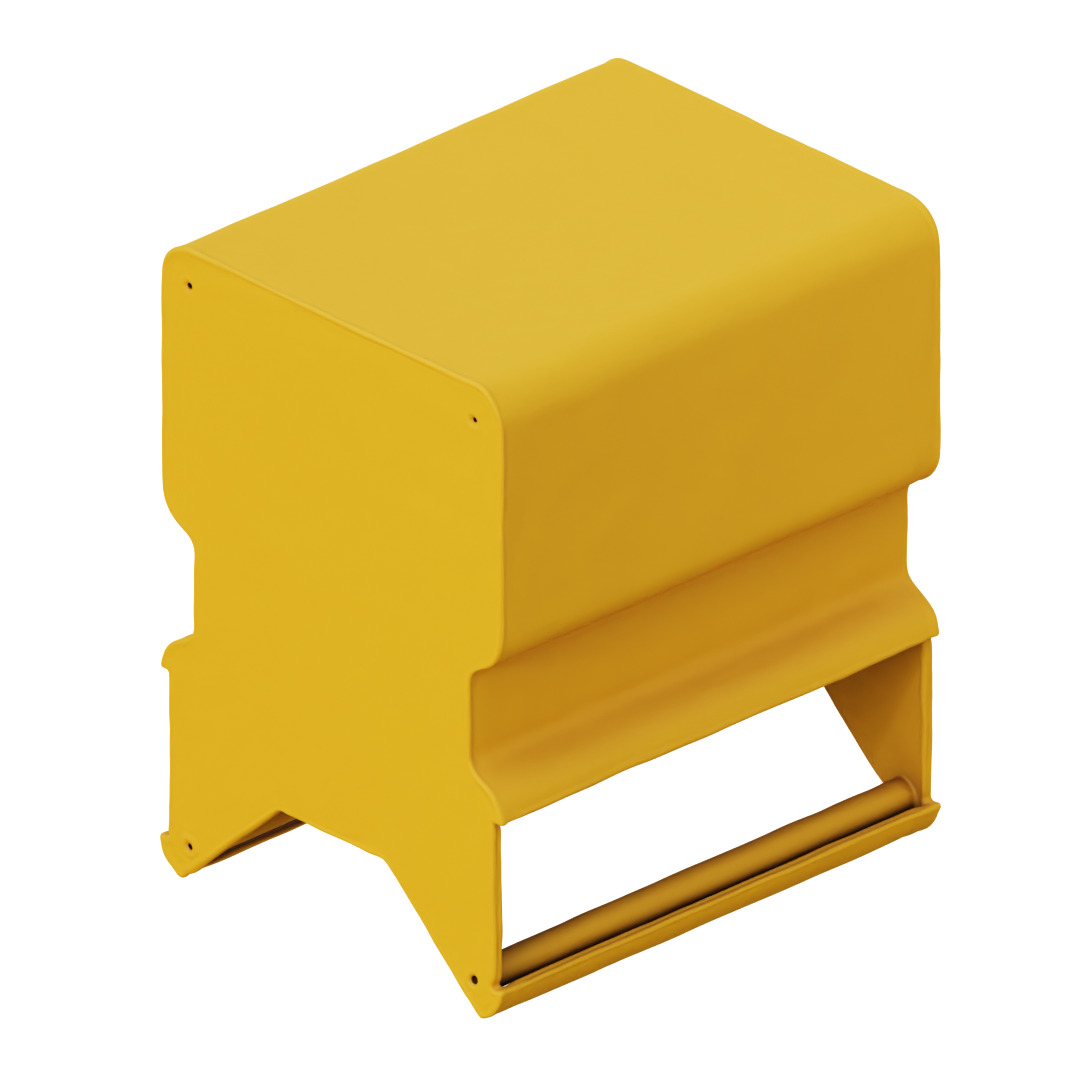}
    & \includegraphics[width=0.2\textwidth,keepaspectratio]{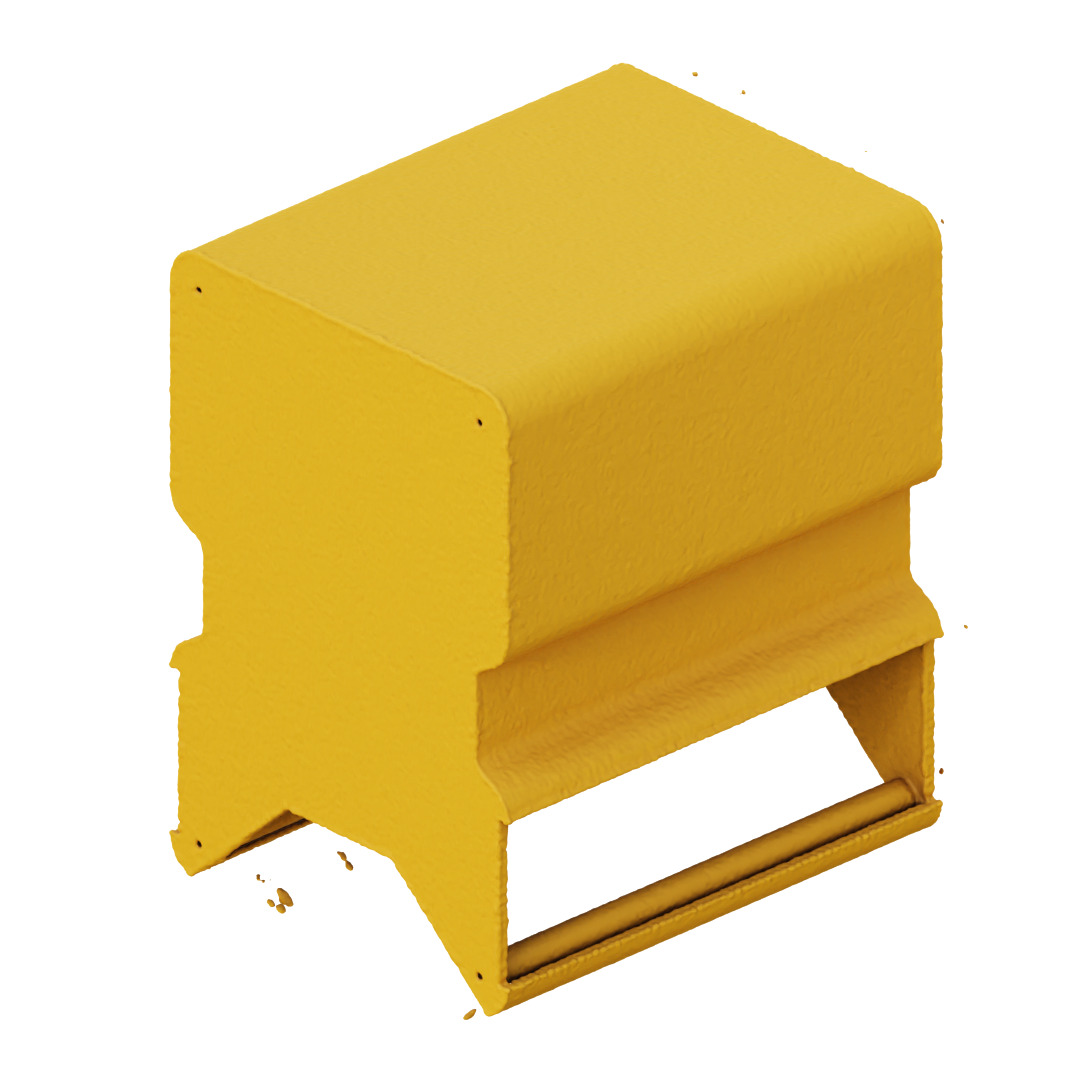}
    & \includegraphics[width=0.2\textwidth,keepaspectratio]{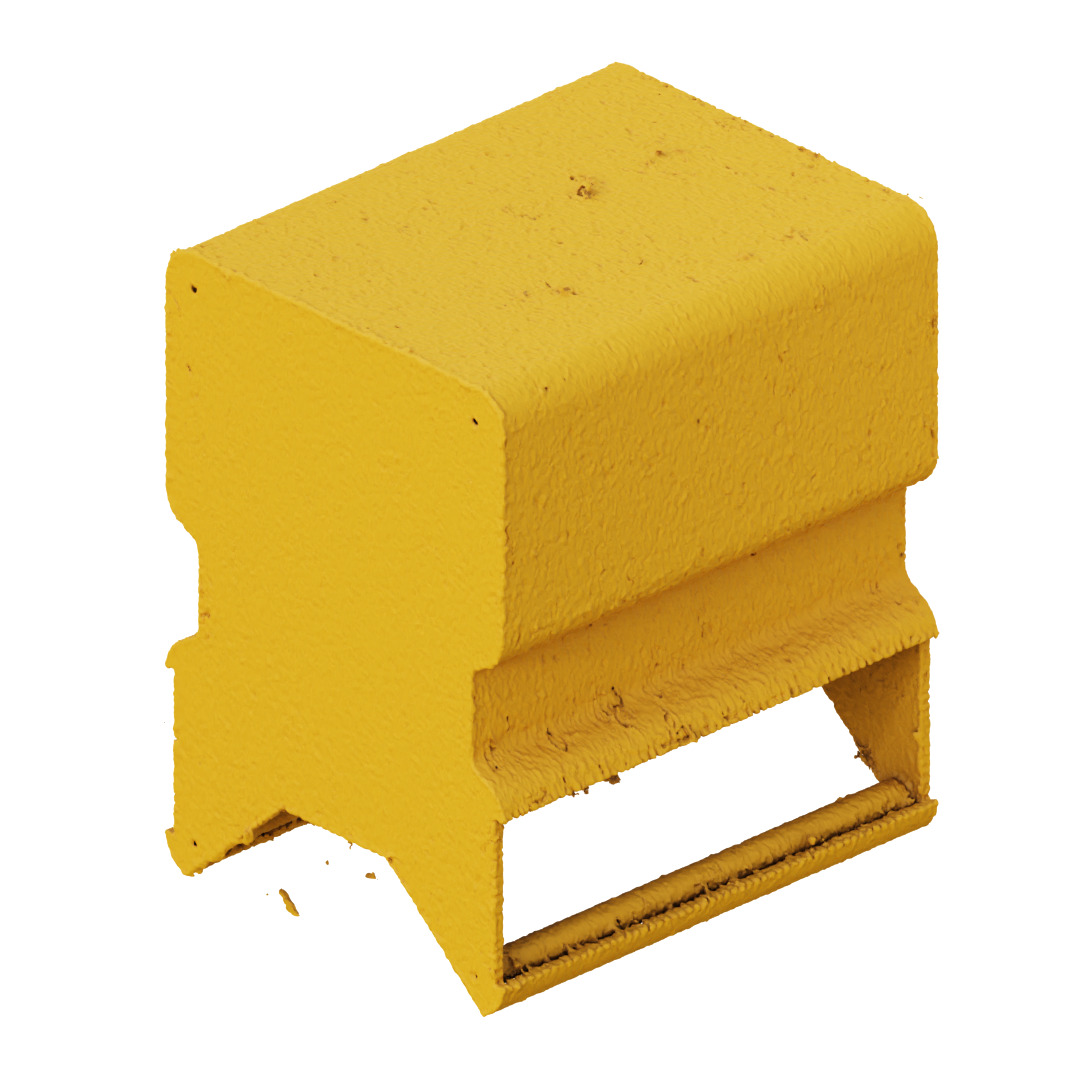} %
    & \includegraphics[width=0.2\textwidth,keepaspectratio]{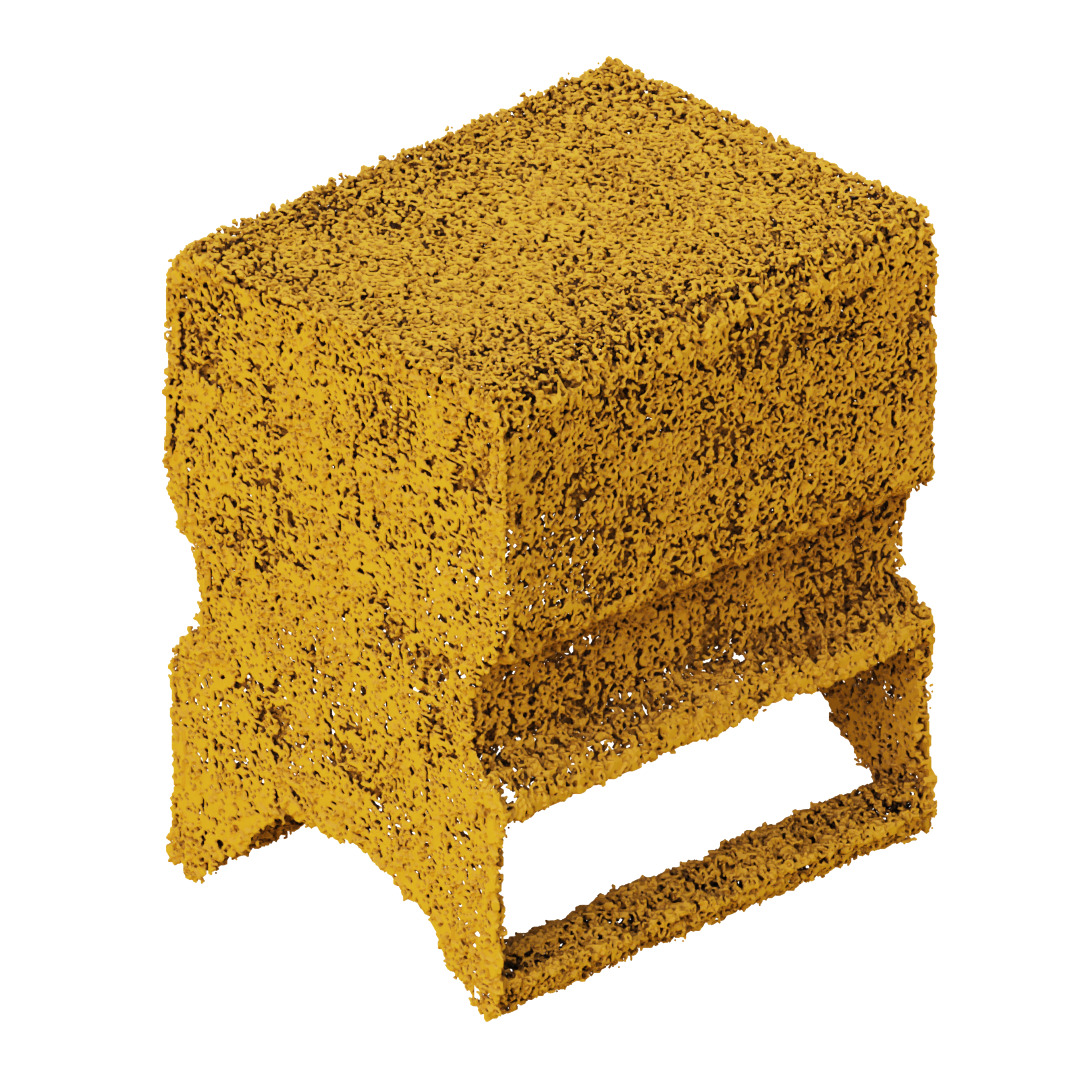}
    & \includegraphics[width=0.2\textwidth,keepaspectratio]{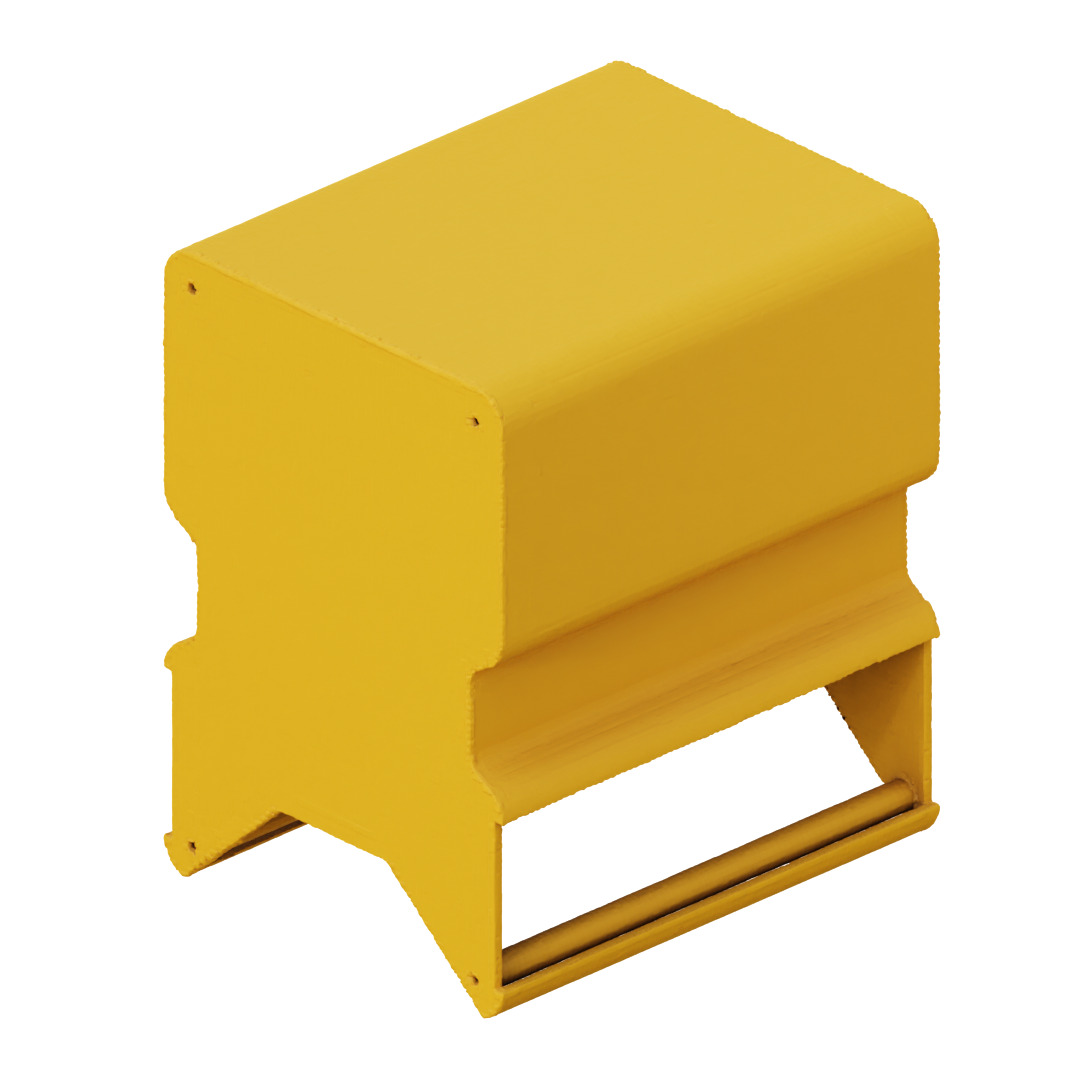}
    & \includegraphics[width=0.2\textwidth,keepaspectratio]{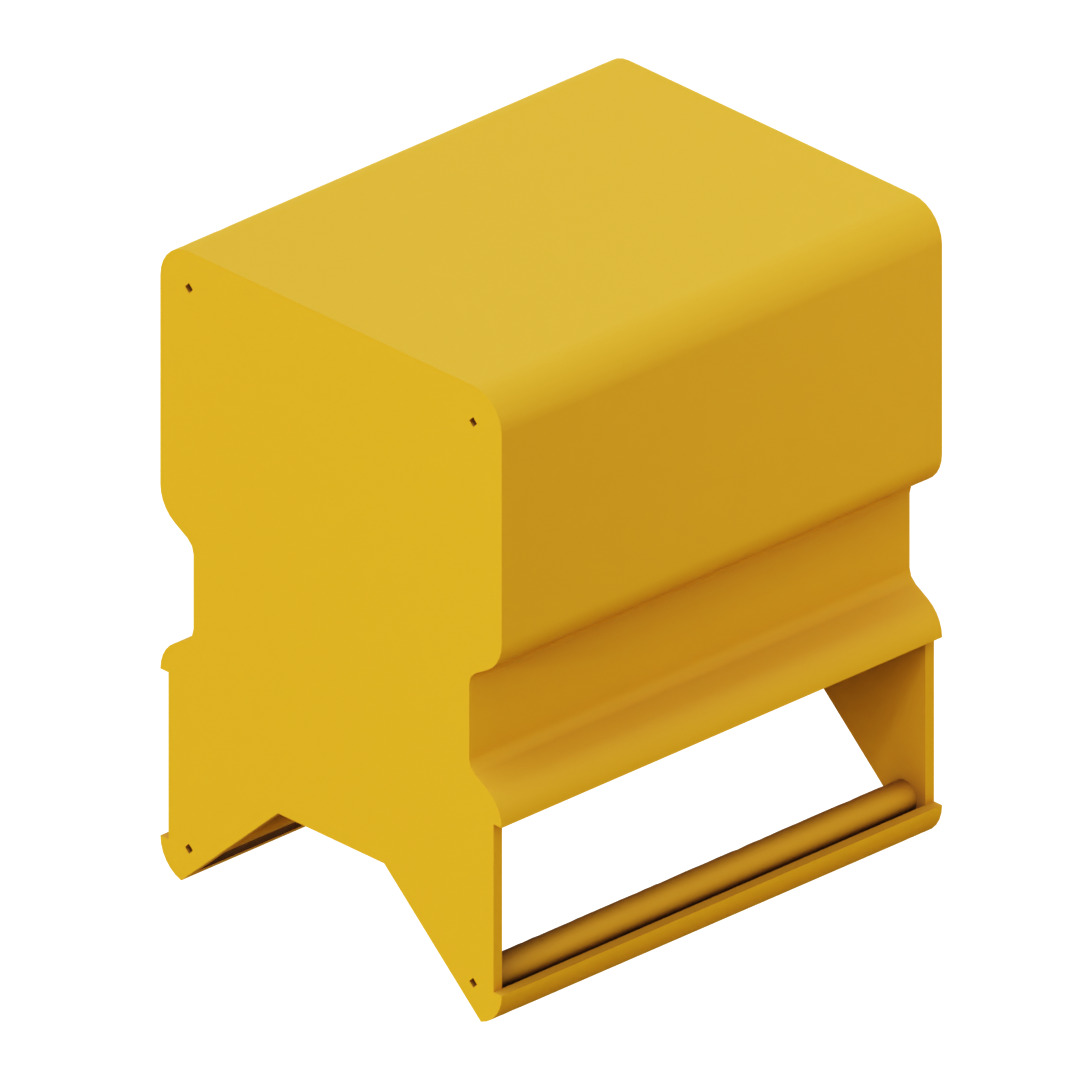} %
    \\
    \includegraphics[width=0.2\textwidth,keepaspectratio]{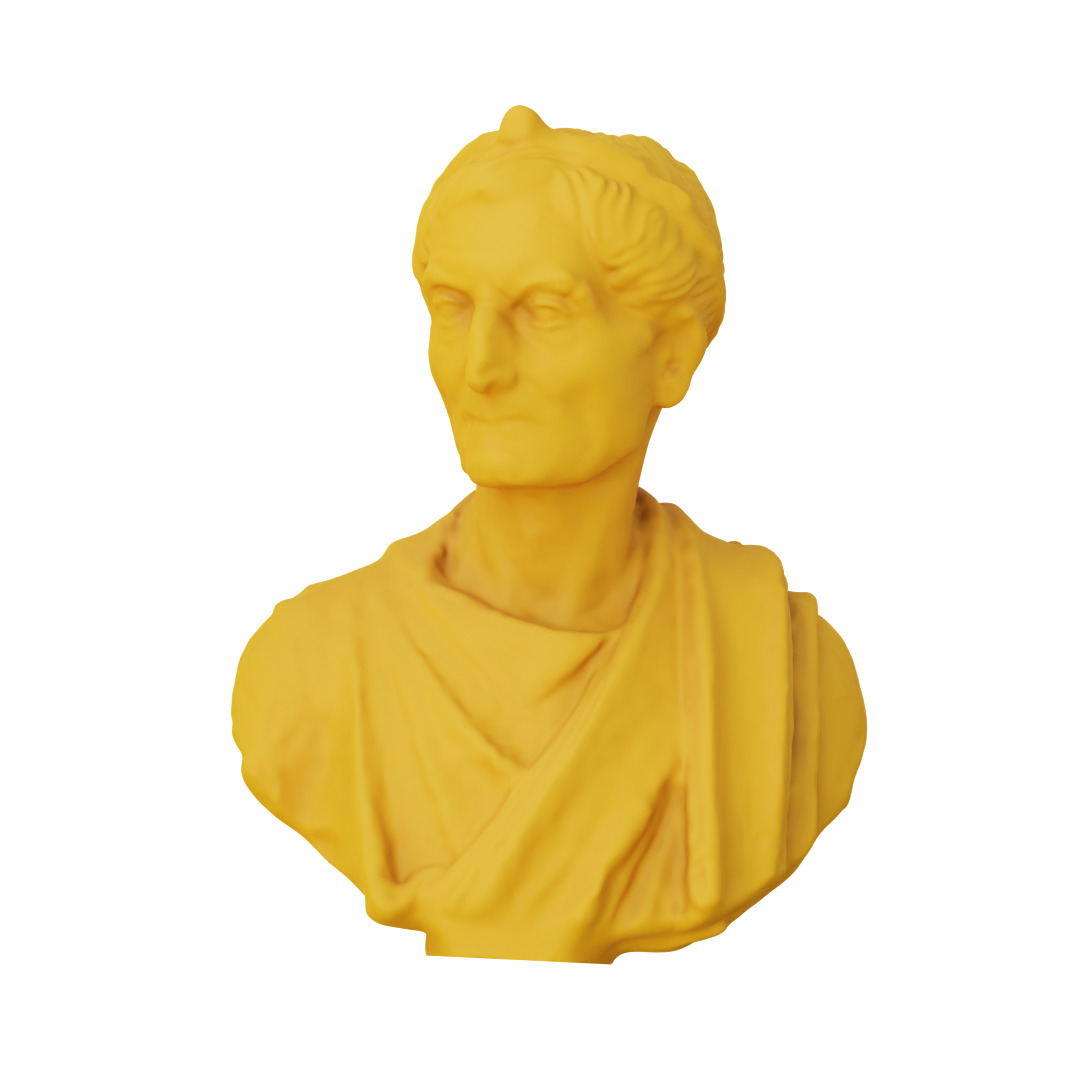}
    & \includegraphics[width=0.2\textwidth,keepaspectratio]{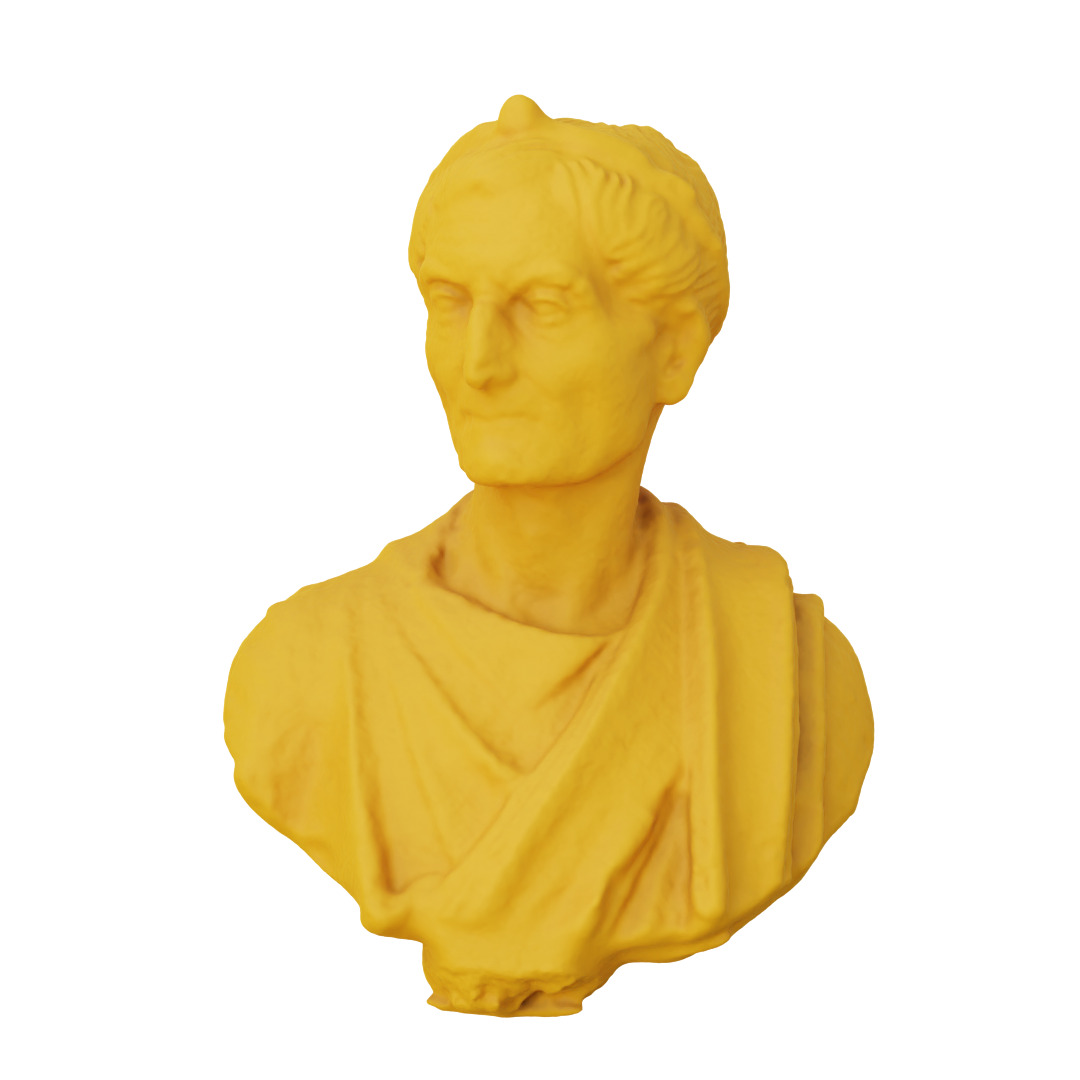}
    & \includegraphics[width=0.2\textwidth,keepaspectratio]{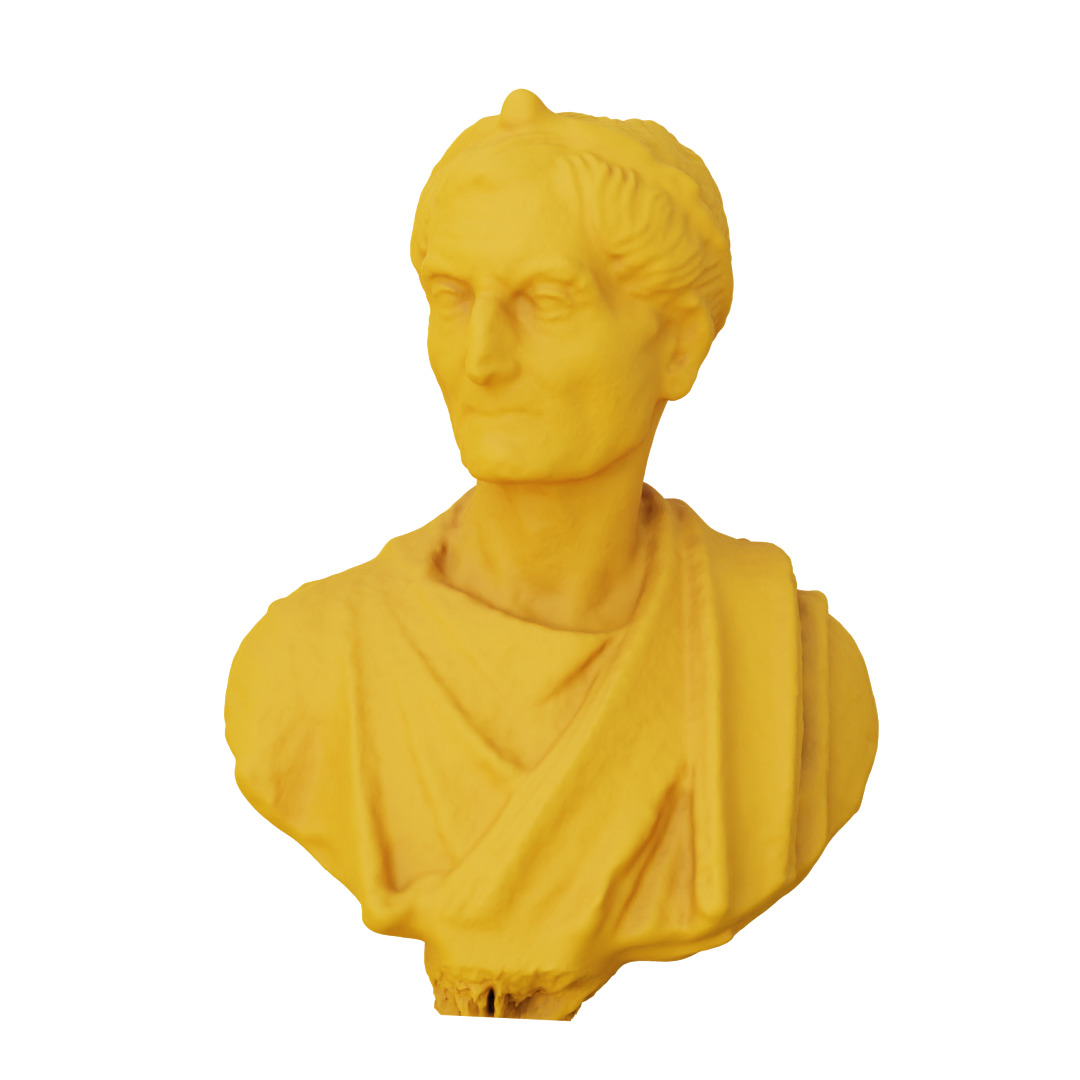}
    & \includegraphics[width=0.2\textwidth,keepaspectratio]{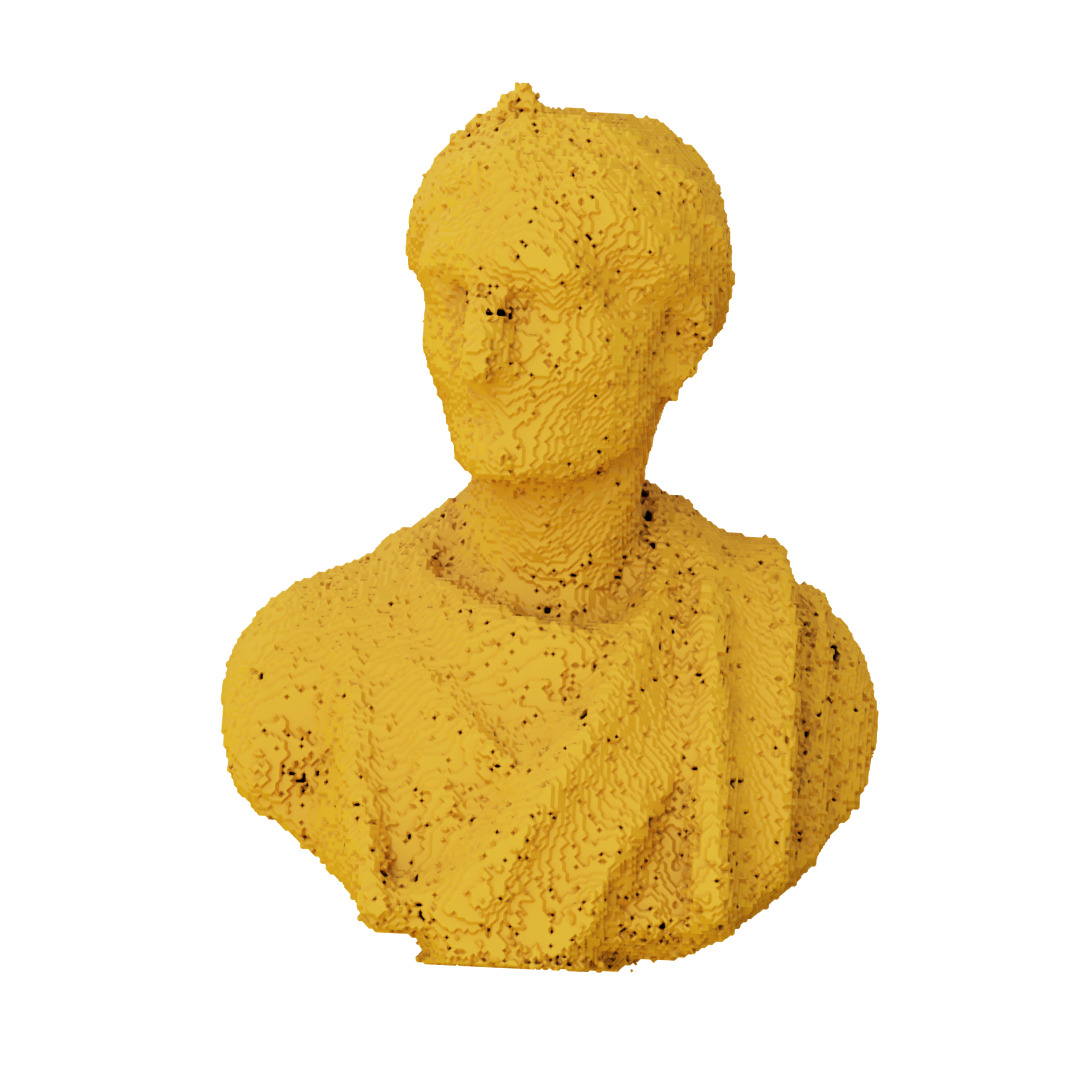}
    & \includegraphics[width=0.2\textwidth,keepaspectratio]{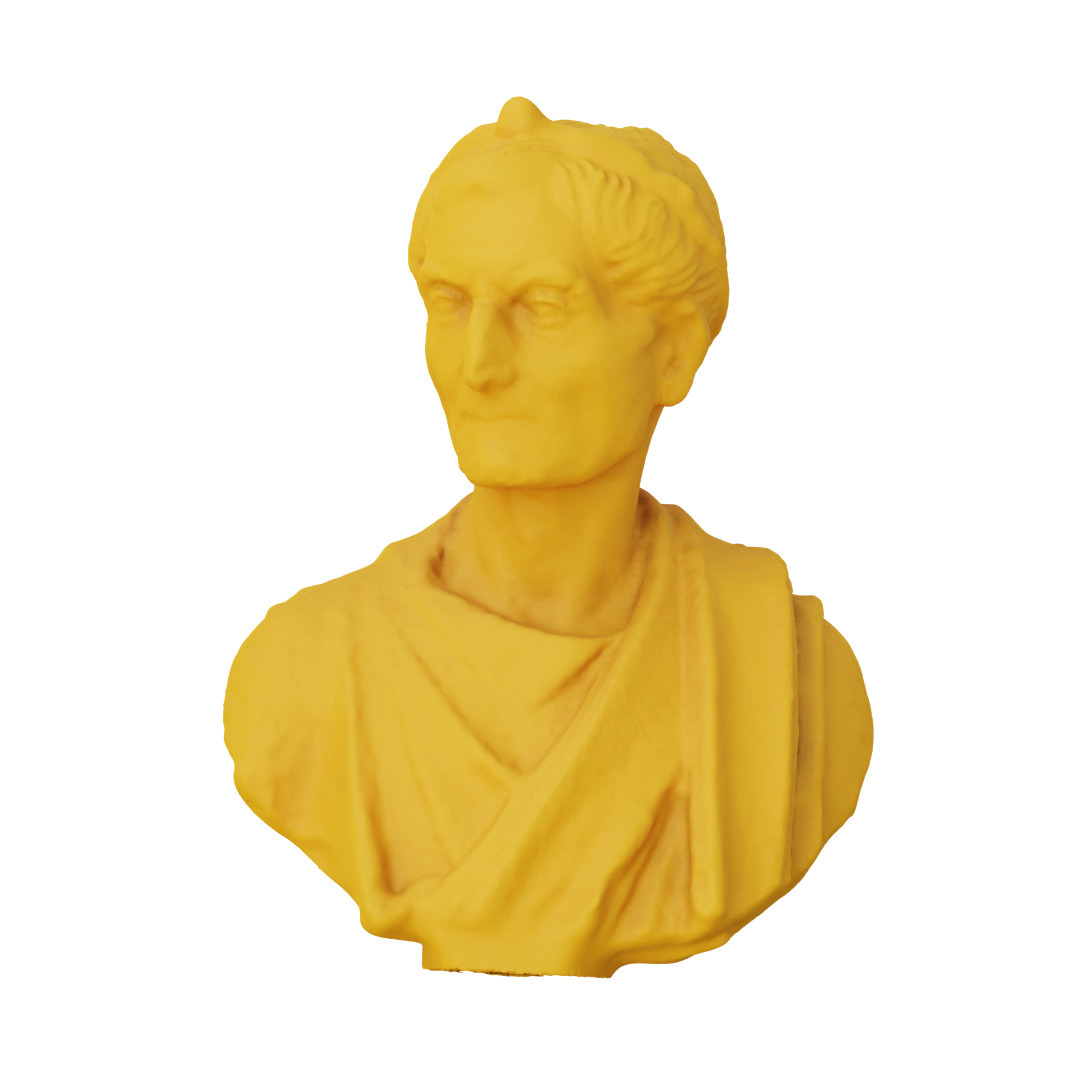}
    & \includegraphics[width=0.2\textwidth,keepaspectratio]{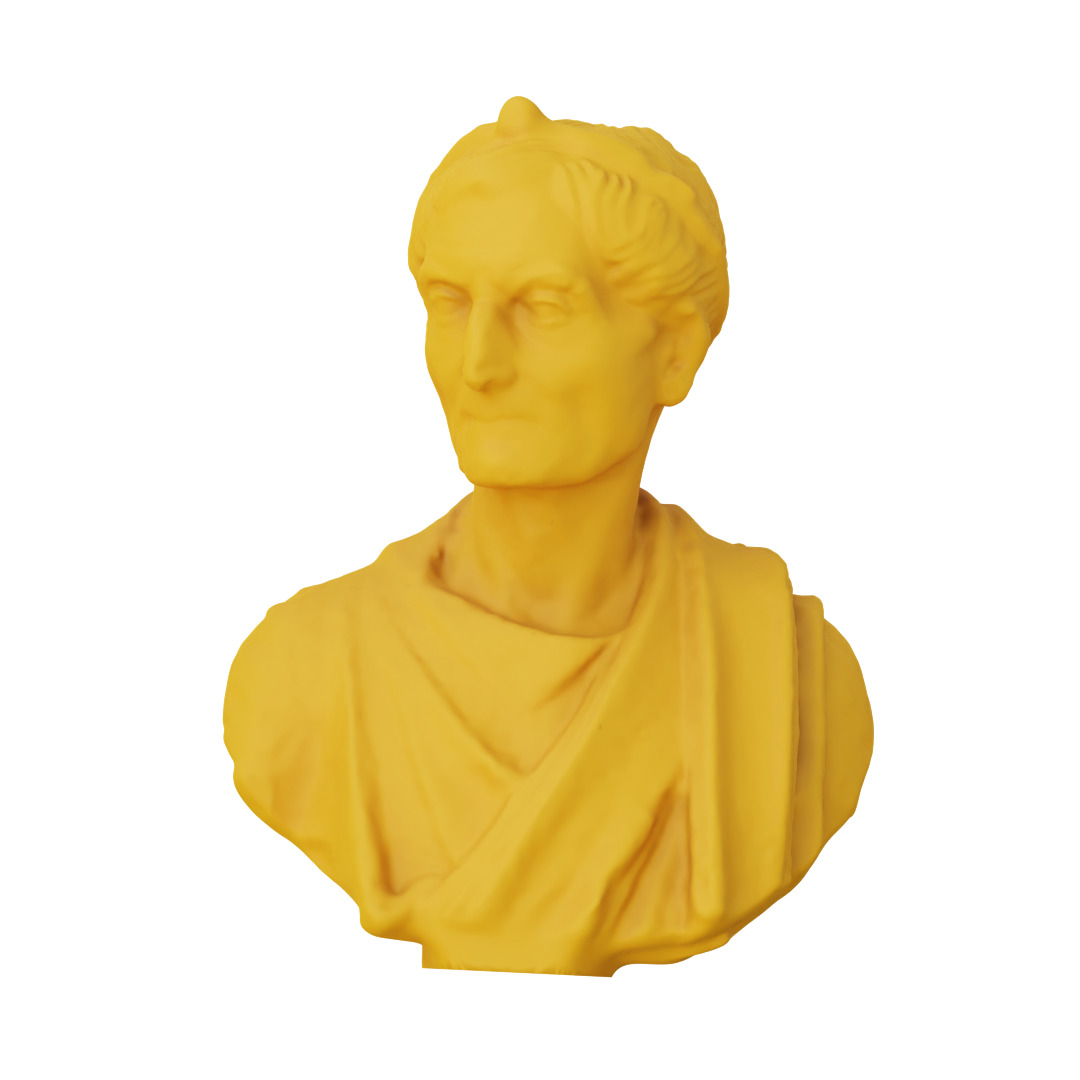}
    \\
    \textbf{SIREN} & \textbf{BACON} & \textbf{FF} & \textbf{NDF} & \textbf{Ours} & \textbf{GT}
    \end{tabular}}
    \caption{\textbf{Additional Baselines}. This figure shows reconstructed meshes in the ABC (above) and Thingi10k (below) datasets for the baselines.}
    \label{fig:baselines_sup}
    \begin{subfigure}[t]{0.65\linewidth}
        \includegraphics[width=\linewidth]{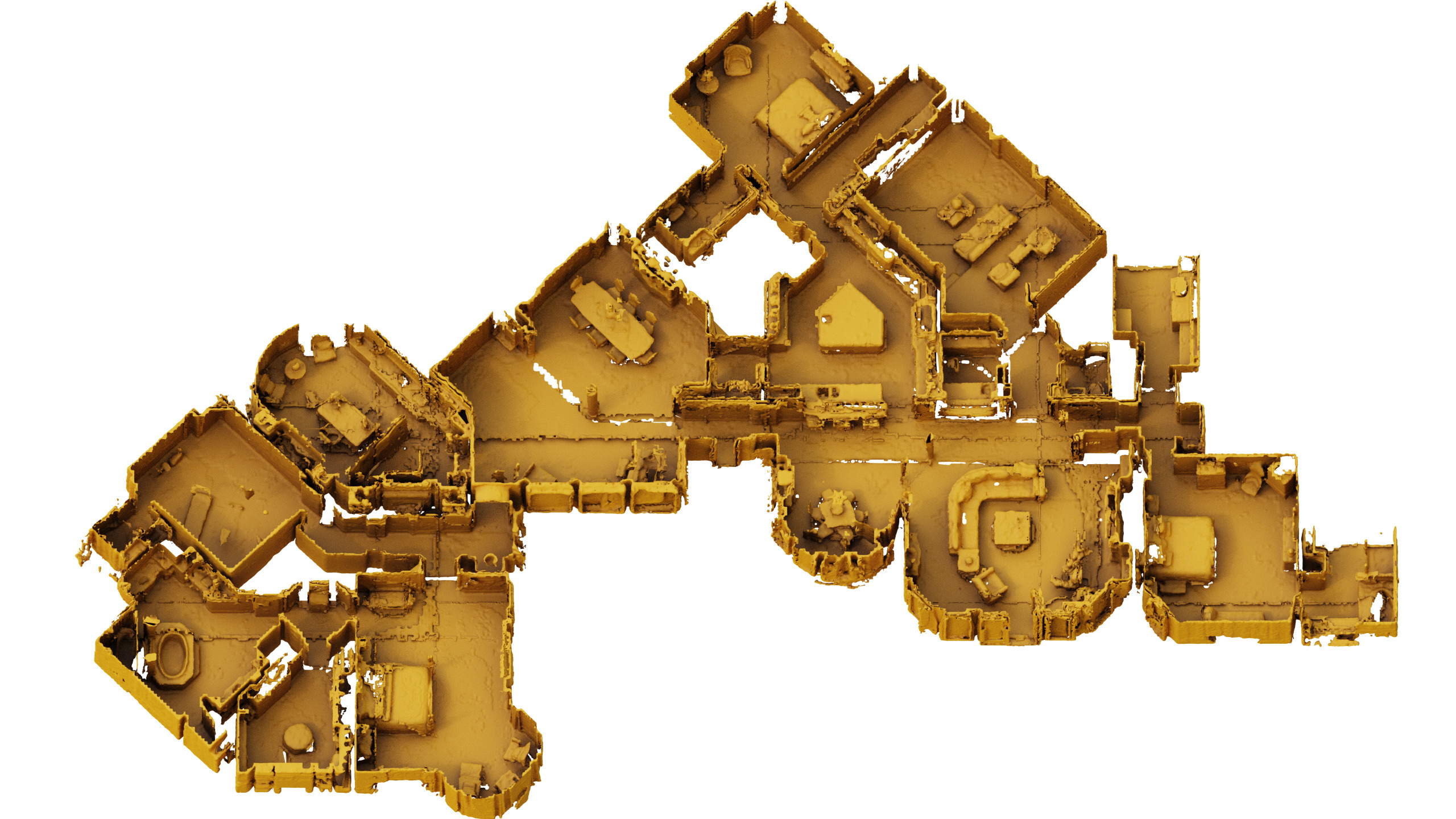}
    \end{subfigure}\\
    \begin{subfigure}[t]{0.65\linewidth}
        \includegraphics[width=\linewidth]{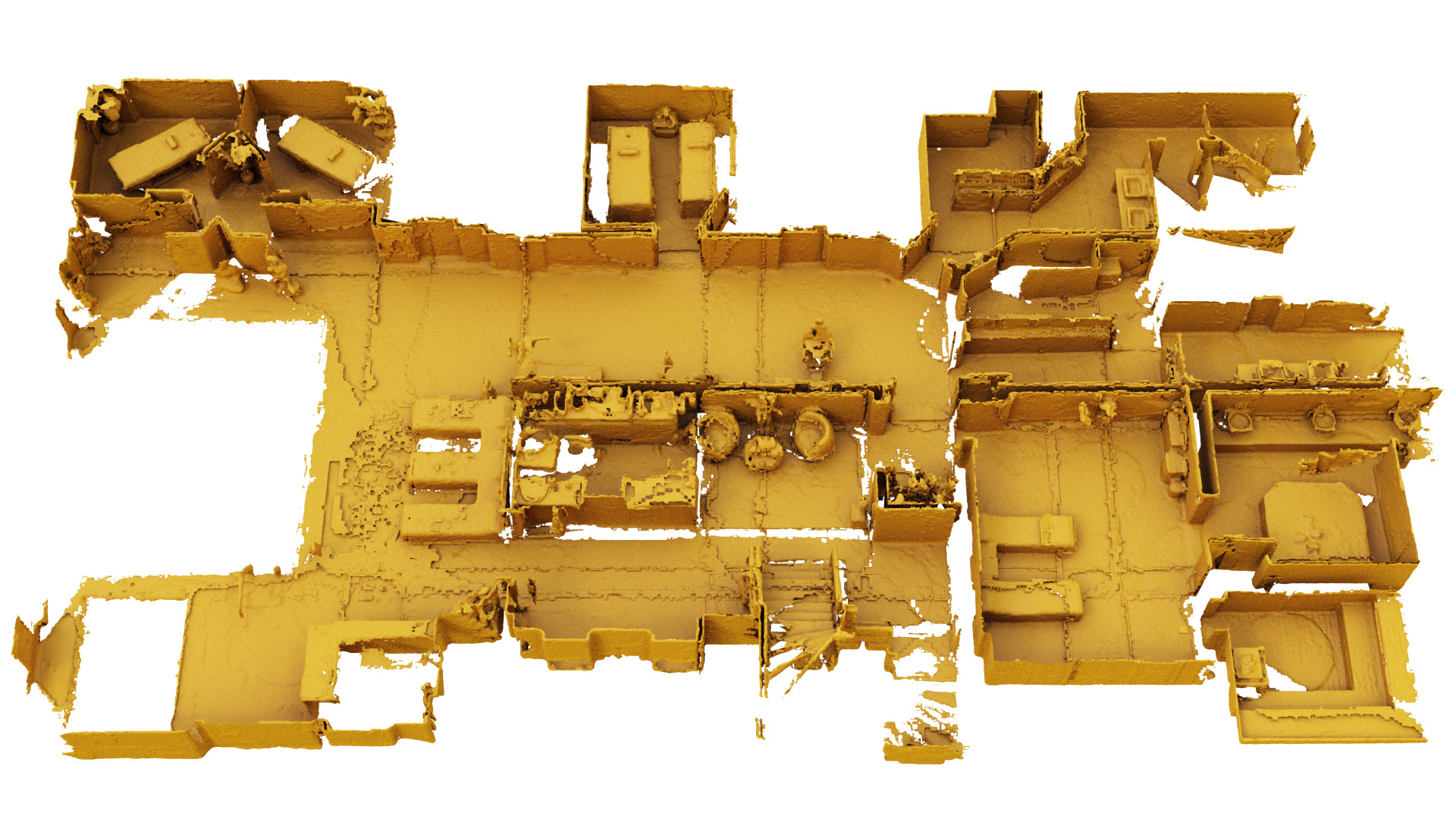}
    \end{subfigure}
    \nextfloat
    \caption{\textbf{Matterport3D scenes}. Matterport sequences mJXqzFtmKg4 and r47D5H71a5s. Stitching effects are caused by scene splitting.}
    \label{fig:scenes_sup}
\end{figure*}

This section presents additional results for the experiments shown in the main paper.
\Cref{app:orivsreg_training} elaborates on the differences between oriented and regular grids during training. \Cref{app:addbaselines} adds additional meshes against baselines, \cref{app:limitation} examples for the current limitations, and~\cref{app:scenes} other scene reconstructions using the proposed encoder.

\subsection{Oriented \textit{vs} Regular grids across training}
\label{app:orivsreg_training}

For regular grids, we solely consider an octree representation for the structure tree with trilinear interpolation. The decoder used is the same as ours and was trained with the same settings. 
This approach contrasts with NGLOD~\cite{takikawa2021neural} in the decoder, where we use an MLP per LOD instead of a shared decoder across LODs, and the output representation (NGLOD uses raycasting to compute occupancy according to sampled cameras, which we change to get occupancies and SDFs directly).
When we switched the output representation, NGLOD performance was significantly degraded without these changes. 

Our approach surpasses regular grids that use an SDF decoder in every aspect. Even though the occupancy framework doesn't perform as well, in both approaches, we notice fewer gaps and artifacts on our meshes, as shown in~\cref{fig:experiments_sdf_occ}.

\Cref{tab:epochs} presents results for our method against regular grids at different epochs. 
We can see a general trend of improvement for both approaches. However, oriented grids outperform regular grids at the first epoch, as shown in Fig. 7 and~\cref{fig:orivsreg_epochs}. Subsequent epochs yield fewer training steps than regular grids; however, oriented grids again outperform regular grids at epoch 30. The normal consistency for oriented grids is consistently better than the ones obtained from regular grids, primarily attributed to smoother planar surfaces and fewer holes in general for our method. This indicates that our method can be used at the first epoch for rendering high-quality meshes.

\subsection{Additional baseline meshes}
\label{app:addbaselines}

\Cref{fig:baselines_sup} shows additional meshes from ABC and Thingi10K for oriented grids {\it vs.} baselines. 

\subsection{Limitation cases}
\label{app:limitation}

We provide some general visualization cases of the limitations of our method and regular grids with SDFs. Due to the normal alignment of our octree and the developed interpolation scheme, the oriented-grid encoder still outperforms the regular grid one. Since both methods have the same type of decoder, similar problems arise. However, our architecture mitigates some of those issues. \cref{fig:lim} shows fewer holes and more robustness for thinner surfaces for oriented grids, which can also be seen for occupancy output in Fig. 6 in the main paper.

\subsection{Additional Matterport3D scenes}
\label{app:scenes}

We show two additional scenes from Matterport3D with oriented grids in~\cref{fig:scenes_sup}. The proposed encoder can reliably obtain high detail in challenging scenes with thin surfaces and complex environments.

}{}

\end{document}